\documentclass[11pt,notitlepage]{article}

 \usepackage[hidelinks]{hyperref}

 \usepackage{colortbl}

\usepackage[super]{nth}
\usepackage{amssymb}
\usepackage{amsmath}
\usepackage{pdfpages}
\usepackage{graphicx}
\usepackage{epstopdf}
\usepackage{pdflscape}
\usepackage{tabularx}
\usepackage{longtable}
\usepackage{setspace}
\usepackage{textcomp}
 \usepackage{graphicx}
   \usepackage{bbm}
\usepackage{makecell}
\usepackage{ragged2e}
\usepackage{float}
\usepackage{geometry}

\usepackage{lscape}
 \usepackage[export]{adjustbox}
\usepackage{morefloats} 
\usepackage{newpxtext}

\usepackage{epsfig}
\usepackage{multirow}
\usepackage{cleveref}
\usepackage{float} 
\usepackage{subcaption}
\usepackage{comment}
\usepackage{caption}
\usepackage{subcaption}
\usepackage{longtable}
\usepackage{float}
\usepackage{threeparttable}
\usepackage{hhline} 
\usepackage{booktabs} 
\usepackage{comment}
\usepackage{tabularx}
\usepackage{pdflscape}

 \usepackage{graphicx}
   \usepackage{bbm}
\usepackage{makecell}
\usepackage{ragged2e}

\usepackage{lscape}
 \usepackage[export]{adjustbox}
\usepackage{morefloats} 
\usepackage{threeparttable}
\usepackage{hhline} 
\usepackage{booktabs} 
\usepackage{comment}
\usepackage{tabularx}
\usepackage{changepage}

\makeatletter
\renewcommand*{\@fnsymbol}[1]{\ifcase#1\or \else\@arabic{\numexpr#1-1\relax}\fi}

\usepackage[style=apa, maxcitenames=2, uniquelist=false]{biblatex}
\defbibfilter{appendixOnlyFilter}{
  segment=1 
  and not segment=0 
}

\DeclareFieldFormat{citehyperref}{%
  \DeclareFieldAlias{bibhyperref}{noformat}
  \bibhyperref{#1}}

\DeclareFieldFormat{textcitehyperref}{%
  \DeclareFieldAlias{bibhyperref}{noformat}
  \bibhyperref{%
    #1%
    \ifbool{cbx:parens}
      {\bibcloseparen\global\boolfalse{cbx:parens}}
      {}}}

\savebibmacro{cite}
\savebibmacro{textcite}

\renewbibmacro*{cite}{%
  \printtext[citehyperref]{%
    \restorebibmacro{cite}%
    \usebibmacro{cite}}}

\renewbibmacro*{textcite}{%
  \ifboolexpr{
    ( not test {\iffieldundef{prenote}} and
      test {\ifnumequal{\value{citecount}}{1}} )
    or
    ( not test {\iffieldundef{postnote}} and
      test {\ifnumequal{\value{citecount}}{\value{citetotal}}} )
  }
    {\DeclareFieldAlias{textcitehyperref}{noformat}}
    {}%
  \printtext[textcitehyperref]{%
    \restorebibmacro{textcite}%
    \usebibmacro{textcite}}}

\makeatletter
\usepackage[page]{appendix}

\usepackage{titlesec}

\usepackage{etoolbox}
\usepackage{etoc}
\newcommand{\sym}[1]{\ifmmode^{#1}\else\(^{#1}\)\fi}

\begin{document}

\etocdepthtag.toc{mtchapter}
\etocsettagdepth{mtchapter}{subsection}
\etocsettagdepth{mtappendix}{none}

\title{Is there ``Secret Sauce'' in Large Language Model Development?\sym{*}
\footnote{\hspace{-0.6cm} \sym{*} Corresponding authors: Matthias Mertens, \text{mmertens@mit.edu}; Neil Thompson, \text{neil\_t@mit.edu}. Massachusetts Institute of Technology, 32 Vassar Street Cambridge, MA 02139, USA.}}

 \author{\Large Matthias Mertens \\ \small Massachusetts Institute of Technology, USA \and \Large Natalia Fischl-Lanzoni\\ \small Massachusetts Institute of Technology, USA  \\  \small \and \Large Neil Thompson \\  \small Massachusetts Institute of Technology, USA  \smallskip }

\sloppy 

\date{ \;\; \\ \;\; \\ \;\; \\ \;\; \\ February 2026    \\ \;\; \\}

\maketitle
\thispagestyle{empty}
 \begin{abstract} 
\noindent 

Do leading LLM developers possess a proprietary “secret sauce,” or is LLM performance driven by scaling up compute? Using training and benchmark data for 809 models released between 2022 and 2025, we estimate scaling-law regressions with release-date and developer fixed effects. We find clear evidence of developer-specific efficiency advantages, but their importance depends on where models lie in the performance distribution. At the frontier, 80–90\% of performance differences are explained by higher training compute, implying that scale---not proprietary technology----drives frontier advances. Away from the frontier, however, proprietary techniques and shared algorithmic progress substantially reduce the compute required to reach \emph{fixed} capability thresholds. Some companies can \emph{systematically} produce smaller models more efficiently. Strikingly, we also find substantial variation of model efficiency \emph{within} companies; a firm can train two models with more than 40x compute efficiency difference. We also discuss the implications for AI leadership and capability diffusion.

 \end{abstract}

\newpage

\onehalfspacing

\clearpage
\setcounter{page}{1}

Large language models (LLMs) have experienced a period of rapid progress and benchmark scores have climbed at an astonishing rate. This rapid progress raises questions about the drivers of improvements in LLM capabilities. Is the frontier of AI advancement propelled by scale; ever-larger models trained on more compute? Or is it fueled by technological progress in the form of openly disseminated algorithmic innovations that raise performance across the field? Or, do leading firms possess a genuine “secret sauce”---proprietary techniques that yield sustained advantages beyond scale and shared algorithmic progress? 

In this paper, we provide empirical evidence that adjudicates among these scenarios by quantifying the components of recent LLM progress. Using a dataset of 809 LLMs released between October 2022 and March 2025 with information on MMLU-Pro benchmark performance and training compute, we decompose observed differences in LLM capabilities into four factors:
\begin{itemize}
    \item \emph{scaling effects}: improvements from models becoming larger, as measured by training compute;
    \item \emph{shared technical/algorithmic progress}: average model efficiency improvements, independent of changes in training compute (e.g., introduction of transformers that are used across the field); 
    \item \emph{developer-specific technology ("secret sauce")}: systematic differences in developer efficiency, whereby firms achieve higher performance using the same amount of compute (e.g., \emph{firm-specific} fine-tuning strategies, high-quality datasets, teacher models, or engineering excellence); and
        \item \emph{model-specific technology}:  all other factors contributing to differences in LLM performance, including \emph{model-specific} specialization, fine-tuning, or optimization. For instance, a Google model fine-tuned in a specific way may exhibit different performance even when trained with identical compute.
\end{itemize}

Understanding the relative importance of these factors informs key questions in contemporary AI: How can firms and countries advance LLM frontier capabilities? Will other countries eventually catch up to the US in frontier AI? And is there a potential threat of malicious actors getting access to advanced AI technologies? Will efficiency improvements diffuse widely and democratize AI, or concentrate rents among a small number of firms with proprietary technologies? How large is the scope for compute efficiency improvements in LLM development?  And, as compute growth slows, will it become a binding constraint on frontier advances, or can algorithmic efficiency sustain past rates of progress? 

While answering these questions directly lies beyond the scope of this paper, our contribution is to provide systematic empirical evidence on the roles of compute-scale and efficiency in shaping LLM performance that can inform these debates.

Our main findings are: 
\textbf{First, there is strong evidence for a "secret sauce" in LLM development.} 14-18\% of LLM performance differences are explained by company-specific effects. 

\textbf{Second, scaling is the most dominant factor in determining \emph{frontier} capabilities} as measured by the highest observed MMLU-Pro benchmark scores: 80\%-90\% of frontier model performance is a consequence of these models' large and increasing compute.

\textbf{Third, although scaling is the dominant driver of frontier performance, algorithmic progress -- either shared or proprietary -- contributes meaningfully to performance differences.} Over our observation window, shared algorithmic progress increased effective compute by a factor of 7.5x. That is, achieving a given MMLU-Pro score in early 2023 would have required 7.5x more compute than achieving the same score in late 2024. Even more striking are the differences in developer- and model-specific compute efficiency. Among smaller models, some developers are up to 61 times more compute-efficient than others, creating a convincing case for a company secret sauce in LLM development. Equally noteworthy, the compute efficiency of individual models after controlling for developer and release-time effects varies substantially, differing by 41x between the 90th and 10th percentiles.

These compute-factor differences are large. Yet, training compute itself also increased dramatically, which puts these numbers into perspective. At the performance frontier, model compute grew by a factor of roughly 5,000 in our data. Thus, while algorithmic progress, company secret sauce, and model-specific factors cause substantial variation in effective compute across models, their impact is dominated by the sheer growth in compute resources. This reiterates our first result: compute growth is the primary driver of performance improvements.

\textbf{Algorithm progress is particularly important for model development that is \textit{not} at the frontier---our fourth key result.} Technical progress has enabled the development of much smaller (and cheaper) models that achieve performance levels previously attainable only by far larger systems. Among major developers, the required compute to reach a 15\% normalized MMLU-Pro score dropped by a factor of 50 in our data, and including smaller developers the factor of compute reduction is close to 8,000. Proprietary technologies (secret sauce) and shared technological advances are equally important in driving these efficiency gains, which are presumably also a root cause of declining inference prices.

Taken together, our results show that scale is the primary driver of frontier performance. This suggests that it is unlikely that any single firm (or country) can sustain AI dominance without securing superior access to compute, and developers lacking such access are likely to fall behind at the frontier. Below the frontier, however, shared and proprietary technologies play an important role and account for substantial compute efficiency differences in the data. While we provide the first robust evidence of a genuine “secret sauce” in LLM development, it manifests less in pushing the frontier itself than in building more efficient, smaller models that deliver relatively high performance with lower compute levels. 

 \paragraph{Related work.}
LLM performance gains follow predictable scaling laws, with accuracy improving as training compute increase (\cite{Henighan2020, kaplan_scaling, rosenfeld2021scaling, hoffmann2022trainingcomputeoptimallargelanguage, chen2024scaling}). Recent work highlights that algorithmic advances further enhance compute efficiency  (\cite{sanderson2025rethinking, ho2024algorithmicprogresslanguagemodels}). We extend previous work by decomposing improvements into scale, shared algorithmic progress, proprietary technology (secret sauce), and remaining model-specific differences. 
Finally, our analysis relates to evidence on declining inference costs (\cite{epoch2025llminferencepricetrends, Gundlach2025}). Our results suggest that these reductions may partly reflect developer- and model-specific efficiency gains, which implies that, even as prices fall, some firms retain a competitive edge.

\section{Results}\label{Results}

\begin{figure}[H]
    \caption{Shapley Variance Decomposition, Different Samples}
    \label{fig: shapley_main}
    \centering

    \begin{subfigure}[t]{0.49\textwidth}
        \centering
        \caption{All 809 LLMs (baseline)}
        \includegraphics[width=\textwidth, height=4cm, keepaspectratio]{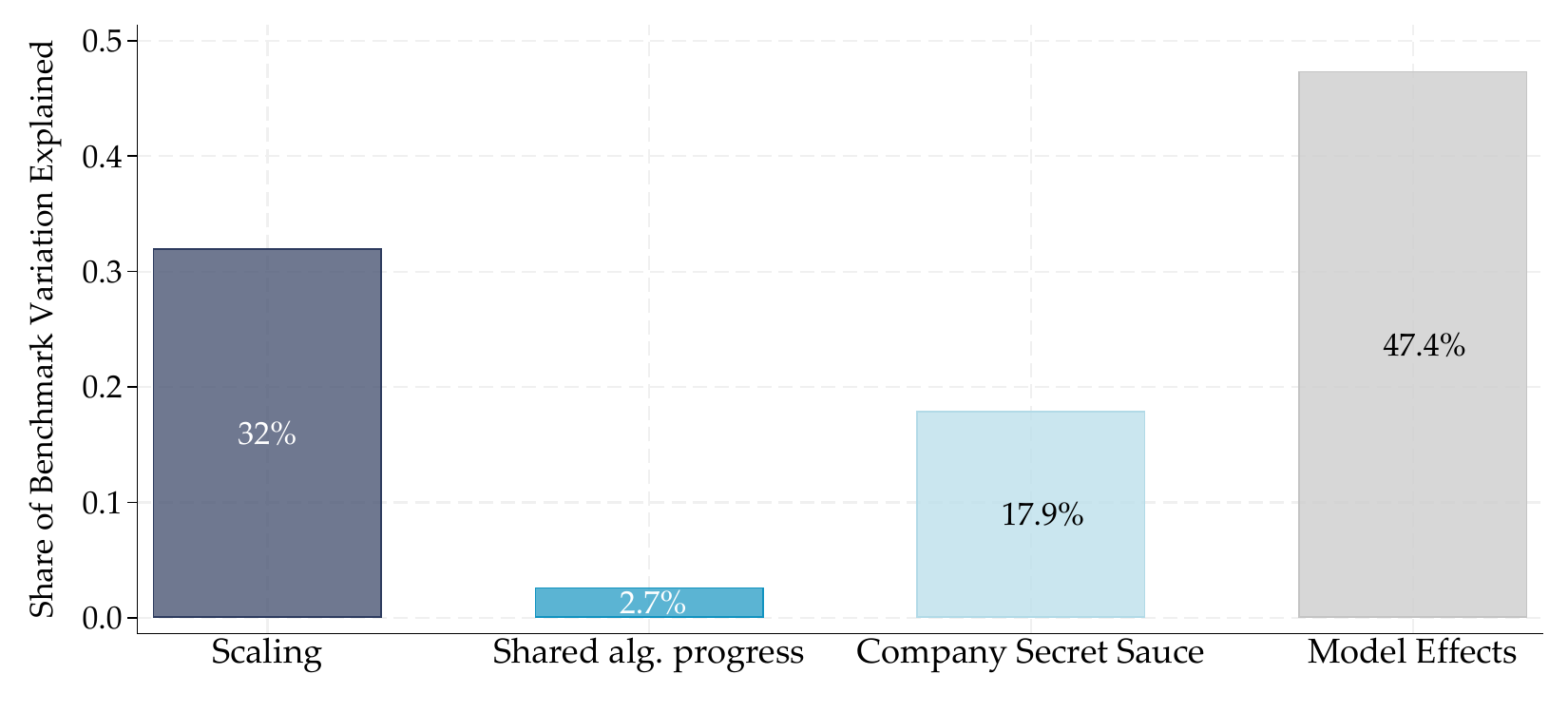}
    \end{subfigure}
    \hfill
    \begin{subfigure}[t]{0.49\textwidth}
        \centering
        \caption{Major Developers only (all sizes) — 122 LLMs}
        \includegraphics[width=\textwidth, height=4cm, keepaspectratio]{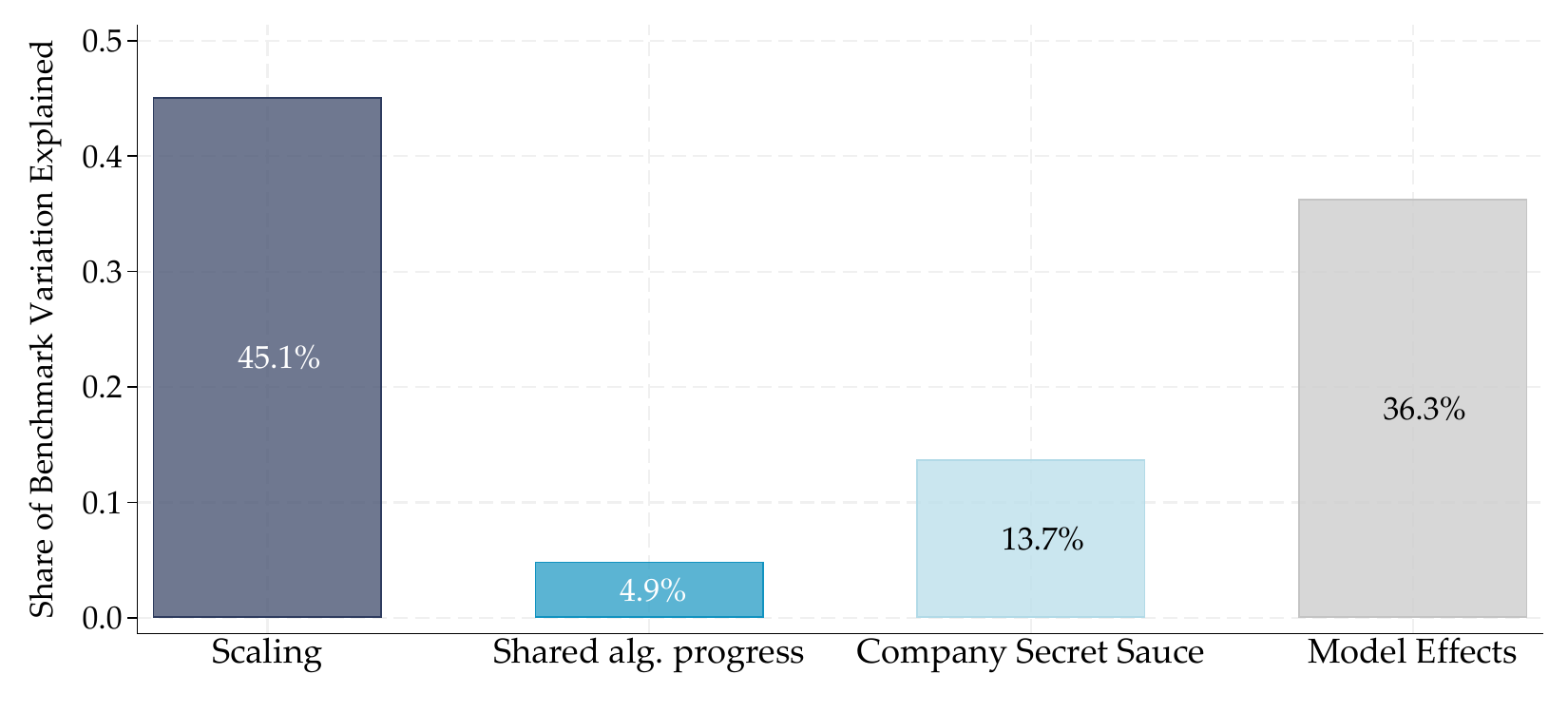}
    \end{subfigure}

    \begin{subfigure}[t]{0.49\textwidth}
        \centering
        \caption{Major Developers, Large Models — 61 LLMs}
        \includegraphics[width=\textwidth, height=4cm, keepaspectratio]{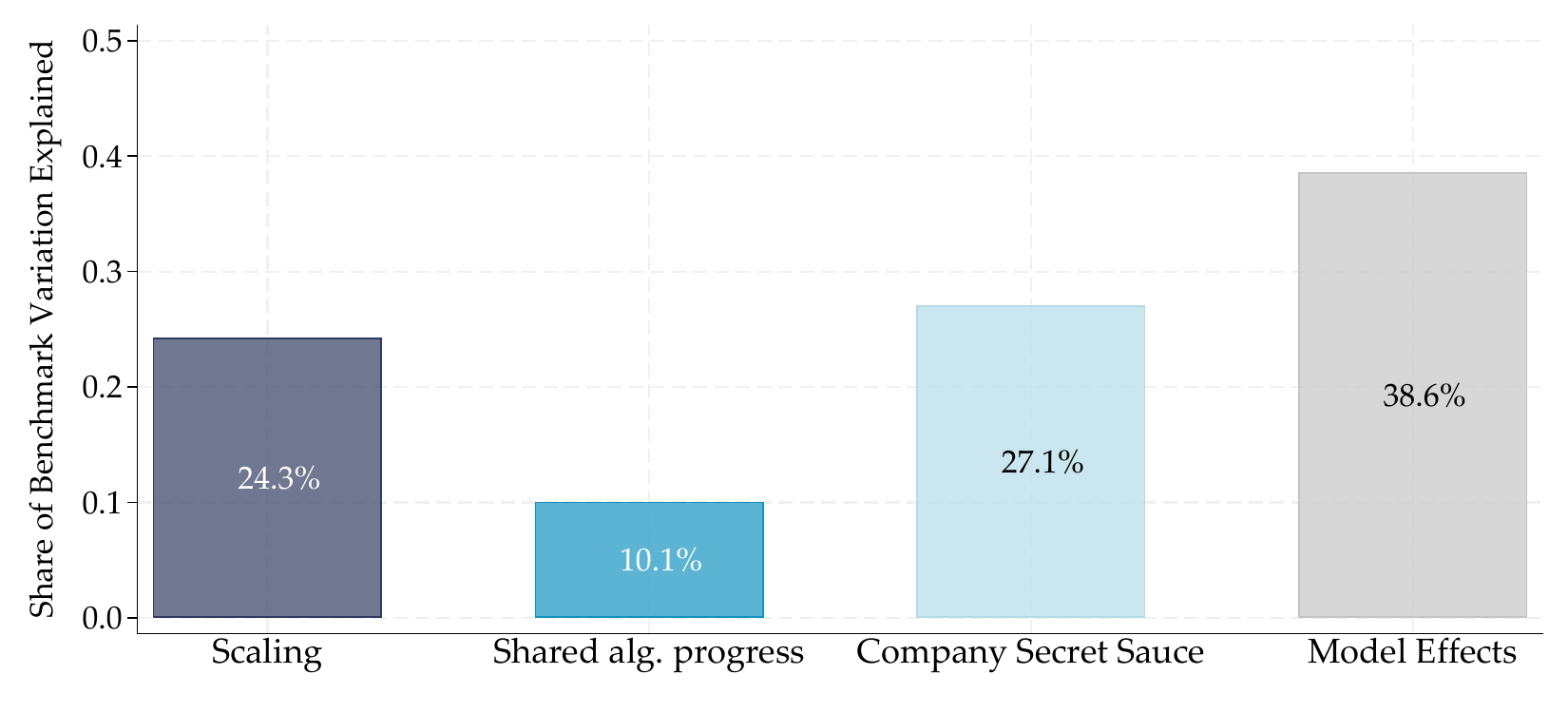}
    \end{subfigure}
    \hfill
    \begin{subfigure}[t]{0.49\textwidth}
        \centering
        \caption{Major Developers, Small Models — 61 LLMs}
        \includegraphics[width=\textwidth, height=4cm, keepaspectratio]{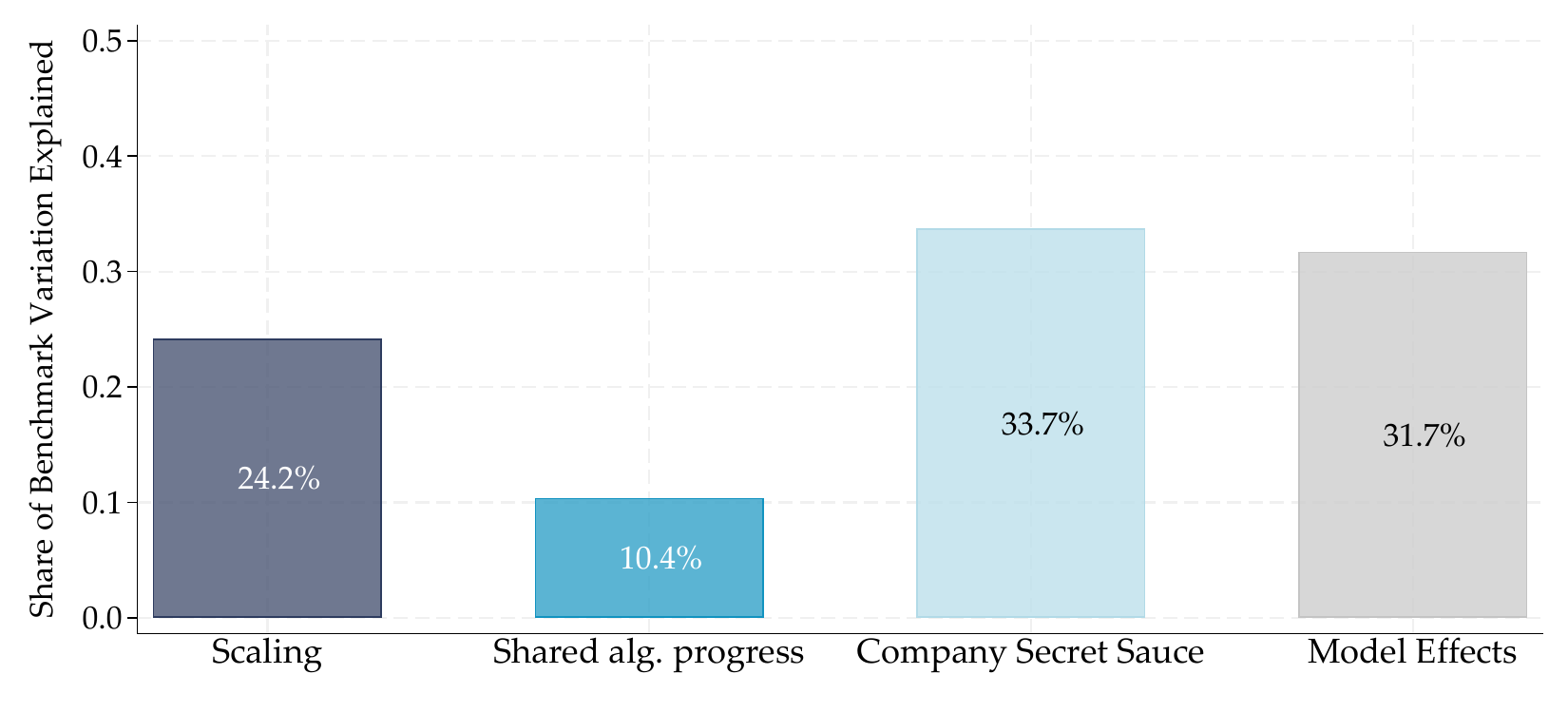}
    \end{subfigure}

    \vspace{-0.6em}

    \begin{minipage}{\textwidth}
        \scriptsize \singlespacing \textit{Notes:}
        Each panel reports a Shapley decomposition of the regression $R^2$ into contributions
        from scale, shared algorithmic progress, and company factors for the indicated sample using the command shapley2 in Stata. The remaining performance variation is explained by residual variation (model effects).    The baseline sample (Panel (a)) includes all 809 LLMs. Subsequent panels restrict attention to      major developers only, and further split the sample by model size (i.e., training compute) at the median of the 122 main developer LLMs. Major developers include: Deepseek, Qwen, Meta, Google, Microsoft, OpenAI, Anthropic, X-AI, 01-AI, and Nvidia.  
    \end{minipage}
\end{figure}

\paragraph{Regression Analysis}
We estimate a regression of LLMs' logit-transformed MMLU-pro benchmark scores, $Y_{i}= \ln\left(\frac{y_{i}}{1-y_{i}}\right)$, where $y_{i}$ denotes the benchmark score, on LLMs' log training compute ($c_i$), a set of dummies for publication periods ($\boldsymbol{{\delta}}_t$), and developer company dummy variables ($\boldsymbol{{\nu}}_j$):
\begin{equation}
 Y_{i} =  \beta_{0} + \underbrace{\beta_{c} \log(c_i)}_{\text{Scaling}}   + \underbrace{\boldsymbol{\delta}_t}_{\text{Shared technical efficiency}} + \underbrace{\underbrace{\boldsymbol{\nu}_j}_{\text{company technical efficiency}}}_{\text{"Secret Sauce"}} + \underbrace{\varepsilon_i}_{\text{Model technical efficiency}},  \label{main_reg}
\end{equation}
where $\varepsilon_i$ captures residual performance not explained by scaling and time and developer effects.\footnote{As discussed in Section \ref{data}, we apply a logit transformation, consistent with the inverse sigmoidal relationship between downstream performance and log compute shown by \textcite{ruan2024observational}.} 
We calculate training compute from training parameters ($N_{i}$) and tokens ($D_{i}$), using $c_i=6\times N_{i}\times D_{i}$ (\cite{kaplan_scaling}). We define periods ($\boldsymbol{{\delta}}_t$) via a group of dummies for three (almost) equally sized periods 2022q4-2023q3, 2023q4-2024q2, and 2024q3-2025q1, with the baseline period 2022q4-2023q3 (time effects are thus expressed relative to 2022q4-2023q3---we show these results are robust to other specifications in Section \ref{data}). Company effects ($\boldsymbol{{\nu}}_i$) are defined by dummies for major companies/model groups (Deepseek, Qwen, Meta, Google, Microsoft, OpenAI, Anthropic, X-AI, 01-AI, and Nvidia) and for the broader groups of small, medium, and large other developers (size measured in average model training compute), with small other developers as baseline (developer effects are thus expressed relative to this group).\footnote{We describe the developer selection criteria in Appendix \ref{detailed_data_methods}. Results are robust to adjusting the set of main developers.} Because Eq. \eqref{main_reg} models a log-additive structure, we can express time ($\boldsymbol{\delta}_t$), company ($\boldsymbol{\nu}_t$), and model ($\varepsilon_i$) effects as factors that increase \emph{effective} compute. We explain this in detail in Section \ref{data}, where we also show that Eq. \eqref{main_reg} is an extension of previous scaling laws that structurally models algorithmic/technical efficiency by time, company, and model effects. We estimate Eq. \eqref{main_reg} using data on 809 LLMs (described in Section \ref{data}).

\paragraph{Results: Variance Decomposition.}\label{reg_results}

Figure \ref{fig: shapley_main} presents a Shapley decomposition of variation in MMLU-Pro benchmark scores based on OLS estimates of Eq. \eqref{main_reg} (regression coefficients in Appendix Table \ref{tab:reg} ). Panels (a)–(d) show results for different samples, with our baseline specification---used throughout the paper---shown in Panel (a).

Across all models, training compute (scaling) explains 32\% of variation in LLM performance. Its contribution increases when focusing on major developers (45\%), as excluding smaller (often non-profit) developers removes idiosyncratic factors that affect performance variation. Shared algorithmic progress contributes relatively little (3\%-10\%) in explaining performance \emph{differences} across LLMs, with stronger effects when we focus on small or large major developer models (which limits variation in model compute). In contrast, proprietary technologies captured by company effects (“secret sauce”) account for a comparably large share of performance differences (14\%-34\%). Finally, a sizable share of variation is captured by model-specific residual factors (32\%-47\%), likely reflecting experimentation, fine-tuning choices, model specialization, time-varying company-specific innovations not captured by a fixed company effect, or other model training idiosyncrasies.

\begin{figure}[H]
    \caption{Main Results}
    \label{fig: mainresults_main}
    \centering
    \begin{subfigure}[t]{0.49\textwidth}
        \centering
            \caption{Scaling and Performance}
        \includegraphics[width=\textwidth, height = 5cm]{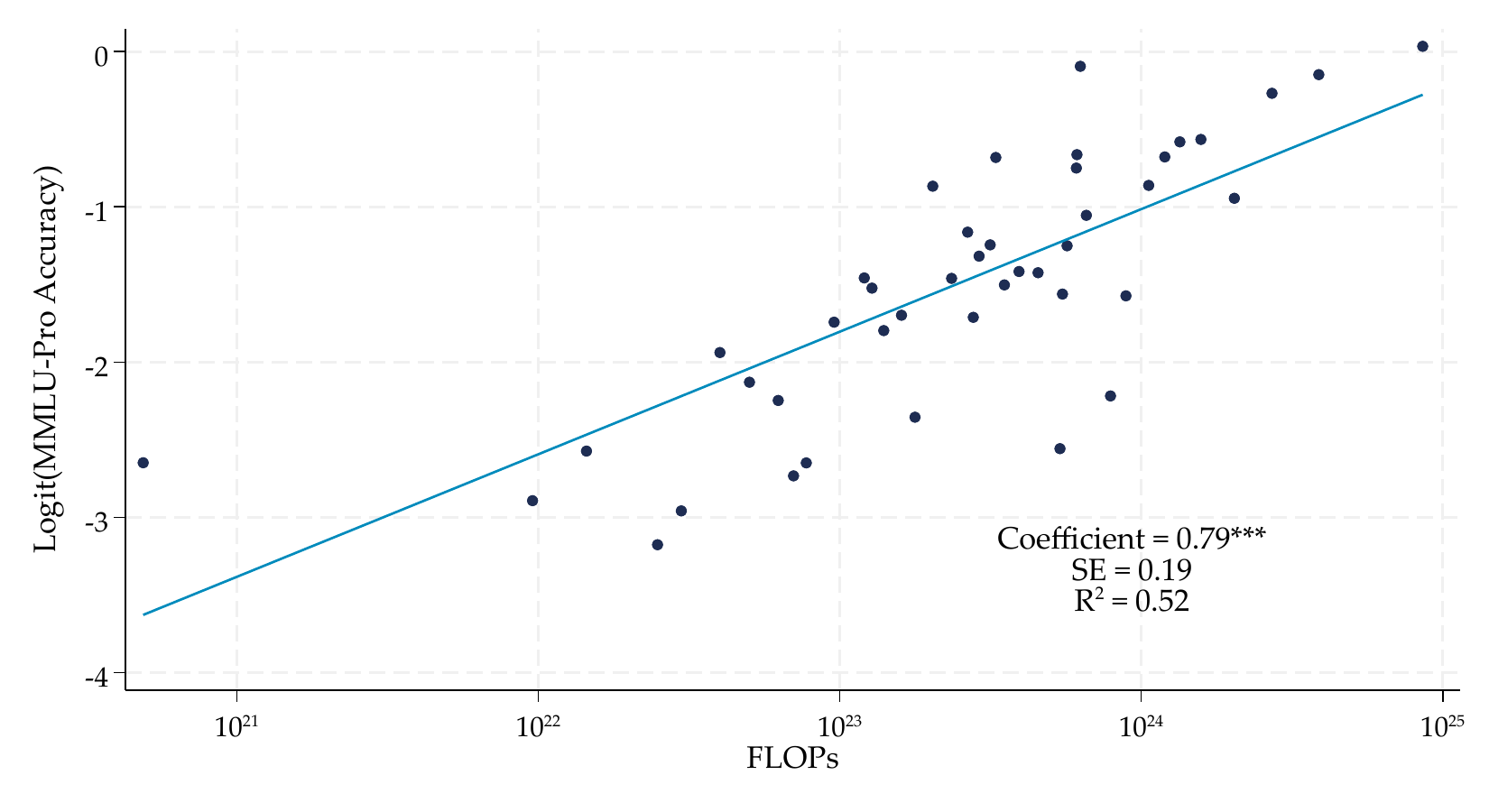}
        \vspace{-0.5em}
    
    \end{subfigure}
    \hfill
    \begin{subfigure}[t]{0.49\textwidth}
        \centering
          \caption{Shared Alg. Progress: Compute Efficiency Gain}
        \includegraphics[width=\textwidth, height = 5cm]{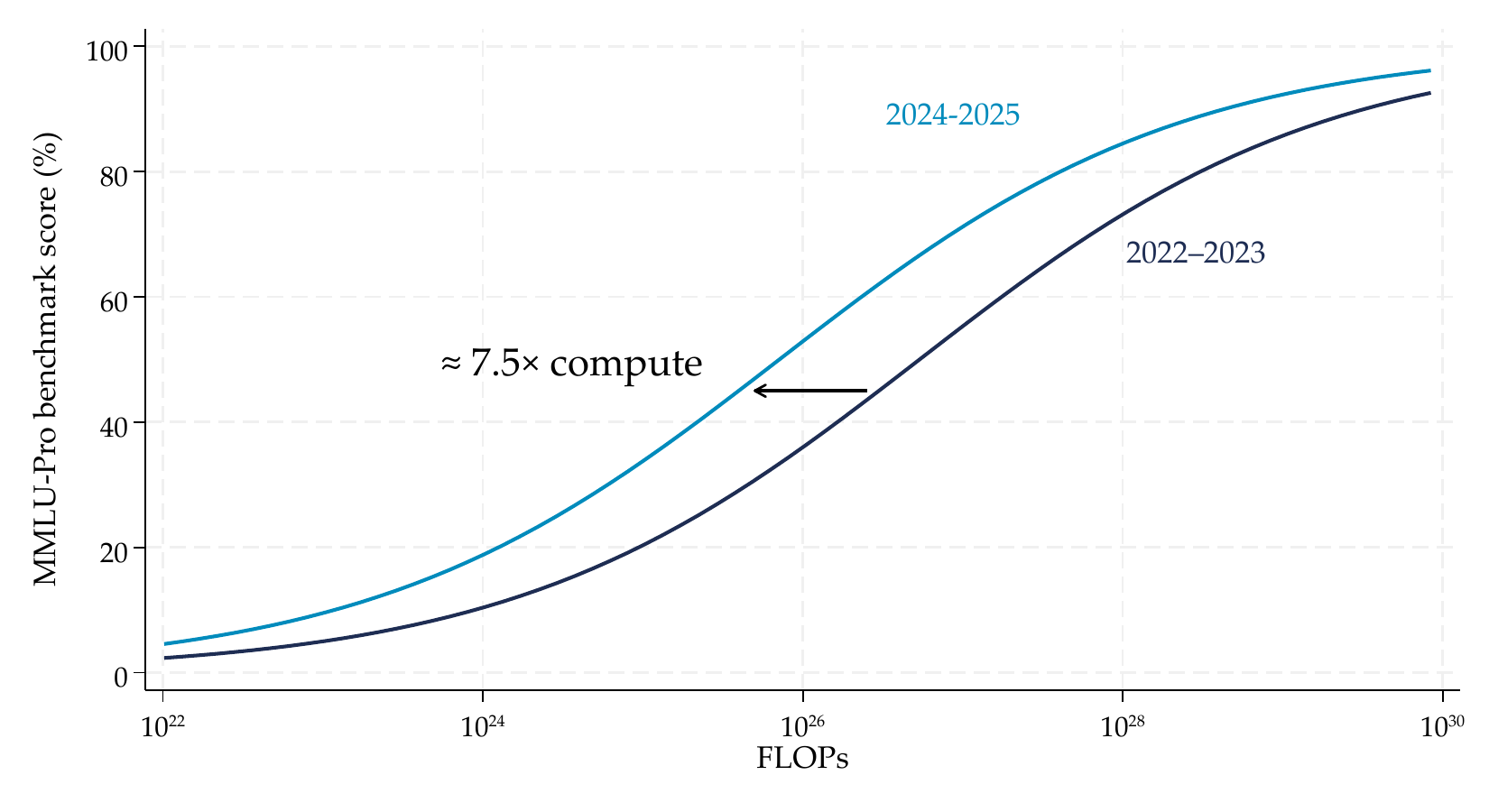}
        \vspace{-0.5em}
      
    \end{subfigure}
    
    \vspace{-0.3em} 

    \begin{subfigure}[t]{\textwidth}
        \centering
                \caption{Company Secret Sauce and Model Effects in Compute Efficiency }
        \includegraphics[width=\textwidth, height=8cm, keepaspectratio]{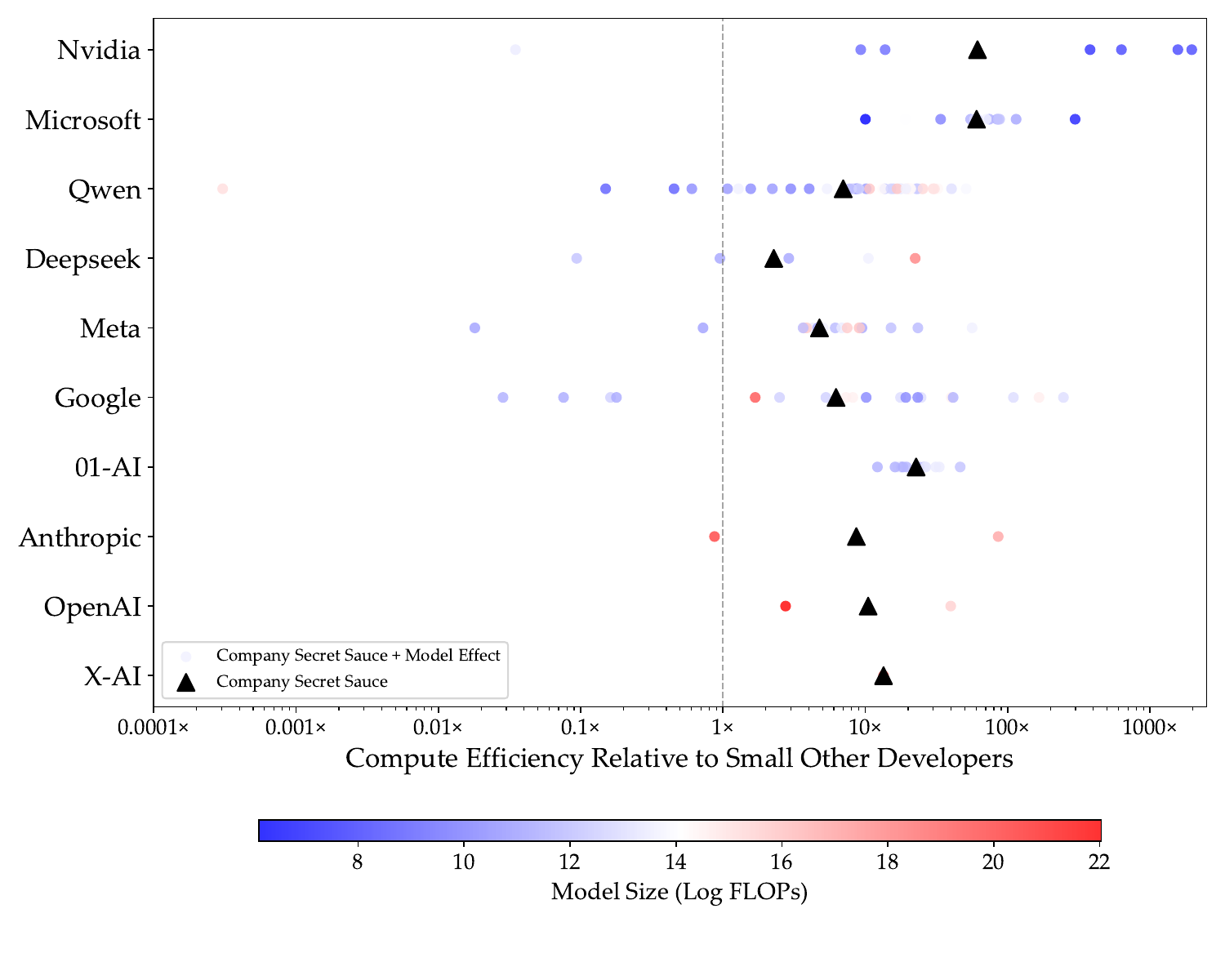}
        \vspace{-0.5em}

    \end{subfigure}
    
    \begin{subfigure}[t]{0.49\textwidth}
        \centering
              \caption{Company Secret Sauce and Relative Model Compute}
        \includegraphics[width=\textwidth, height=5cm, keepaspectratio]{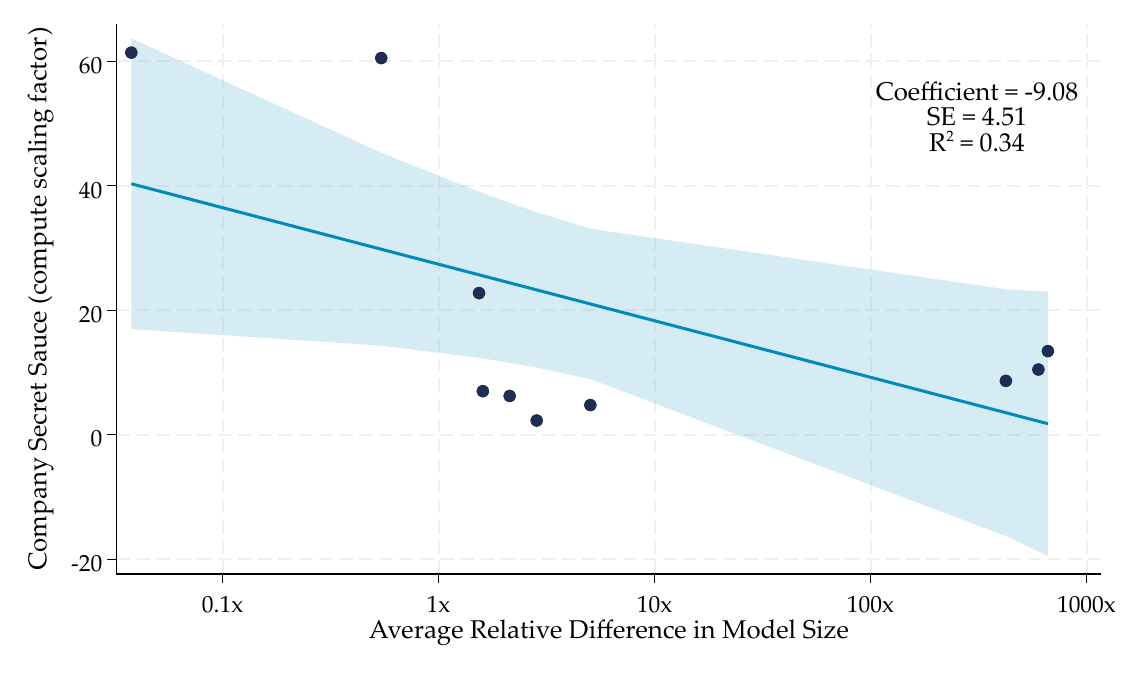}
        \vspace{-0.5em}
  
    \end{subfigure}
    \hfill
    \begin{subfigure}[t]{0.49\textwidth}
        \centering
                \caption{Model-Specific Effects in Compute Efficiency}
        \includegraphics[width=\textwidth, height=5cm, keepaspectratio]{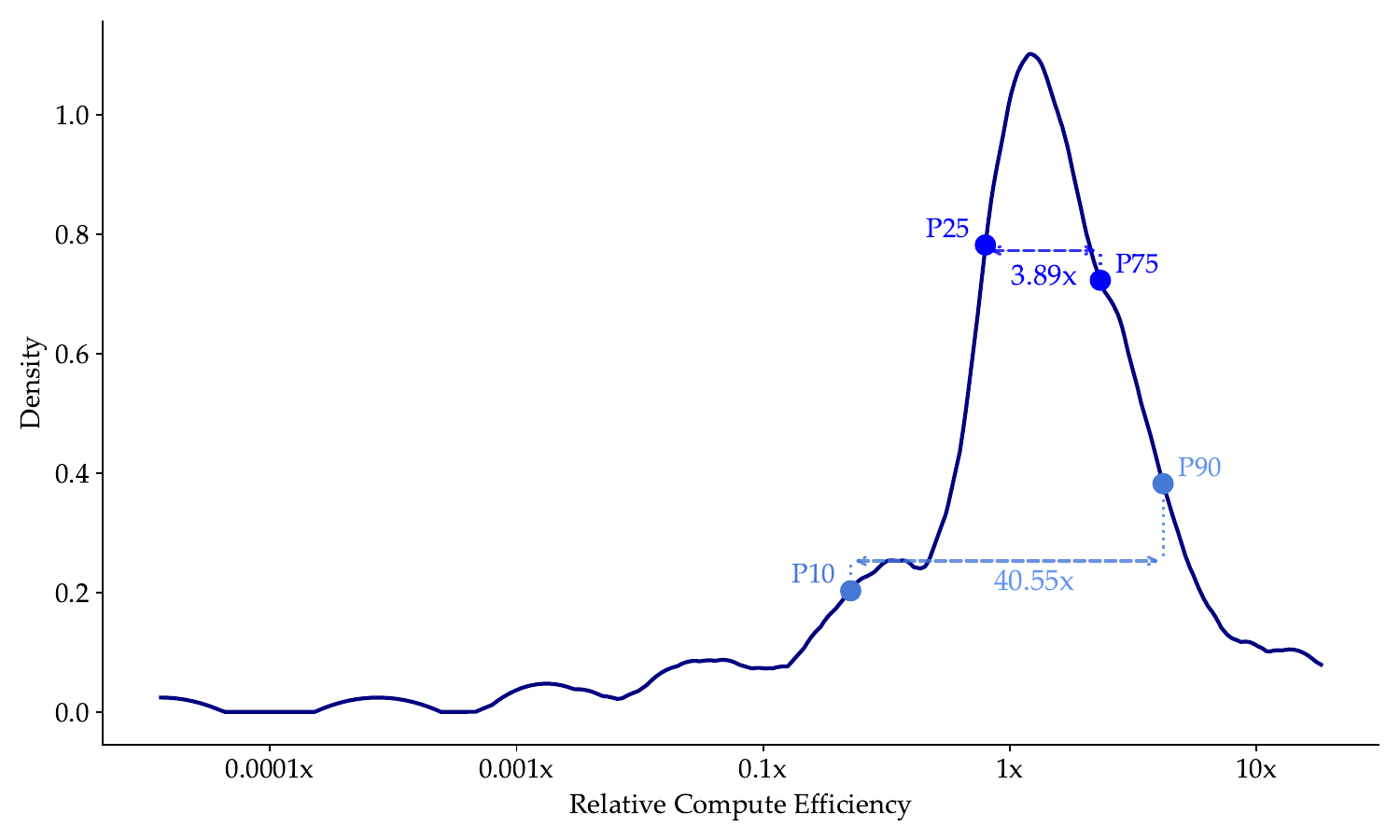}
        \vspace{-0.5em}

    \end{subfigure}
    
    \vspace{-0.3em} 

    \begin{minipage}{\textwidth}
        \scriptsize \singlespacing \textit{Notes:} Panel (a) visualizes Eq. \eqref{main_reg} using a binned scatter plot that absorbs developer and period effects. Panel (b) shows the implied compute–performance curves for the first and last periods as predicted by the regression.  Panel (c) reports the estimated developer fixed effects (triangles). The shaded dots add model effects ($\varepsilon_i$) to the developer fixed effects with colors indicating the compute of the models. Panel (d) uses a binned scatter plot to project company effects against the average of model compute deviations from the average compute of published models in a given period (log differences). Panel (e) displays the distribution of model effects/residuals for main developers ($\varepsilon_i$). Significance levels in Panel (a) and (d): \sym{*} $p<0.10$, \sym{**} $p<0.05$, \sym{***} $p<0.01$. Panel (d) shows 95\% confidence bands.
    \end{minipage}
\end{figure}

\paragraph{Results: scaling, shared progress, secret sauce, and model-specific factors.}\label{reg_results}
Figure \ref{fig: mainresults_main} further investigates each performance driver. Increasing compute is a strong predictor of higher model performance. As shown in Panel (a), a ten-fold increase in model compute increases the log-odds of performance by 0.79  (controlling for period and company effects). 

 Panel (b) visualizes the impact of shared algorithmic progress by plotting predicted benchmark scores for different compute levels. Shared algorithmic progress (measured by period fixed effects) increased effective compute by a factor of 7.5x. That is, matching the MMLU-Pro performance achieved in 2024q3–2025q1 would have required 7.5× more compute in 2022q4–2023q3, indicating significant technological advances that increase LLM performance. 

Panel (c) shows the estimated developer fixed effects (“secret sauce”). Triangles report developer effects expressed as compute factors that scale \emph{effective} compute. Recall: developer effects are normalized relative to small “other” developers. Shaded dots add model-level residuals to each developer effect, illustrating within-developer heterogeneity in performance (discussed further in Panel (e)).

Several patterns emerge. First, all major developers achieve higher performance per unit of compute than small developers, as indicated by positive developer effects. Second, developer compute efficiency varies significantly: for example, DeepSeek is estimated to use compute 2.3× more efficiently than small developers, while Microsoft’s estimated compute factor is 60.5×. Third, there is weak evidence that developers focusing on smaller models exhibit higher compute efficiency as suggested by the color gradient of the model-level dots, with blue indicating smaller models. This pattern may reflect that proportional improvements at smaller compute baselines yield smaller absolute gains and/or that repeated experimentation and optimization are easier at smaller scales. Panel (d) examines this relationship formally by regressing developer compute factors on average relative model compute (within release cohorts). While the estimated coefficient implies a large effect---developers producing 10× larger models have a 6.6-point lower compute factor---it is imprecise and not statistically significant. Fourth, although developer compute factors are sizable, they should be interpreted relative to the extreme dispersion in training compute: models at the 95th percentile use 1,321× more compute than those at the 5th percentile (not shown here). Taken together, the results reveal substantial developer-specific differences in compute efficiency, providing clear evidence of a meaningful “secret sauce” in LLM development that shapes performance outcomes.

Panel (e) plots the distribution of model-specific factors ($\varepsilon_i$) for our 10 main developers and reveals substantial dispersion in model-specific compute efficiency, possibly reflecting experimentation, domain specialization, or model-specific fine-tuning. Models at the 90th percentile translate compute 41× more efficiently into benchmark performance than models at the 10th percentile. Recall that these residuals come from a regression with developer and period fixed effects that controls for log compute. Hence, the figure reflects effective compute differences net of publication time differences between models coming from the \emph{same developer}.

\begin{figure} [H]
    \caption{Contributions to Top Model Over Time}
  \label{base_top_models_main}
    \centering
    \begin{subfigure}[b]{0.89\textwidth}
    \caption{Top Scoring Models}
    \includegraphics[width=\textwidth]{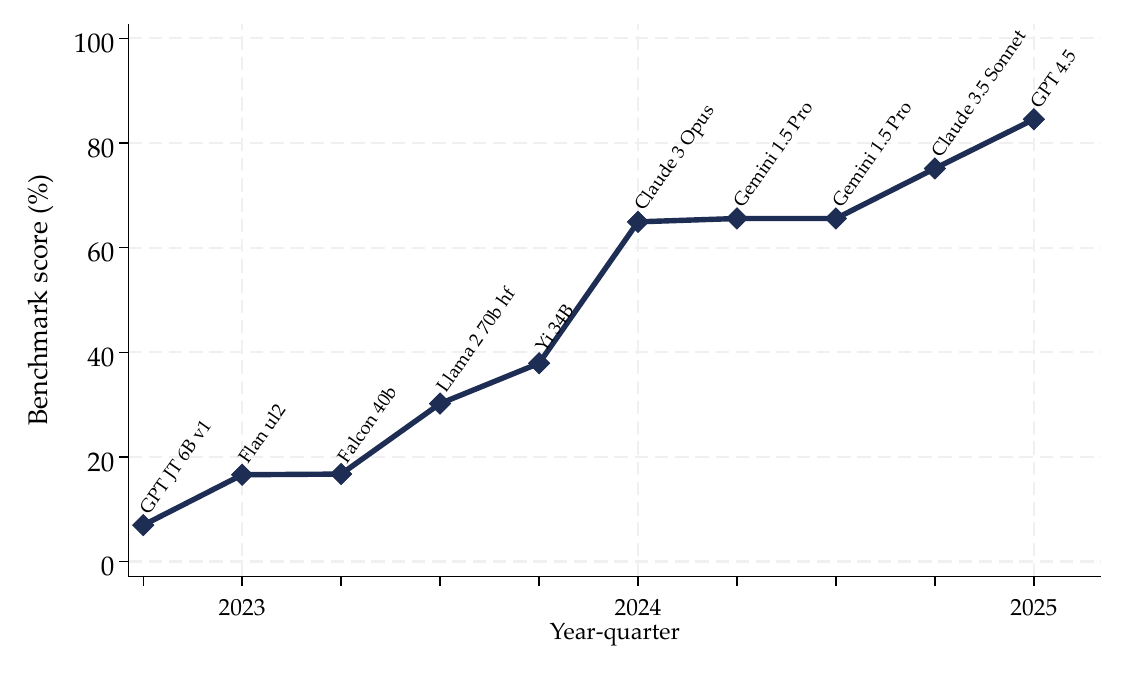}
    \end{subfigure}
    \begin{subfigure}[b]{0.89\textwidth}
    \caption{Required FLOPs for a Given MMLU-Pro Score}
    \includegraphics[width=\textwidth]{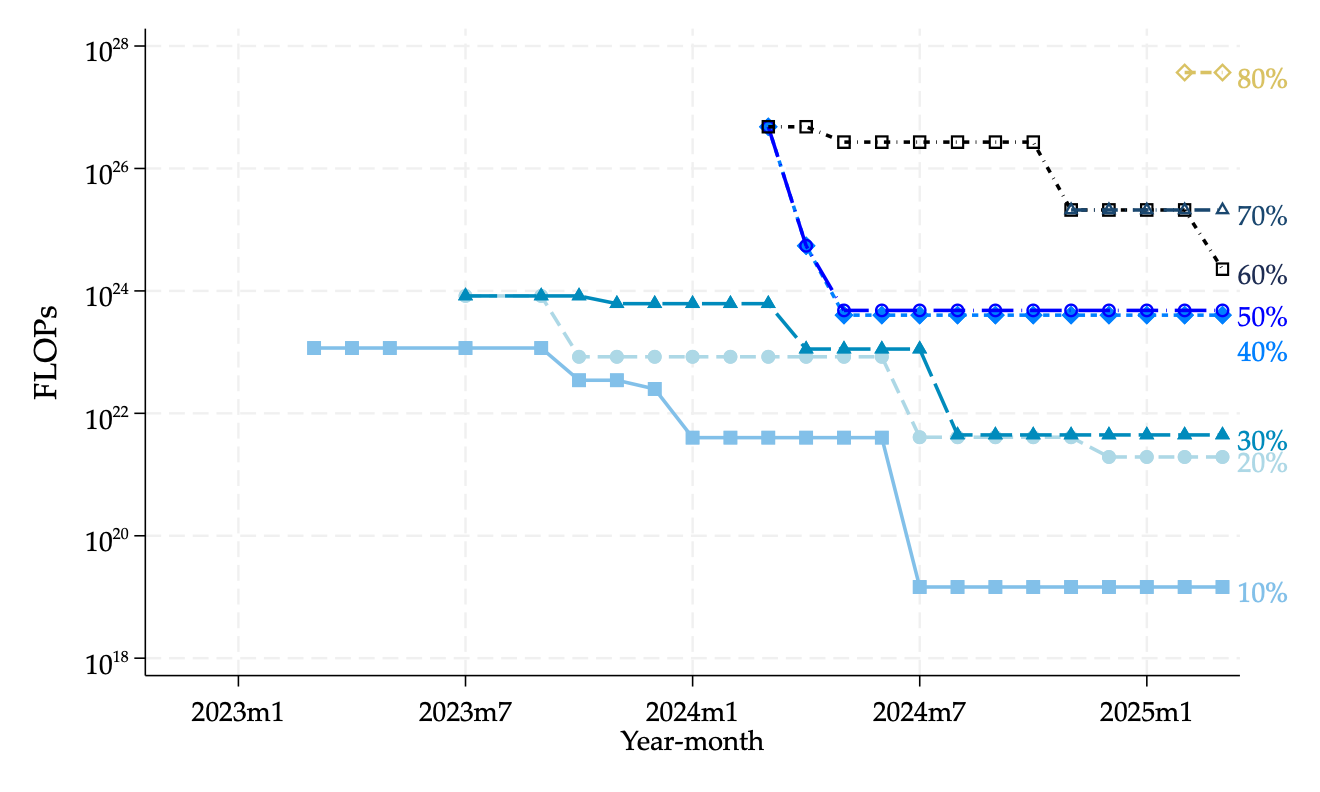}
    \end{subfigure}
                   \begin{minipage}{\textwidth}
        \scriptsize\singlespacing \textit{Notes:} Panel (a) shows the top scoring model on the normalized MMLU-Pro benchmark each quarter (MMLU-pro scores are normalized to account for random guessing). Panel (b) shows, over time, the smallest models (in FLOPs) in the data that reach a given MMLU-Pro performance level. 
     \end{minipage}

    \end{figure}

\paragraph{LLM-development.}
What do these findings imply for LLM development? To provide context, Figure \ref{base_top_models_main} presents two key trends in LLM-development which we aim to better understand below. Panel (a) shows the remarkable performance increase in top scoring models in our data. Panel (b), shows that the minimum model compute required to reach a given benchmark score has fallen rapidly.

\begin{figure} [H]
    \caption{Sources of Performance Growth: Frontier Models and Smaller, Efficient Models}
  \label{base_meek_models_main}
    \centering
    \begin{subfigure}[b]{0.95\textwidth}
    \caption{Source of Benchmark Score Growth: How Top Performing Models Become Better}
    \includegraphics[width=\textwidth]{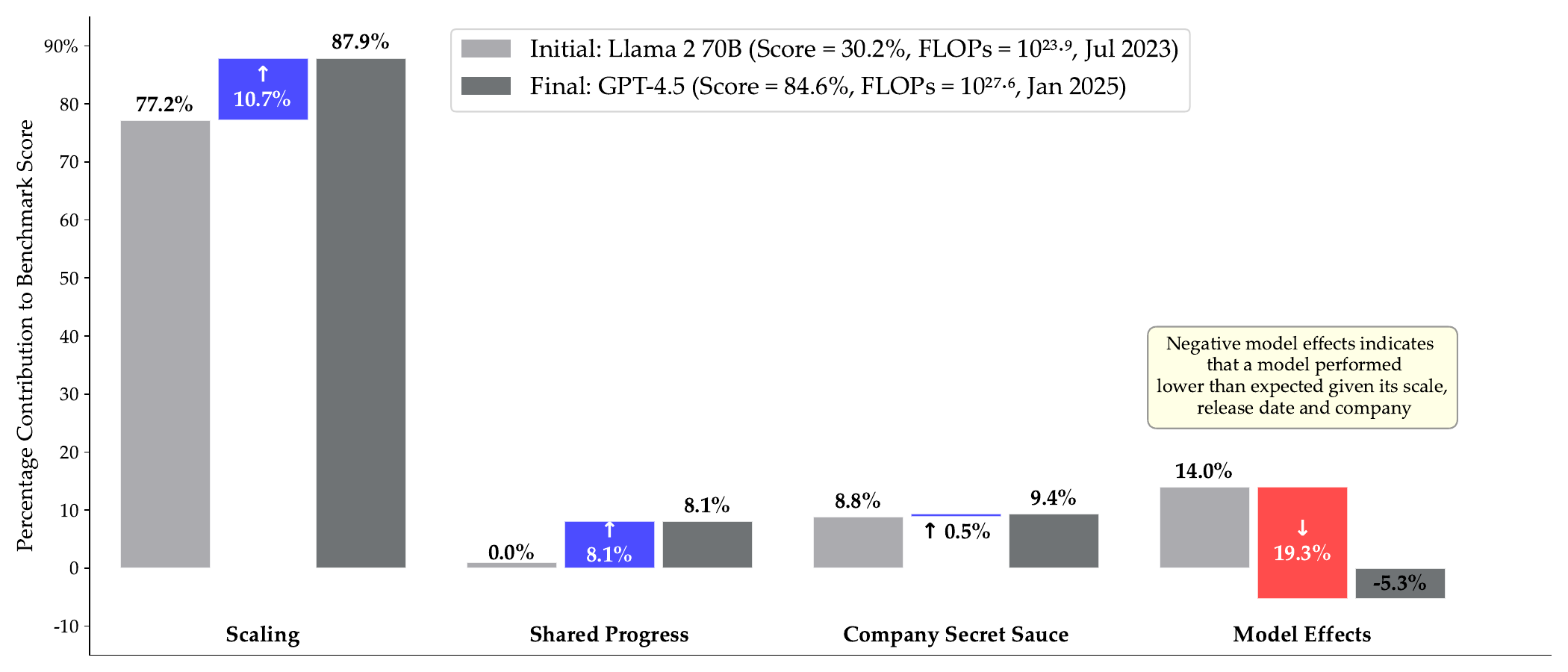}
    \end{subfigure}
    \begin{subfigure}[b]{0.95\textwidth}
    \caption{Source of Benchmark Score Growth: How Models Become More Efficient}
    \includegraphics[width=\textwidth]{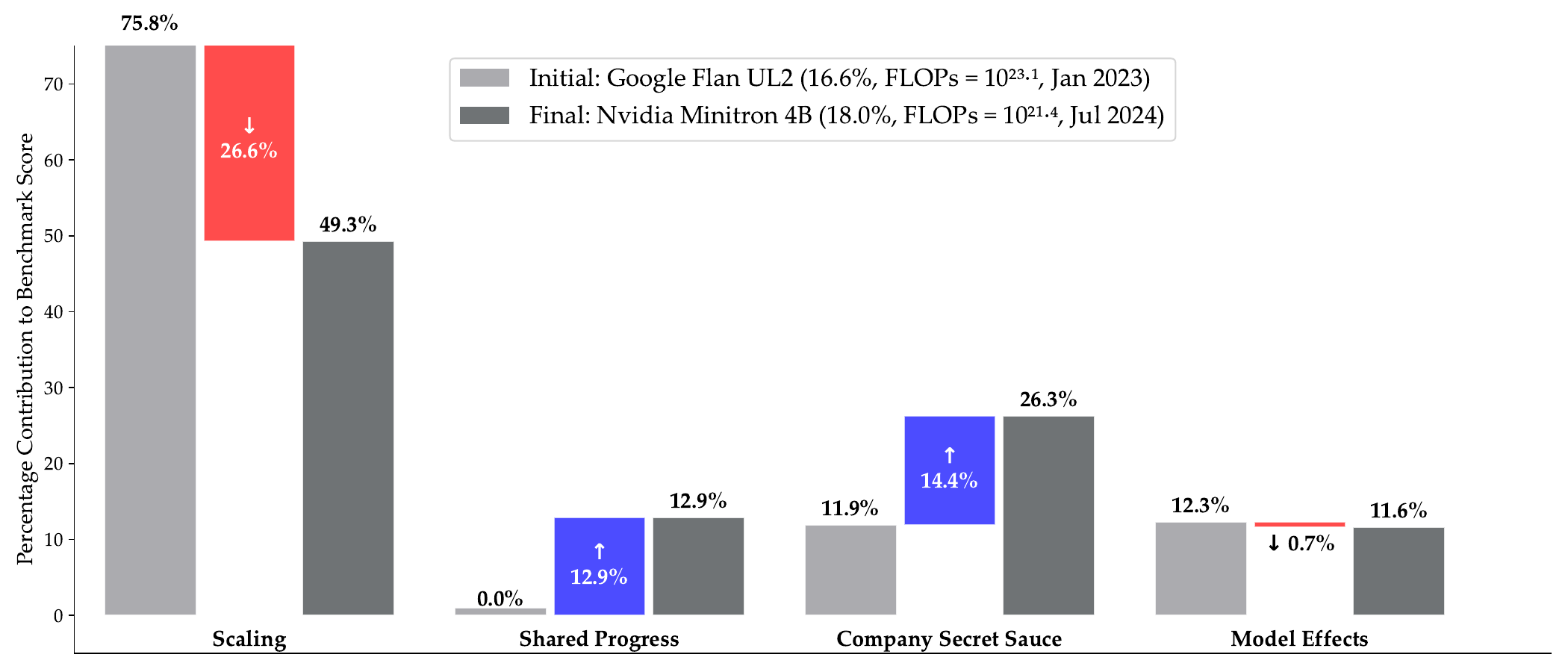}
    \end{subfigure}
          \begin{minipage}{\textwidth}
        \scriptsize\singlespacing \textit{Notes:} Panel (a) applies our regression-based decomposition to the highest-performing models for MMLU-Pro in the first and last time periods of our sample. Panel (b) applies our regression-based decomposition to the first model and the smallest models to achieve at least 15\% on MMLU-Pro. We compute contribution shares of scaling, shared progress, company effects, and model effects by summing the estimated components of Eq. \eqref{main_reg} in logit-space and calculating each component’s share of the total. These logit-space shares are then visually applied to the benchmark scores to ease interpretation. MMLU-Pro scores are normalized to account for random guessing.
     \end{minipage}
    \end{figure}

\paragraph{Decomposing improvements at the frontier.}
 Figure \ref{base_meek_models_main}, Panel (a) uses our regression to decompose benchmark performance for the best-performing model in 2022q4-2023q3 (Llama 70B) and 2024q3-2025q1 (GPT-4.5). Performance levels at the frontier are mostly explained by high compute levels. Over time, frontier performance is becoming \emph{even more} compute driven: The share of performance explained by scaling rose from 77\% to 88\%. Consistent with that, the top models' training compute rose by nearly a factor of 5,000, far exceeding estimated compute factors of shared algorithmic progress and company secret sauce in Figure \ref{fig: mainresults_main}.
The contributions of shared algorithmic progress and the secret sauce in LLM development to advancing frontier LLM capabilities have therefore been modest. And model effects imply even a negative impact. That is, GPT-4.5 is relatively less compute-efficient within OpenAI’s model lineup than Llama 2 70B is within Meta’s model lineup (conditional on publication time effects). In Appendix Table \ref{tab:model_comparison}, Column (1) we use a counterfactual decomposition analysis to show how the changes in Figure \ref{base_meek_models_main}, Panel (a) translate into actual performance changes and find that scaling alone can account for the \emph{entire} performance increase at the frontier. 

\paragraph{How technical progress gives rise to small efficient models.}\label{smaller_model_results}

 Figure \ref{base_meek_models_main}, Panel (b) runs the same decomposition for the smallest models that can achieve a given performance level. We focus on our main developers (as we estimated individual company effects for them) and examine the 15\% (normalized) MMLU-Pro threshold, which was already passed in January 2023 (and thus provides a long time series for tracking improvements). 
 For the 15\% threshold, required log FLOPs declined from 23.1 to 21.4, a 50-fold reduction (in unreported results, we find an 8,000-fold compute reduction when we also include small other developers). Accordingly, the importance of scaling in explaining benchmark performance declined by 26.6\%. Despite the compute reduction,  model performance slightly \emph{increased} (from 16.6\% to 18.0\%) due to gains from company secret sauce and shared algorithmic progress that both became much more important in explaining benchmark performance at the given threshold.\footnote{Appendix Table \ref{tab:model_comparison} reports the counterfactual performance contributions.}

\paragraph{Specialized Capabilities: MATH Level 5}\label{Disc_Math}

While our main analysis focuses on MMLU-Pro as a broad measure of LLM capabilities, Appendix \ref{Sec:Math_results} replicates all analyses for the MATH Level 5 benchmark. Results are qualitatively similar. Model compute remains the dominant driver of frontier performance despite its \emph{relative} contribution share to benchmark performance slightly declined from 75\% to 73\%. Interestingly, secret sauce plays a somewhat larger role in math-specific capability growth. Additionally, implied compute gains from shared algorithmic efficiency increase from 7.5-fold to 11-fold, and the 90–10 percentile gap in model-specific compute factors rises to 186×. These patterns suggest that some developers may have targeted math capabilities, resulting in larger technological gains and greater variation in compute efficiency within this domain.

\section{Discussion}\label{Discussion}

Our findings have important implications for AI progress, competition, and governance. Advances at the frontier of large language models are driven primarily by increases in training compute, with only modest contributions from shared algorithmic progress or developer-specific technologies. As a result, sustained leadership in frontier AI capabilities appears unlikely without continued access to rapidly expanding compute resources. This implies that access to compute is central for AI leadership and helps explain the ongoing race to invest in compute infrastructure. However, sustaining historical rates of compute growth seems at least uncertain. In that case, our findings suggest that future frontier progress may significantly \emph{slow down} as it needs to increasingly rely on algorithmic and technical improvements.

Despite the central role of compute at the frontier (with a compute growth factor of close to 5,000x), algorithmic efficiency is far from irrelevant. Over time, shared algorithmic improvements increased compute efficiency by a factor of 7.5x, while developer-specific technologies generate substantial dispersion in compute efficiency, with differences of up to a factor of 61x in the compute required to reach a given capability. Significant performance variation also persists across models within the same developer, consistent with ongoing experimentation, domain specialization, and a general uncertainty of training outcomes, which could indicate significant remaining scope for efficiency gains from improvements in model design.

The largest effects of technical progress arise below the frontier. Over the sample period, the compute required to reach modest capability thresholds declined by factors of up to 8,000x, reflecting a combination of shared algorithmic advances, developer-specific technologies, and model-specific innovations. Thus, the secret sauce of LLM development is less about sustaining a large performance lead at the very top and more about compressing capabilities into smaller, cheaper models. These efficiency gains lower user prices for given below-frontier AI capabilities and help democratize existing capabilities through lower-cost models, which, however, can also increase risk of AI misuse as access costs are also lowered for malicious actors (but not necessarily to the most advanced technologies).  

Finally, our results suggest that algorithmic efficiency gains contain a meaningful proprietary component, implying that technology diffusion, despite notable, may be slower than often assumed. This creates scope for firm-specific rents that shape competitive dynamics in AI markets. 

Our analysis focuses on non-reasoning models (see discussion in Section \ref{data}) and a limited set of benchmarks. Extensions to reasoning systems, task-specific and labor-market-relevant evaluations, and studying the economic implications of developer-specific efficiency advantages represent important directions for future research.

\section{Methods}\label{data}

\paragraph{Data}
We describe the data in further detail in Appendix \ref{detailed_data_methods}. Our main source is the Hugging Face Open LLM Leaderboard, which operated from March 2022 to March 2025 and reports harmonized benchmark scores, parameter counts, and publication dates for open-weight models. We augment these data with information on proprietary models from Epoch AI and TIGER Labs. Because these sources provide only parameter counts and benchmark scores, we hand-collect training-token counts from developer pages and model release papers. One challenge is the limited availability of data on parameter counts and training tokens for large proprietary models. To our knowledge, the only available source for these models consists of \emph{estimates} from lifearchitect.ai. We recognize that these figures are subject to substantial uncertainty. However, since they allow us to include important frontier models that would otherwise be excluded, we view their use as a worthwhile trade-off. To address concerns about relying on these estimates, we replicate our analysis excluding all observations sourced from lifearchitect.ai in Appendix \ref{app_no_life}. Reassuringly, our qualitative patterns are confirmed throughout. In particular, scaling remains the dominant source of frontier performance growth (despite omitting large proprietary models).

We focused on MMLU-Pro because it offers the most comparable cross-source performance measure (we studied MATH Level 5 in a robustness check). We restrict our analysis to non-reasoning models, as our data do not capture compute used during reasoning time, which would underestimate compute for reasoning models and bias comparisons.\footnote{We exclude 10 reasoning models (including GPT-o1) from our data. Including them leaves the Shapley decomposition unchanged and makes frontier progress less scale-dependent, though scaling remains the most dominant driver.} Properly extending the analysis to reasoning models will be feasible once inference-compute data become available.  We begin our sample in Q4 2022, when model release frequency became sufficient for our analysis. 

\paragraph{Regression framework.}
Our regression starts from the established power-law relationship (\cite{kaplan_scaling}), between a model \emph{i}'s training compute, $c_{i}$, and its error, $e_{i}$, that can be modeled in logs: $\log(e_{i})  = b \log(c_{i}) + a$, where $b$ and $a$ are parameters. This relationship is based on compute scaling laws. For instance, \textcite{hoffmann2022trainingcomputeoptimallargelanguage} describe predicted Loss, $L$, as a function of parameters ($N$) and data tokens ($D$): $L=F+\gamma^NN^{-\theta^{N}}+\gamma^DD^{-\theta^{D}}$, where $F$ is a residual error. \textcite{ruan2024observational} generalized this relationship to downstream performance with an inverse sigmoidal link function: 
\begin{equation}\label{EQ_invers_sigma}
\sigma^{-1}(y_{i})  = b \log(c_{i}) + a.
\end{equation}
 $0 < y_{i} < 1$ measures downstream performance (we use normalized MMLU-Pro scores corrected for random guessing). Based on this relationship (and because $\sigma^{-1}(y_i)=\text{logit}(y_i)$), we estimate the regression Eq. \eqref{main_reg}, where proprietary and shared algorithmic/technical progress is captured by developer and publication time dummies (equivalent to $a$ in Eq. \eqref{EQ_invers_sigma}). Appendix Figure \ref{fig: raw_data_scatter} shows that the logit-transformation we use fits the raw data well, consistent with \textcite{ruan2024observational}. 

\paragraph{Implied compute factors.}
We can rewrite time, developer, and model effects in Eq. \eqref{main_reg} in terms of computer factors. We illustrate this derivation for time effects (the derivation is analogous for developer and model effects): Let $M$ denote the compute factor we report throughout the paper, such that for any period effect, $\delta_t$, we can rewrite our estimated regression (left-hand side) as:
\begin{equation} \label{compute_factor_eq}
\beta_{0} + \beta_{c} \log(c_i) + \delta_t +\nu_i + \varepsilon_i = \beta_{0} + \beta_{c} \log(Mc_i)  + \nu_i + \varepsilon_i \quad \Leftrightarrow \quad M=10^{\left(\frac{\delta_t}{\beta_{c} }\right)}.
\end{equation}

\paragraph{Robustness.}
We discuss robustness tests for our empirical specification in Appendix \ref{robustness}. Appendix \ref{flex_fits} shows that our results are robust to more flexible logit transformations. In Appendix \ref{Sec:alt_time}, we use more detailed period fixed effects (quarterly fixed effects---with the drawback that some quarters contain insufficient model publication counts) and, instead of time dummies, linear and quadratic trends in publication months. These specifications address concerns about misattributing performance variation to company effects due to company-specific publication timelines. Reassuringly, under these alternative specifications, the variance decomposition yields very similar results and the estimated company secret sauce remains similar. In a third robustness test (Appendix \ref{time_interac}), we study if compute coefficients ($\beta_c$) change over time and find no statistically significant differences in time-specific compute coefficients.\footnote{We cannot conduct a similar analysis for company effects due to insufficient observation numbers, but note that model-specific factors ($\epsilon_i$) capture changing developer effects.}

\section{Data Availability}
We publish our novel dataset together with all program files required to produce our results under this \href{https://drive.google.com/file/d/1Mqy0_n-4ZC4J6NQlL7T1jiDQVHfEkESs/view?usp=sharing}{\textcolor{blue}{link}}.

\section{Funding Declaration}
Funding: Not applicable.

\section{Acknowledgments}
We thank Aaron Kaye, Andrew Lohn, Alexander Fogelson, Emanuele Del Sozzo, Hans Gundlach, Haoran Lyu, Jayson Lynch, Jonathan Rosenfeld, Omeed Maghzian, Martin Fleming, Zachary Brown, as well as other seminar participants at MIT FutureTech for insightful discussions and comments. We thank Haoran Lyu for sharing data and Anthony Meng for his excellent work as a research assistant.

\singlespacing


\printbibliography[segment=0]

\clearpage


\onehalfspacing
\newpage
\appendix
 \newrefsegment 
 \counterwithin{equation}{section}
\renewcommand\theequation{\thesection\arabic{equation}}
\renewcommand\thefigure{\thesection\arabic{figure}}
\renewcommand\thetable{\thesection\arabic{table}}

\setcounter{figure}{0} \renewcommand{\thefigure}{A.\arabic{figure}}
\setcounter{table}{0} \renewcommand{\thetable}{A.\arabic{table}}
\setcounter{equation}{0} \renewcommand{\theequation}{A.\arabic{equation}}

\singlespacing
\addtocontents{lof}{\protect\setcounter{tocdepth}{1}}
\vspace{-.2cm}
\addtocontents{lot}{\protect\setcounter{tocdepth}{1}}
\etocsettagdepth{mtchapter}{none}
\etocsettagdepth{mtappendix}{subsection}

\begin{center}
\vspace{-2cm}
\textbf{\Large{}Online Appendix of:} \\
\textbf{\Large{Is there “Secret Sauce” in Large Language Model
Development?}}{\Large\par}
 \smallskip
\textbf{Matthias Mertens, Natalia Fischl-Lanzoni, and Neil Thompson}{\par}
\par\end{center}
\vspace{-.5cm}

    \setcounter{figure}{0} \renewcommand{\thefigure}{A.\arabic{figure}}
    \setcounter{table}{0} \renewcommand{\thetable}{A.\arabic{table}}
    \setcounter{equation}{0} \renewcommand{\theequation}{A.\arabic{equation}}

    \singlespacing
\addtocontents{lof}{\protect\setcounter{tocdepth}{1}}
\vspace{-.2cm}
\addtocontents{lot}{\protect\setcounter{tocdepth}{1}}
\etocsettagdepth{mtchapter}{none}
\etocsettagdepth{mtappendix}{subsection}


\vspace{-.2cm}



\etocdepthtag.toc{mtappendix}

\section{Detailed Data Collection Process} \label{detailed_data_methods}

We download data from the HuggingFace Open LLM Leaderboard, which operated from March 2022 to March 2025 and reports publication (upload) dates, parameter counts, and harmonized evaluation results for  MMLU-Pro, BBH, GPQA, MATH Level 5, and MUSR benchmark scores for more than 4,000 models.

\subsection{Other Data Sources}
The Hugging Face Open LLM Leaderboard has three drawbacks  i) it does not report benchmark scores for proprietary models; ii) it does not report results for several of the largest open-weight models; and iii) it does not report training or fine-tuning dataset sizes. To address these gaps, we supplement the data with four additional sources: EpochAI, TIGER Lab, Life Architect AI, and hand-collected information from publications. We now describe the information obtained from each of these sources.

\paragraph{TIGER Labs.} 
TIGER Labs, the creator of the MMLU-Pro benchmark, maintains its own leaderboard that reports scores for both proprietary and open-weight models. These scores are either produced under the same benchmark specifications used by Hugging Face (5-shot prompting) or are self-reported by model developers. Although self-reported results are not ideal, the added coverage of large open-weight and proprietary models is highly valuable. We therefore append 87 additional models from TIGER Labs to our dataset.

\paragraph{Epoch AI.}
Epoch AI reports MATH Level 5 benchmark scores for a wide range of models, complementing the scores available on the Hugging Face Leaderboard and TIGER Labs. We use the Epoch AI Benchmarking Hub to obtain results for proprietary and larger open-weight models that are missing from the Hugging Face data, analogous to how TIGER Labs supplements MMLU-Pro coverage. A limitation is that Epoch AI reports 0-shot performance, whereas the Hugging Face Leaderboard uses 4-shot prompting. However, because we use the MATH Level 5 benchmark only as a robustness check, this discrepancy is not central to our analysis. In total, we added benchmark data for 11 additional models from Epoch AI.
 
\paragraph{Life Architect AI.}
To further supplement our dataset, we also use the LifeArchitect.AI models table. Although the parameter and dataset token counts are estimates, we believe that they provide a general approximation of the FLOPs used to train the model. We add 16 additional models with dataset size and parameter counts from LifeArchitect.AI.

\paragraph{Hand-collected data from publications.}
To complement our data with dataset token counts, we first examined Hugging Face model cards to identify cases where creators reported this information. For models not hosted on Hugging Face, or when token counts were missing, we reviewed the corresponding model release papers. For example, the token count for Qwen-2.5 was obtained directly from its release paper. When possible, we recorded both pre-training and fine-tuning token counts, though most models do not disclose this level of detail. In practice, we define dataset size as the sum of pre-training and fine-tuning tokens; when fine-tuning tokens are unavailable, we use the pre-training token count alone. Because pre-training determines model weights, pre-training tokens are the most important measure for our purposes.\footnote{For the subsample of models with both pre-training and fine-tuning token information, regression results were insensitive to whether we use only base-model pre-training tokens or the sum of pre-training and fine-tuning tokens, as these values typically differ only modestly.} We also cross-checked all entries against third-party sources reporting dataset sizes, but because these often rely on speculative estimates, every model was manually verified following the procedure above.

Across multiple sources---LifeArchitect.AI, Epoch AI, and model publications---we obtain total training dataset sizes for 291 models. We also identify dataset sizes for 211 base models. If a model is listed as a base model on Hugging Face, its total dataset size corresponds to its base-model token count; otherwise, we assign the token count used to train its underlying base model. These 211 base models appear in 770 models on the Hugging Face Leaderboard.

\paragraph{Choice of main developers.}
In our analysis, we estimate developer-specific effects for a set of “main developers.” This set is constructed in two steps. First, we include major companies for which we observe more than three models in our data and that developed at least one model achieving an MMLU-pro score above 30\% after adjustment for random guessing (Microsoft, Nvidia, Qwen, DeepSeek, Meta, Google, and 01-AI). Second, we include developers that released well-known, high-performing models during our observation period, even if they do not meet the first criterion (OpenAI, xAI, and Anthropic).

Our results are robust to alternative definitions of the main-developer set. For example, we obtain similar results when including IBM, Mistral, and/or Tencent.

\newpage
\section{Contributions of Scaling, Secret Sauce, Shared Algorithmic Progress, and Model Effects to Benchmark Score Changes}
 \counterwithin{equation}{section}
\renewcommand\theequation{\thesection\arabic{equation}}
\renewcommand\thefigure{\thesection\arabic{figure}}
\renewcommand\thetable{\thesection\arabic{table}}

\setcounter{figure}{0} \renewcommand{\thefigure}{B.\arabic{figure}}
\setcounter{table}{0} \renewcommand{\thetable}{B.\arabic{table}}
\setcounter{equation}{0} \renewcommand{\theequation}{B.\arabic{equation}}

Table \ref{tab:model_comparison} reports the implied contributions of the regression components underlying the benchmark score changes shown in Figure \ref{base_meek_models_main}. We express all changes in log-odds to respect the non-linear relationship between benchmark performance and compute implied by the logit specification. Performing the decomposition in actual benchmark score space would linearize this relationship and therefore bias the attribution. Each entry in the table reflects the contribution of a single component (scaling, shared progress, firm-specific secret sauce, and model effects) to the change in the logit benchmark score between the first and last models, holding all other components fixed.

At the frontier (Column (1)), the increase in benchmark performance from LLaMA-2-70B to GPT-4.5 corresponds to a 54.4-percentage-point gain in MMLU scores, equivalent to a 2.5-point increase in log-odds. Column (1) shows that scaling alone accounts for more than the full observed increase, indicating that model size is the dominant driver of performance improvements at the frontier.

Column (2) presents the decomposition corresponding to Figure \ref{base_meek_models_main}, Panel (b), where performance is held approximately constant and we track the smallest model capable of achieving a given benchmark score. Over this comparison, benchmark performance changes only marginally (from 16.6\% to 18.0\%), while model size declines substantially (from $10^{23.1}$ to $10^{21.4}$ FLOPs). Column (2) shows that firm-specific factors (secret sauce) and shared algorithmic progress contribute roughly equally to this improvement in compute efficiency, jointly enabling large reductions in model size at nearly constant performance levels.

\begin{table}[H]
\caption{Counterfactual Contributions to MMLU-Pro Benchmark Scores in Log-Odds} \label{tab:model_comparison}
\begin{adjustbox}{width=1\linewidth}

\centering
\begin{tabular}{lcc}

\hline\hline
& \textbf{Top Performance Growth} & \textbf{Smallest Models Reaching 15\% Benchmark Score} \\
& \makecell[c]{From Llama2 70B (Score: 30.2\%)\\ to GPT 4.5 (Score: 84.6\%)} & \makecell[c] {From Google Flan UL2 (Score: 16.6\%)\\ to Nvidia Minitron 4B (Score: 18.0\%)} \\
& (1) & (2) \\
\hline
$\Delta$ Scaling in log-odds & +2.9   & -1.4  \\
$\Delta$ Shared progress in log-odds & +0.7  & +0.7  \\
$\Delta$ Company secret sauce in log-odds & +0.3  & +0.8  \\
$\Delta$ Model effects in log-odds & -1.3  & -0.0  \\
\hline
Total benchmark score change in log-odds  & + 2.5  & +0.1  \\
\hline\hline
\end{tabular}
\end{adjustbox}
\vspace{-5mm}
\begin{minipage}{\textwidth}\scriptsize\singlespacing \textit{Notes:} The table shows the contributions to change of benchmark scores (in log-odds) for top performing models and more efficient models shown in Figure \ref{base_meek_models_main}. The numbers in the table  show how much each component (scaling, shared progress, company secret sauce, and model effects) changed the logit benchmark score from the first model to the last model, holding fixed all other factors. We compute these numbers by calculating the share of changes in logit benchmark scores associated with each regression component. For the top performing models, the change in score is calculated with Llama2 70B ($10^{23.9}$ Flops) and GPT 4.5 ($10^{27.6}$ Flops). For the more efficient models, the change in benchmark score is calculated with Google Flan UL2 ($10^{23.1}$ Flops) and Nvidia Minitron 4B  ($10^{21.4}$ Flops).
\end{minipage}
\end{table}

\newpage
\section{Additional Tables and Figures}
 \counterwithin{equation}{section}
\renewcommand\theequation{\thesection\arabic{equation}}
\renewcommand\thefigure{\thesection\arabic{figure}}
\renewcommand\thetable{\thesection\arabic{table}}

\setcounter{figure}{0} \renewcommand{\thefigure}{C.\arabic{figure}}
\setcounter{table}{0} \renewcommand{\thetable}{C.\arabic{table}}
\setcounter{equation}{0} \renewcommand{\theequation}{C.\arabic{equation}}

\begin{figure} [H]
    \caption{MMLU-Pro Score vs Log(FLOPs) Data Visualization with a Logistic Curve Fit}
  \label{fig: raw_data_scatter}
    \centering
    \includegraphics[width=0.8\textwidth]{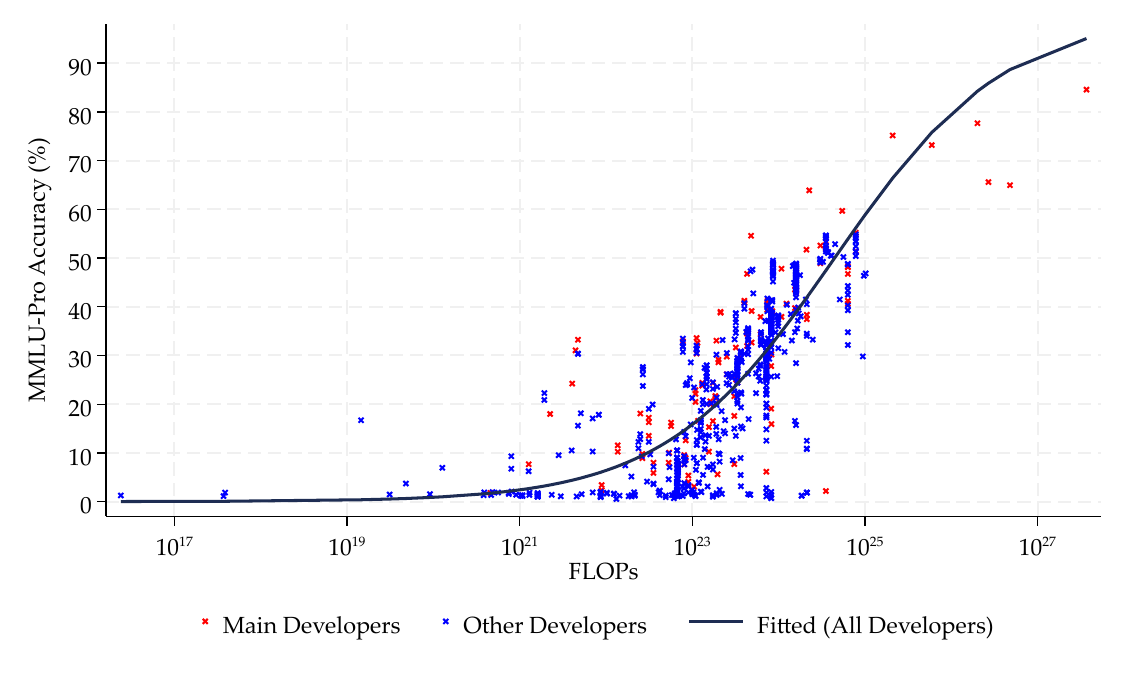}
   \begin{minipage}{\textwidth}
        \scriptsize \textit{Notes:} The figure show the raw data of MMLU-Pro benchmark scores (vertical axis) against log FLOPs (horizontal axis). We fit the following two parameter logistic curve through the data: $\text{Benchmark score} = \frac{1}{1 + exp(-(.4400164 \log(FLOPs) + (-6.736996)))}$, where the parameters where estimated using non-linear least squares.
     \end{minipage}
\end{figure}

\begin{table}[H]
\centering
\caption{Main Regression Specification}
\label{tab:reg}
\begin{adjustbox}{width=.95\linewidth}
\def\sym#1{\ifmmode^{#1}\else\(^{#1}\)\fi}
\renewcommand{\arraystretch}{1.2}
\begin{tabular}{l*{4}{c}}
\hline\hline
                & \multicolumn{4}{c}{\emph{Dependent var: Logit(MMLU-Pro scores)}} \\ \cmidrule(lr){2-5}
                & Full sample & Major developers & \shortstack{Major developers,\\below median FLOPs} & \shortstack{Major developers,\\above median FLOPs} \\
                  & (1) & (2) & (3) & (4) \\
\hline
Log(FLOPs)      & 0.789\sym{***} & 0.885\sym{***} & 1.096\sym{***} & 0.467\sym{***} \\
                & (0.192) & (0.142) & (0.240) & (0.116) \\

2023q4--2024q2  & 0.702\sym{***} & 0.152 & -0.240 & 0.748\sym{***} \\
                & (0.182) & (0.203) & (0.317) & (0.148) \\

2024q3--2025q1  & 0.692\sym{***} & 0.307 & 0.257 & 0.799\sym{**} \\
                & (0.128) & (0.230) & (0.468) & (0.271) \\

Deepseek        & 0.282 & -1.160\sym{***} & -1.880\sym{**} & 2.434\sym{***} \\
                & (0.224) & (0.327) & (0.647) & (0.201) \\

Qwen            & 0.667\sym{***} & -0.825\sym{**} & 1.432\sym{**} & 2.080\sym{***} \\
                & (0.198) & (0.166) & (0.168) & (0.078) \\

Meta            & 0.535\sym{*} & -1.145\sym{***} & -2.163\sym{***} & 2.028\sym{***} \\
                & (0.278) & (0.331) & (0.441) & (0.183) \\

Google          & 0.627\sym{***} & -0.829\sym{**} & -1.826\sym{**} & 2.447\sym{***} \\
                & (0.198) & (0.307) & (0.515) & (0.216) \\

Microsoft       & 1.407\sym{***} & -0.058 & -0.448 & 2.485\sym{***} \\
                & (0.086) & (0.215) & (0.467) & (0.101) \\

OpenAI          & 0.805 & -0.913 & --- & 2.726\sym{***} \\
                & (0.692) & (0.623) &  & (0.346) \\

Anthropic      & 0.740 & -0.964 & --- & 2.612\sym{***} \\
                & (0.663) & (0.602) &  & (0.329) \\

X-AI            & 0.890 & -0.925 & --- & 2.830\sym{***} \\
                & (0.745) & (0.602) &  & (0.284) \\

01-AI           & 1.071\sym{***} & -0.308 & -0.717 & 2.321\sym{***} \\
                & (0.159) & (0.314) & (0.671) & (0.209) \\

Nvidia          & 1.412\sym{***} & --- & --- & --- \\
                & (0.070) &  &  &   \\

Constant        & -21.10\sym{***} & -21.40\sym{***} & -25.49\sym{***} & -14.67\sym{***} \\
                & (4.216) & (2.925) & (4.814) & (2.516) \\

\hline
N               & 809 & 122 & 61 & 61 \\
R$^2$           & 0.526 & 0.637 & 0.683 & 0.614 \\
\hline\hline
\end{tabular}
\end{adjustbox}

\begin{minipage}{\textwidth}
\vspace{0.5em}
\scriptsize \textit{Notes:} The table reports OLS regressions of logit-transformed MMLU-Pro accuracy on log training compute (FLOPs), time fixed effects, and developer fixed effects. Baseline categories are the period 2022q4–2023q3 and the group of other small developers. Standard errors clustered at the developer level are reported in parentheses. Columns (2)-(4) focus on major developers. Columns (3) and (4) split the sample by median training compute. Significance levels: \sym{*} $p<0.10$, \sym{**} $p<0.05$, \sym{***} $p<0.01$.
\end{minipage}
\end{table}

\newpage
\section{Results Replicated Without LifeArchitect Proprietary Models}\label{app_no_life}

\begin{figure} [H]
  \label{fig: shapley nolife}
    \centering
    \caption{Shapley $R^2$ Decomposition: MMLU-Pro}
    \includegraphics[width=0.8\textwidth]{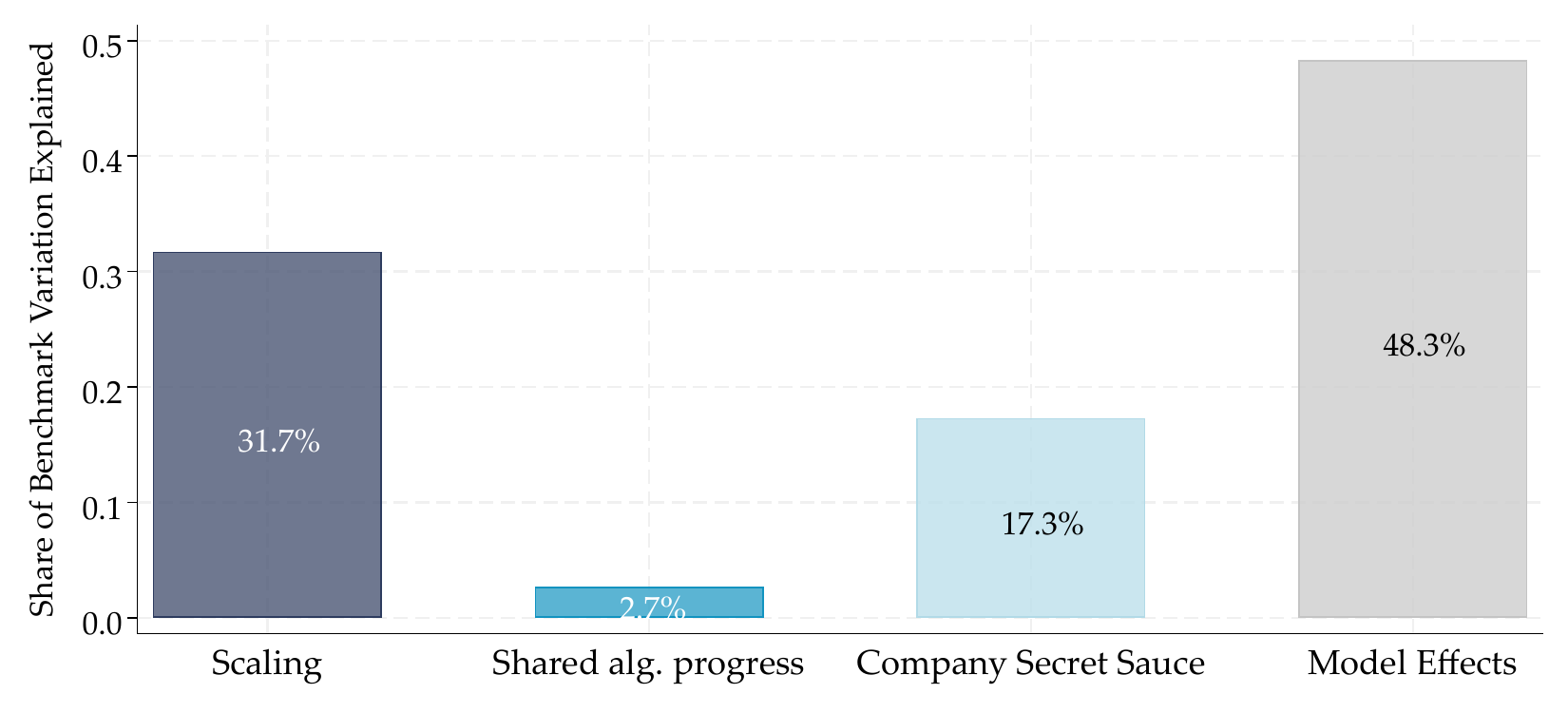}
    \begin{minipage}{\textwidth}
        \scriptsize \textit{Notes:} The figure reports a Shapley decomposition of the regression $R^2$ into contributions from scale, shared algorithmic progress,
and company factors for 793 LLMs with MMLU Pro benchmark scores using the command shapley2 in Stata. Model effects captures all variance in MMLU Pro benchmark score not explained by scaling, shared algorithmic progress, or company effects. 
    \end{minipage}
    \end{figure}

    \begin{figure}[H]
    \caption{Main Regression results: MMLU-Pro}
    \label{fig: mainresults_nolifearch}
    \centering
    \begin{subfigure}[t]{0.49\textwidth}
        \centering
            \caption{Scaling and Performance}
        \includegraphics[width=\textwidth, height=5cm, keepaspectratio]{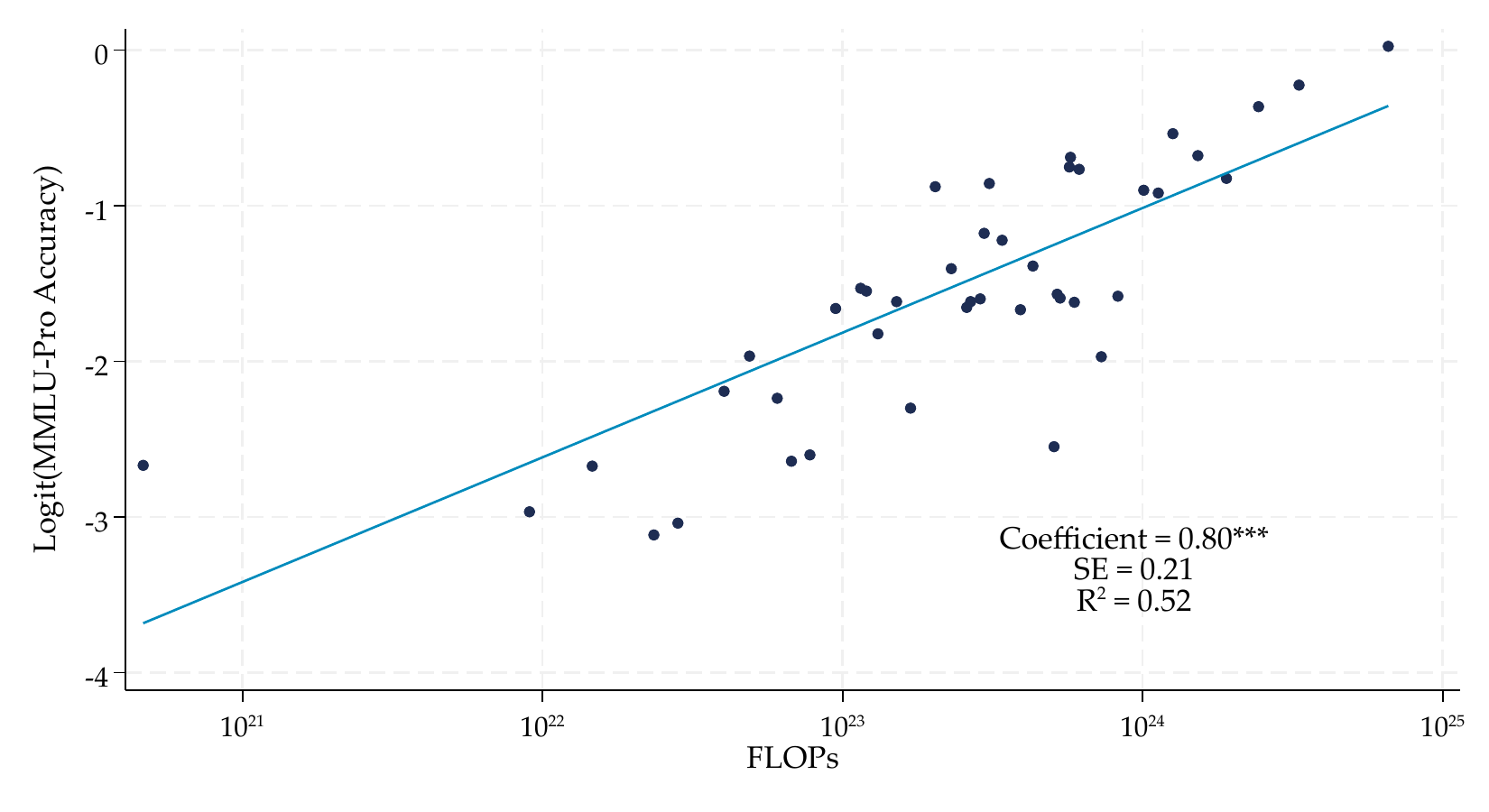}
        \vspace{-0.5em}
    
    \end{subfigure}
    \hfill
    \begin{subfigure}[t]{0.49\textwidth}
        \centering
          \caption{Shared Algorithmic Progress: Compute Factor Gain}
        \includegraphics[width=\textwidth, height=5cm, keepaspectratio]{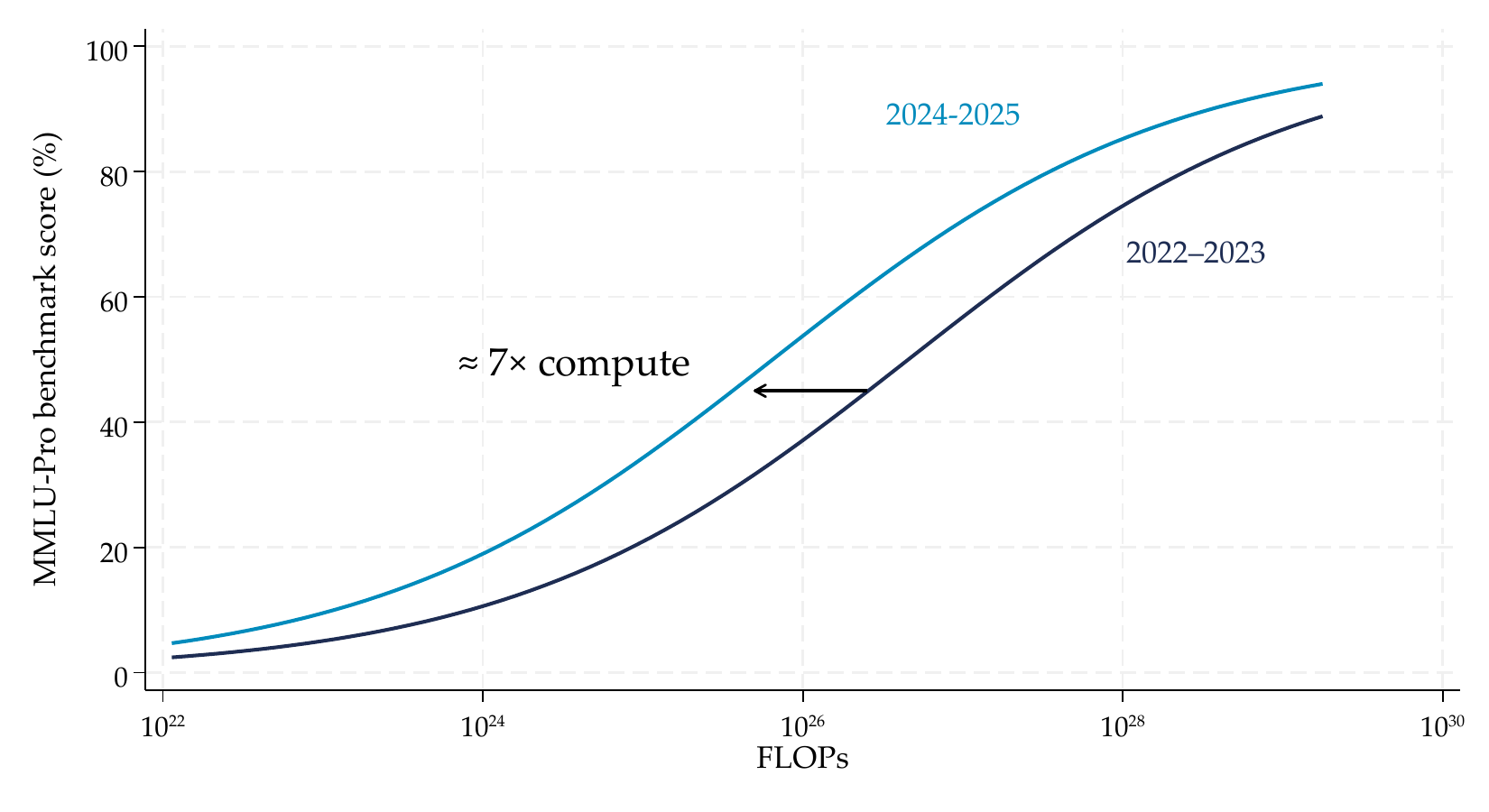}
        \vspace{-0.5em}
      
    \end{subfigure}
    
    \vspace{-0.3em} 

    \begin{subfigure}[t]{\textwidth}
        \centering
                \caption{Company and Model Effects in Compute Factors }
        \includegraphics[width=\textwidth, height=8cm, keepaspectratio]{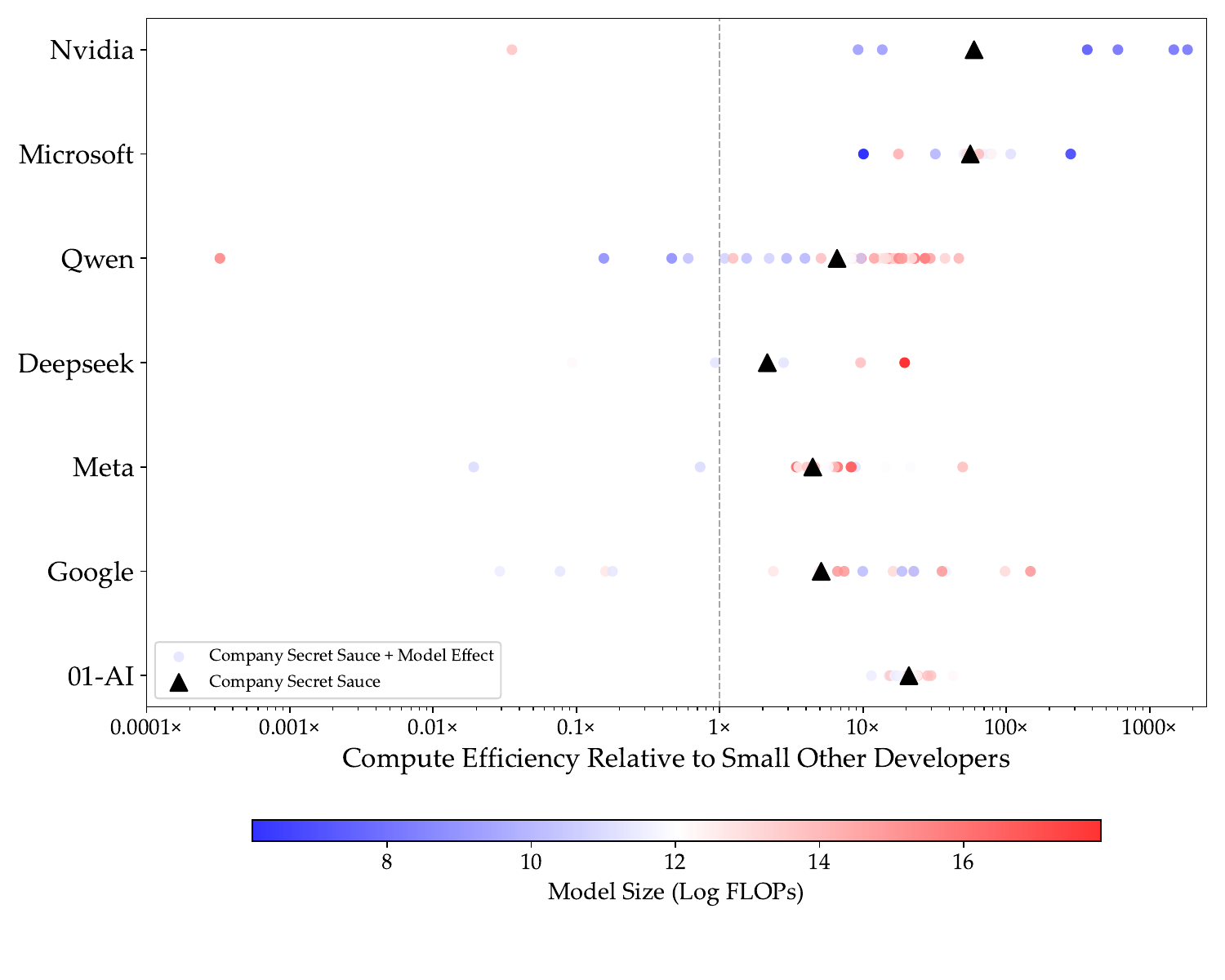}
        \vspace{-0.5em}

    \end{subfigure}
    
    \begin{subfigure}[t]{0.49\textwidth}
        \centering
              \caption{Company Effects and Relative Model Size}
        \includegraphics[width=\textwidth, height=5cm, keepaspectratio]{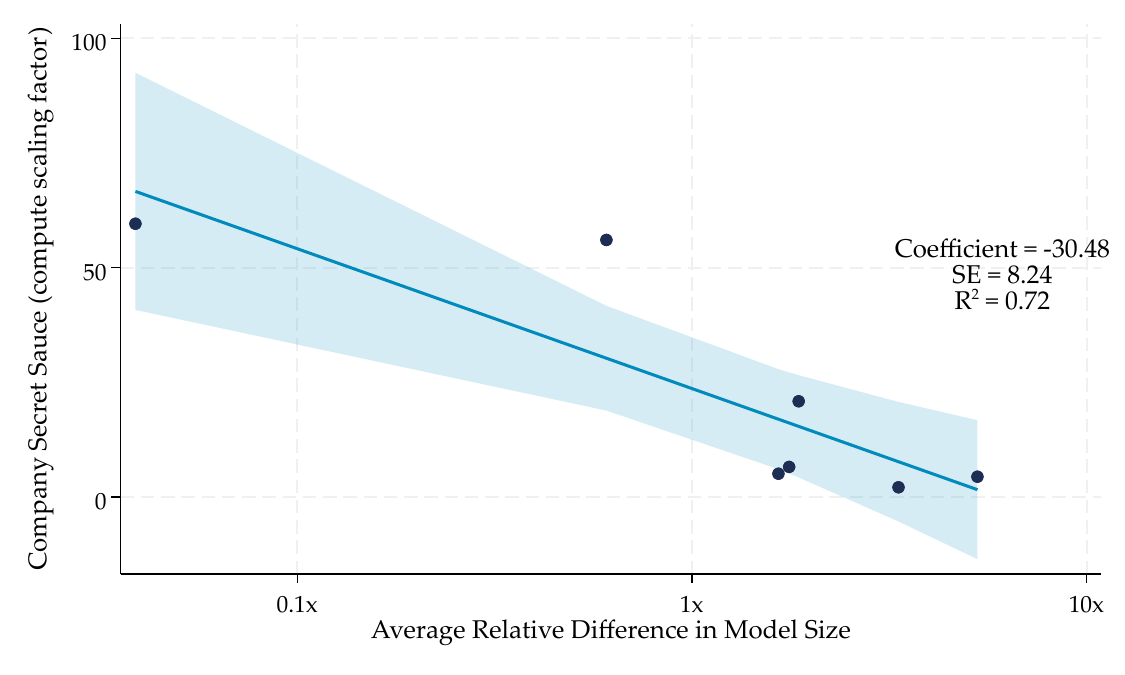}
        \vspace{-0.5em}
  
    \end{subfigure}
    \hfill
    \begin{subfigure}[t]{0.49\textwidth}
        \centering
                \caption{Model-Specific Effects in Compute Factors}
        \includegraphics[width=\textwidth, height=5cm, keepaspectratio]{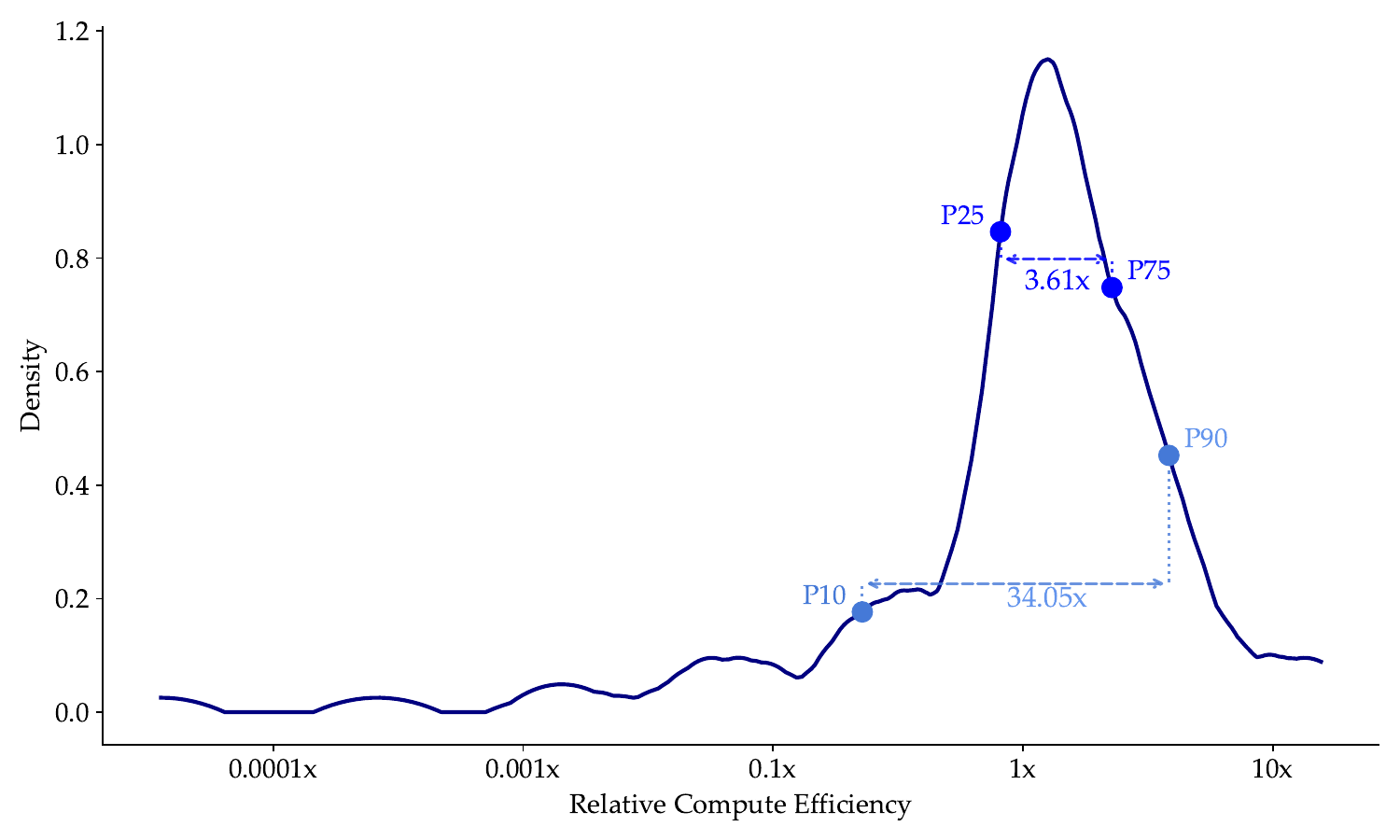}
        \vspace{-0.5em}

    \end{subfigure}
    
    \vspace{-0.3em} 

    \begin{minipage}{\textwidth}
        \scriptsize \singlespacing \textit{Notes:} The figure reports results for MMLU-Pro benchmark scores (793 LLMs). Panel (a) visualizes Eq. \eqref{main_reg} using a binned scatter plot that absorbs developer and period effects. Panel (b) shows the implied compute–performance curves for the first and last periods as predicted by the regression.  Panel (c) reports the estimated developer fixed effects (triangles). The shaded dots add model effects ($\varepsilon_i$) to the developer fixed effects with colors indicating the size of the models. Panel (d) uses a binned scatter plot to project company effects against the average of model size deviations from the average size of published models in a given period (log differences). Panel (e) displays the distribution of model effects/residuals ($\varepsilon_i$) for main developers. Significance levels in Panel (a) and (d): \sym{*} $p<0.10$, \sym{**} $p<0.05$, \sym{***} $p<0.01$
    \end{minipage}
\end{figure}

\begin{figure} [H]
    \caption{Contributions to Top Model Over Time}
  \label{top_models_main_nolifearch}
    \centering
    \begin{subfigure}[b]{0.89\textwidth}
    \caption{Top Scoring Models}
    \includegraphics[width=\textwidth]{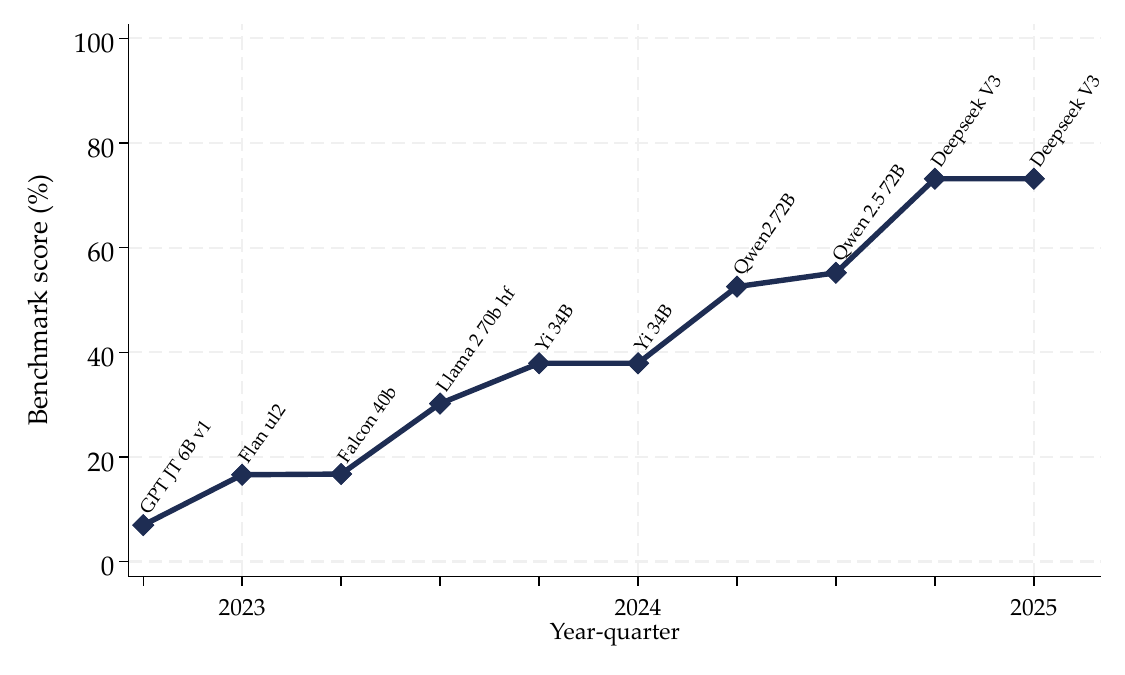}
    \end{subfigure}
    \begin{subfigure}[b]{0.89\textwidth}
    \caption{Required FLOPs for a Given MMLU-Pro Score}
    \includegraphics[width=\textwidth]{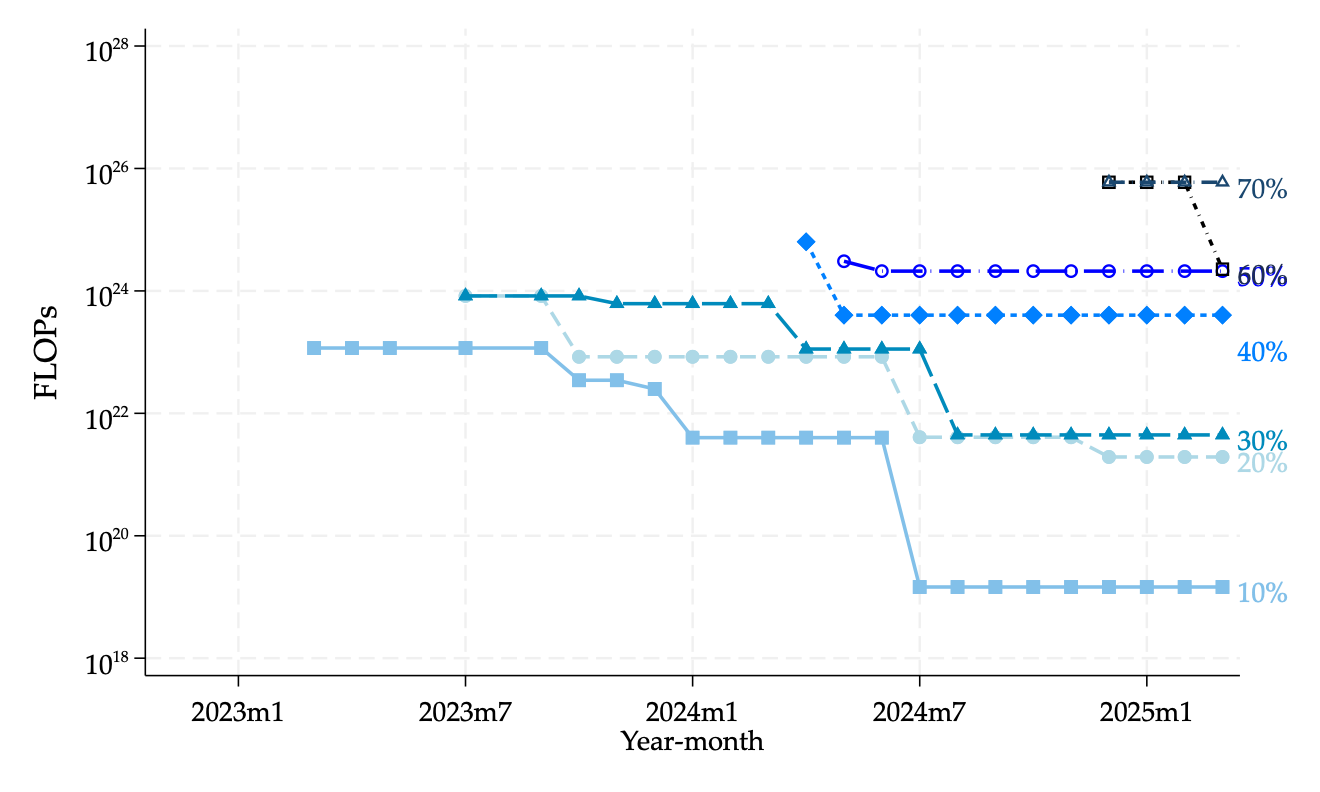}
    \end{subfigure}
                   \begin{minipage}{\textwidth}
        \scriptsize\singlespacing \textit{Notes:} Panel (a) shows the top scoring model on the MMLU-Pro benchmark each quarter. Panel (b) shows, over time, the smallest models (in FLOPs) in the data that reach a given MMLU-Pro performance level. 
     \end{minipage}

    \end{figure}

    \begin{figure} [H]
    \caption{Sources of Performance Growth: Frontier Models and Smaller, Efficient Models}
  \label{nolifearch_meek_models_main}
    \centering
    \begin{subfigure}[b]{0.95\textwidth}
    \caption{Source of Benchmark Score Growth: How Top Performing Models Become Better}
    \includegraphics[width=\textwidth]{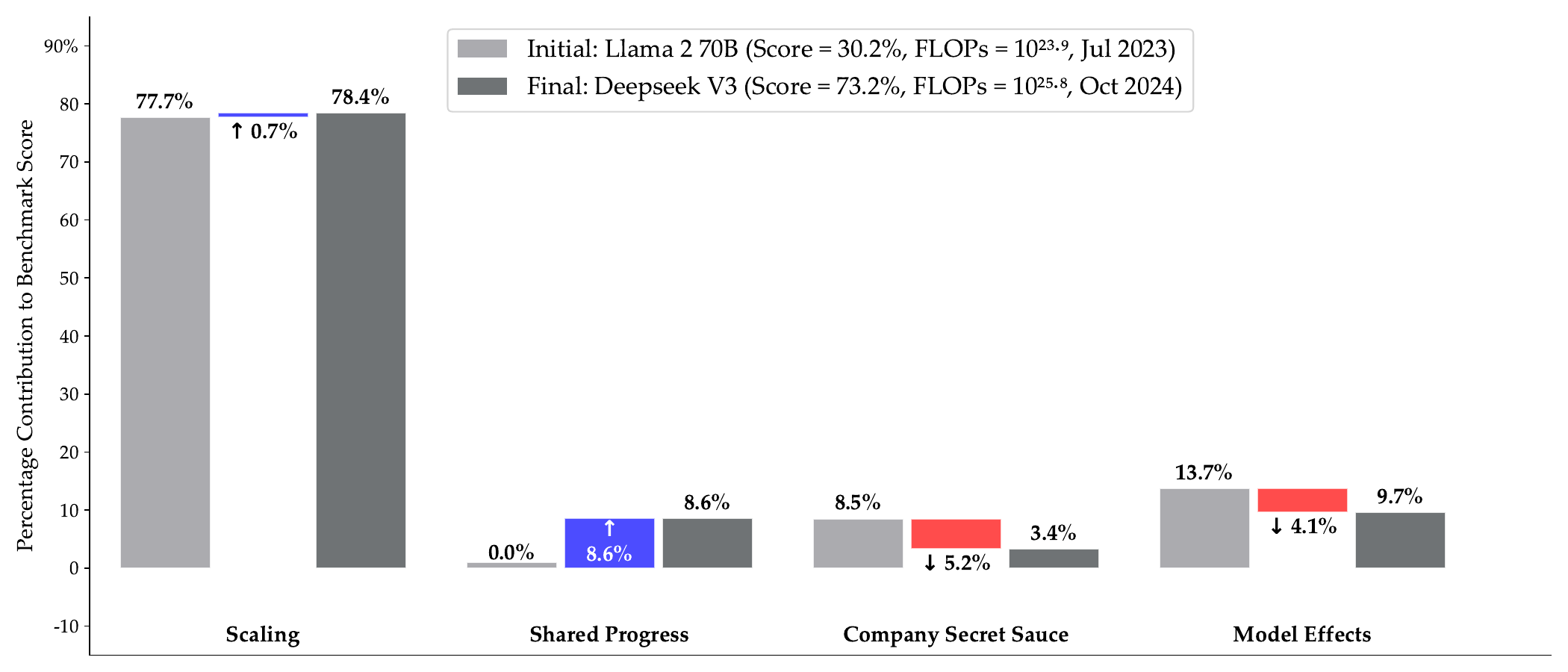}
    \end{subfigure}
    \begin{subfigure}[b]{0.95\textwidth}
    \caption{Source of Benchmark Score Growth: How Models Become More Efficient}
    \includegraphics[width=\textwidth]{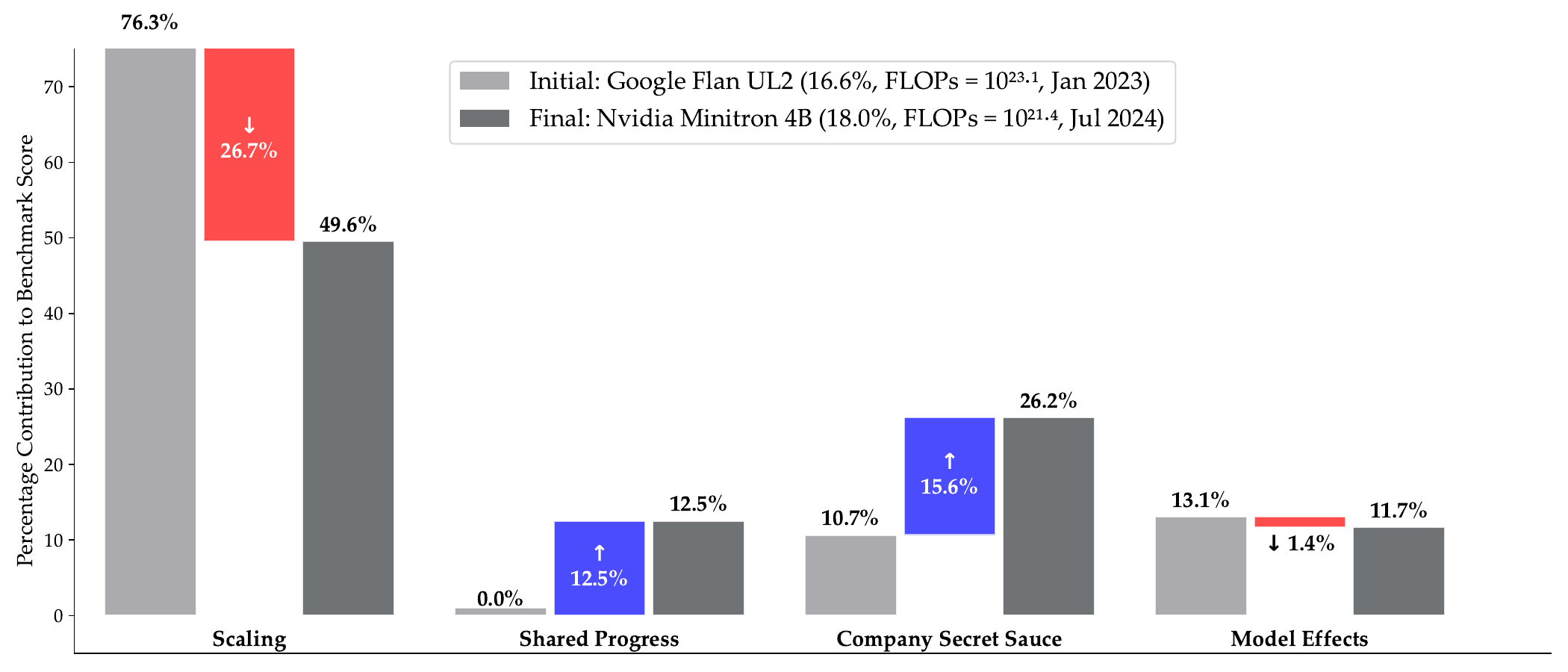}
    \end{subfigure}
          \begin{minipage}{\textwidth}
        \scriptsize\singlespacing \textit{Notes:} Panel (a) applies our regression-based decomposition to the highest-performing models for MMLU-Pro in the first and last time periods of our sample. Panel (b) applies our regression-based decomposition to the first model and the smallest models to achieve at least 15\% on MMLU-Pro. We compute contribution shares of scaling, shared progress, company effects, and model effects by summing the estimated components of Eq. \eqref{main_reg} and calculating each component’s share of the total. These shares are then applied to the benchmark scores.
     \end{minipage}
    \end{figure}

\begin{table}[H]
\caption{Counterfactual Contributions to MMLU-Pro Benchmark Scores in Log-Odds} \label{tab:model_comparison_nolifearch}
\begin{adjustbox}{width=1\linewidth}

\centering
\begin{tabular}{lcc}

\hline\hline
& \textbf{Top Performance Growth} & \textbf{Smallest Models Reaching 15\% Benchmark Score} \\
& \makecell[c]{From Llama2 70B (Score: 30.2\%)\\ to Deepseek V3 (Score: 73.2\%)} & \makecell[c] {From Google Flan UL2 (Score: 16.6\%)\\ to Nvidia Minitron 4B (Score: 18.0\%)} \\
& (1) & (2) \\
\hline
$\Delta$ Scaling in log-odds & +1.5   & -1.4  \\
$\Delta$ Shared progress in log-odds & +0.7  & +0.7  \\
$\Delta$ Company secret sauce in log-odds & -0.3 & +0.9  \\
$\Delta$ Model effects in log-odds & -0.7  & -0.1  \\
\hline
Total benchmark score change in log-odds  & + 2.5  & +0.1  \\
\hline\hline
\end{tabular}
\end{adjustbox}
\vspace{-5mm}
\begin{minipage}{\textwidth}\scriptsize\singlespacing \textit{Notes:} The table shows the contributions to change of benchmark scores (in log-odds) for top performing models and more efficient models shown in Figure \ref{base_meek_models_main}. The numbers in the table  show how much each component (scaling, shared progress, company secret sauce, and model effects) changed the logit benchmark score from the first model to the last model, holding fixed all other factors. We compute these numbers by calculating the share of changes in logit benchmark scores associated with each regression component. For the top performing models, the change in score is calculated with Llama2 70B ($10^{23.9}$ Flops) and Deepseek V3 ($10^{25.8}$ Flops). For the more efficient models, the change in benchmark score is calculated with Google Flan UL2 ($10^{23.1}$ Flops) and Nvidia Minitron 4B  ($10^{21.4}$ Flops).
\end{minipage}
\end{table}

\newpage
\section{Results Replicated With the MATH Level 5 Benchmark}\label{Sec:Math_results}
 \counterwithin{equation}{section}
\renewcommand\theequation{\thesection\arabic{equation}}
\renewcommand\thefigure{\thesection\arabic{figure}}
\renewcommand\thetable{\thesection\arabic{table}}

\setcounter{figure}{0} \renewcommand{\thefigure}{D.\arabic{figure}}
\setcounter{table}{0} \renewcommand{\thetable}{D.\arabic{table}}
\setcounter{equation}{0} \renewcommand{\theequation}{D.\arabic{equation}}

\begin{figure} [H]
    \caption{MATH Level 5 Score vs Log FLOPs Data Visualization with a Logistic Curve Fit}
  \label{fig: raw_data_scatter_math}
    \centering
    \includegraphics[width = 0.8\textwidth]{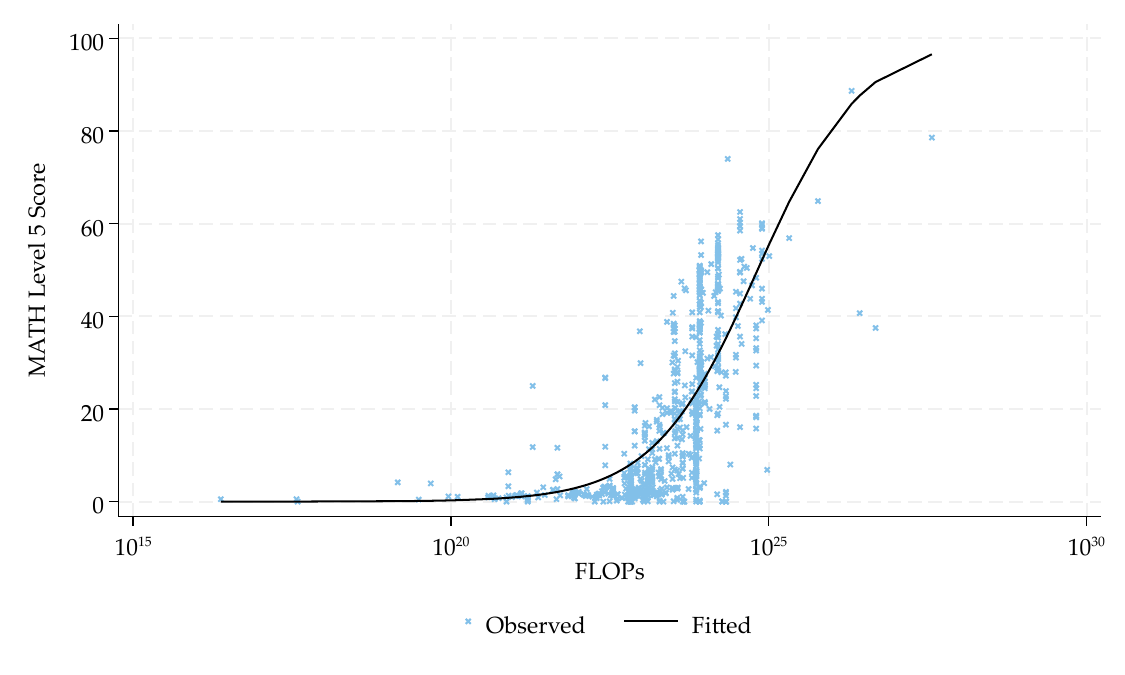}
   \begin{minipage}{\textwidth}
        \scriptsize \textit{Notes:}  The figure show the raw data of MATH Level 5 benchmark scores (vertical axis) against log FLOPs (horizontal axis). We fit the following two parameter logistic curve through the data: $\text{Benchmark score} =  \frac{1}{1 + exp(-(0.5305669   \ln(FLOPs) + (-8.300407)))}$, where the parameters where estimated using non-linear least squares.
     \end{minipage}
\end{figure}

\begin{figure} [H]
  \label{fig: min FLOPs to score}
    \centering
    \caption{Shapley $R^2$ Decomposition: MATH Level 5}
    \includegraphics[width=0.8\textwidth]{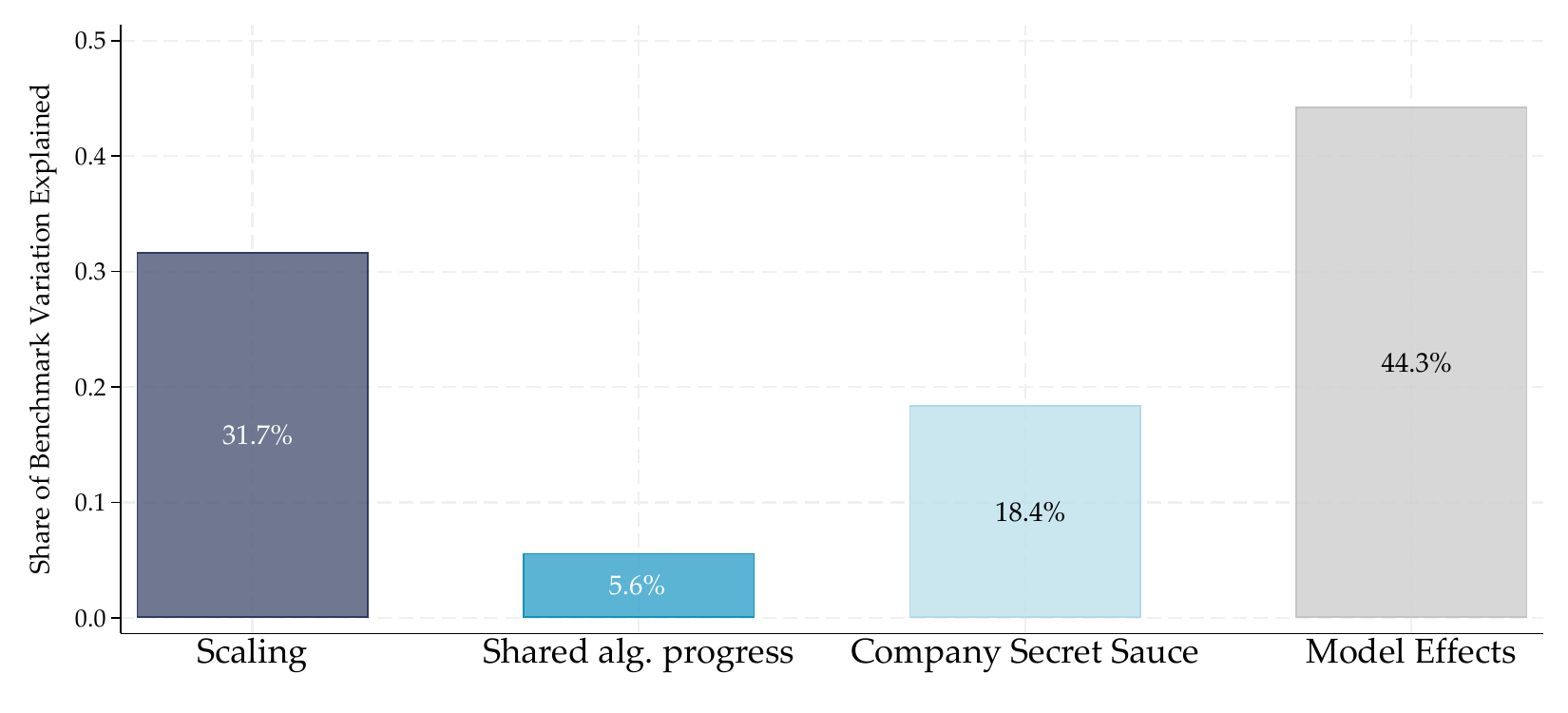}
    \begin{minipage}{\textwidth}
        \scriptsize \textit{Notes:} The figure reports a Shapley decomposition of the regression $R^2$ into contributions from scale, shared algorithmic progress,
and company factors for 785 LLMs with MATH Level 5 benchmark scores using the command shapley2 in Stata. Model effects captures all variance in MATH Level 5 benchmark score not explained by scaling, shared algorithmic progress, or company effects. 
    \end{minipage}
    \end{figure}

\begin{figure}[H]
    \caption{Main Regression results: MATH Level 5}
    \label{fig: mainresults}
    \centering
    \begin{subfigure}[t]{0.49\textwidth}
        \centering
            \caption{Scaling and Performance}
        \includegraphics[width=\textwidth, height=5cm, keepaspectratio]{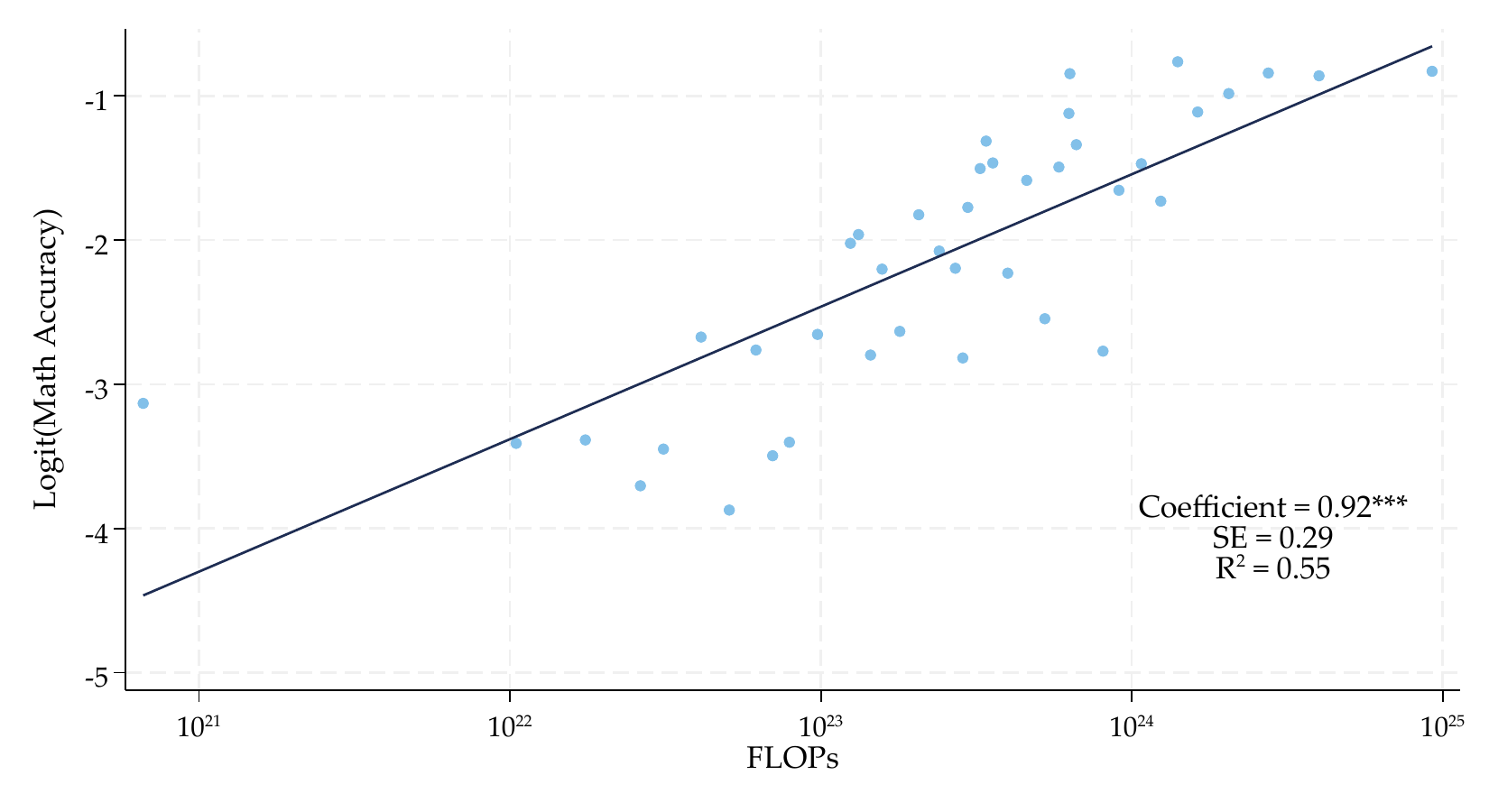}
        \vspace{-0.5em}
    
    \end{subfigure}
    \hfill
    \begin{subfigure}[t]{0.49\textwidth}
        \centering
          \caption{Shared Algorithmic Progress: Compute Factor Gain}
        \includegraphics[width=\textwidth, height=5cm, keepaspectratio]{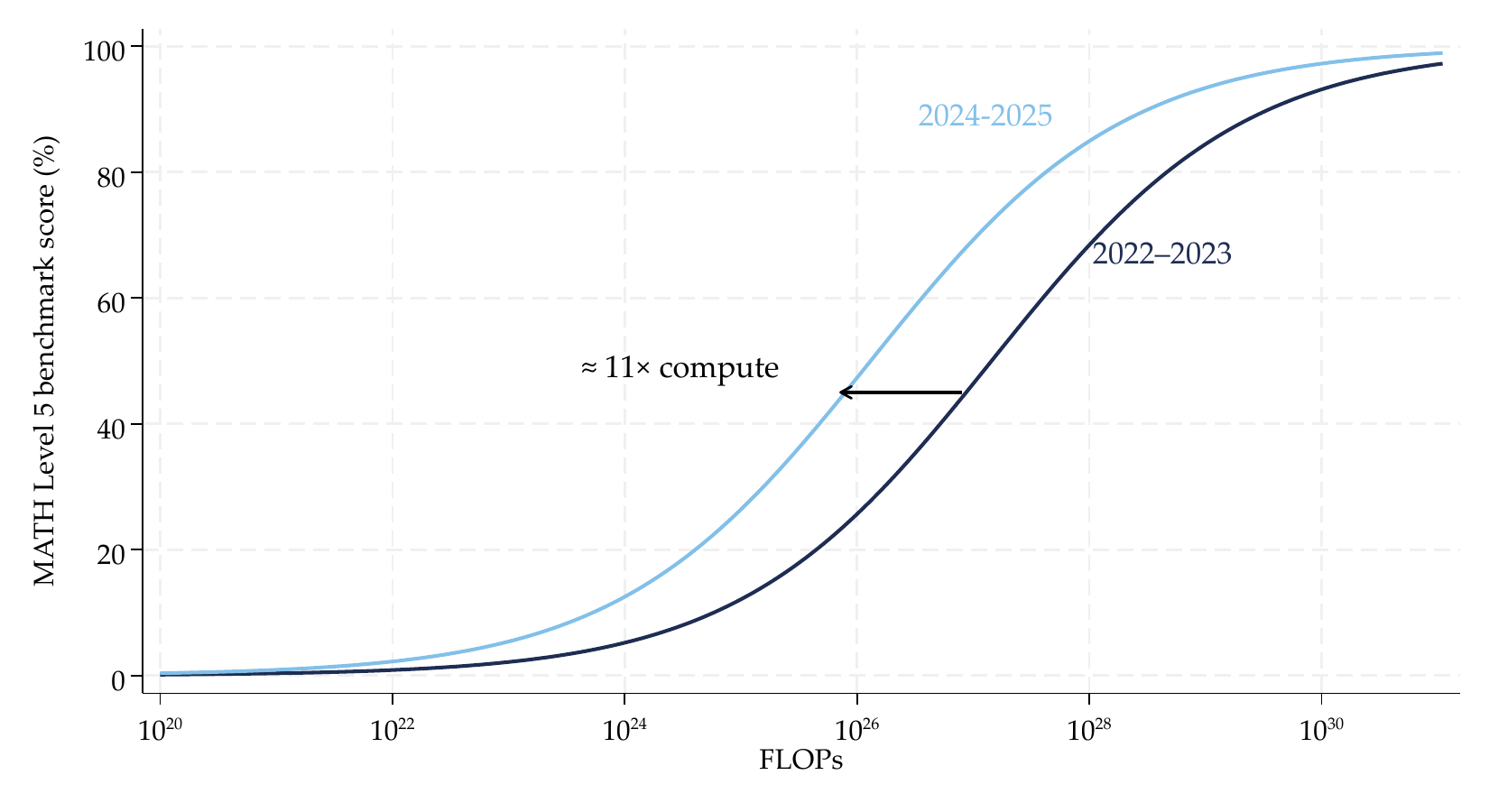}
        \vspace{-0.5em}
      
    \end{subfigure}
    
    \vspace{-0.3em} 

    \begin{subfigure}[t]{\textwidth}
        \centering
                \caption{Company and Model Effects in Compute Factors }
        \includegraphics[width=\textwidth, height=8cm, keepaspectratio]{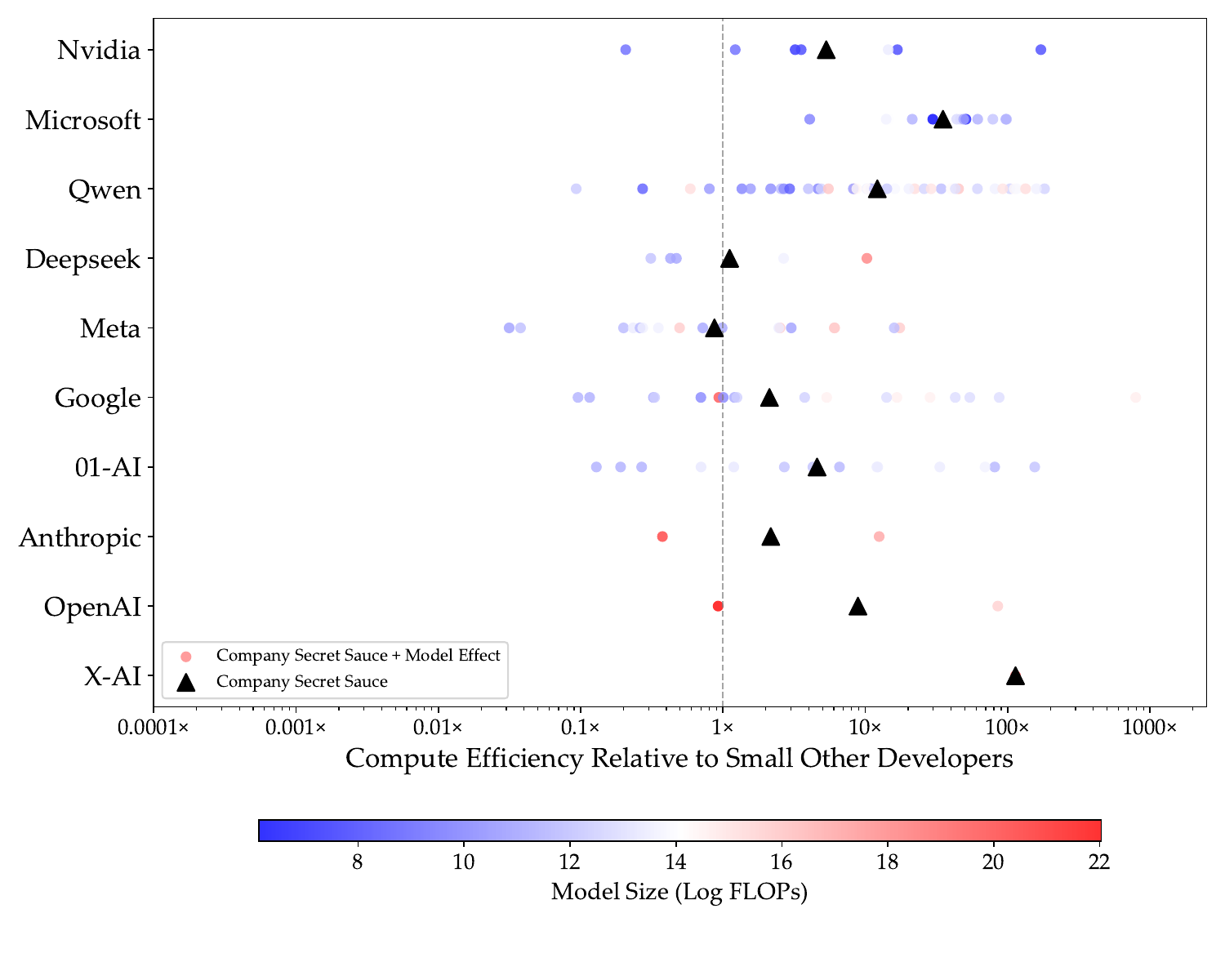}
        \vspace{-0.5em}

    \end{subfigure}
    
    \begin{subfigure}[t]{0.49\textwidth}
        \centering
              \caption{Company Effects and Relative Model Size}
        \includegraphics[width=\textwidth, height=5cm, keepaspectratio]{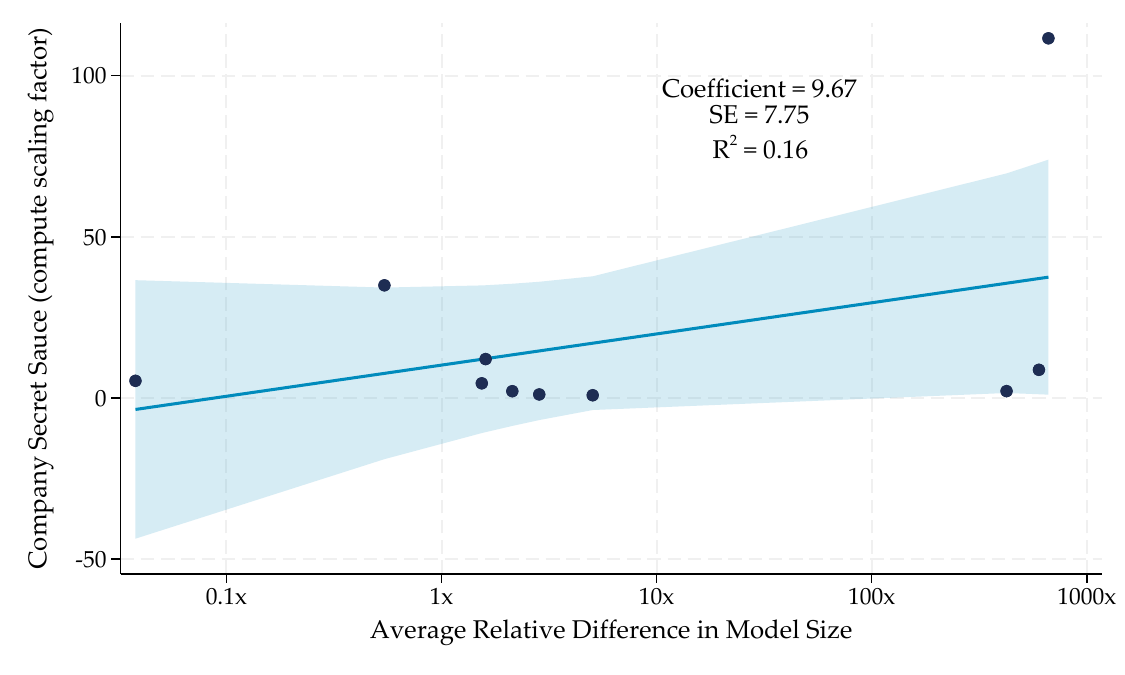}
        \vspace{-0.5em}
  
    \end{subfigure}
    \hfill
    \begin{subfigure}[t]{0.49\textwidth}
        \centering
                \caption{Model-Specific Effects in Compute Factors}
        \includegraphics[width=\textwidth, height=5cm, keepaspectratio]{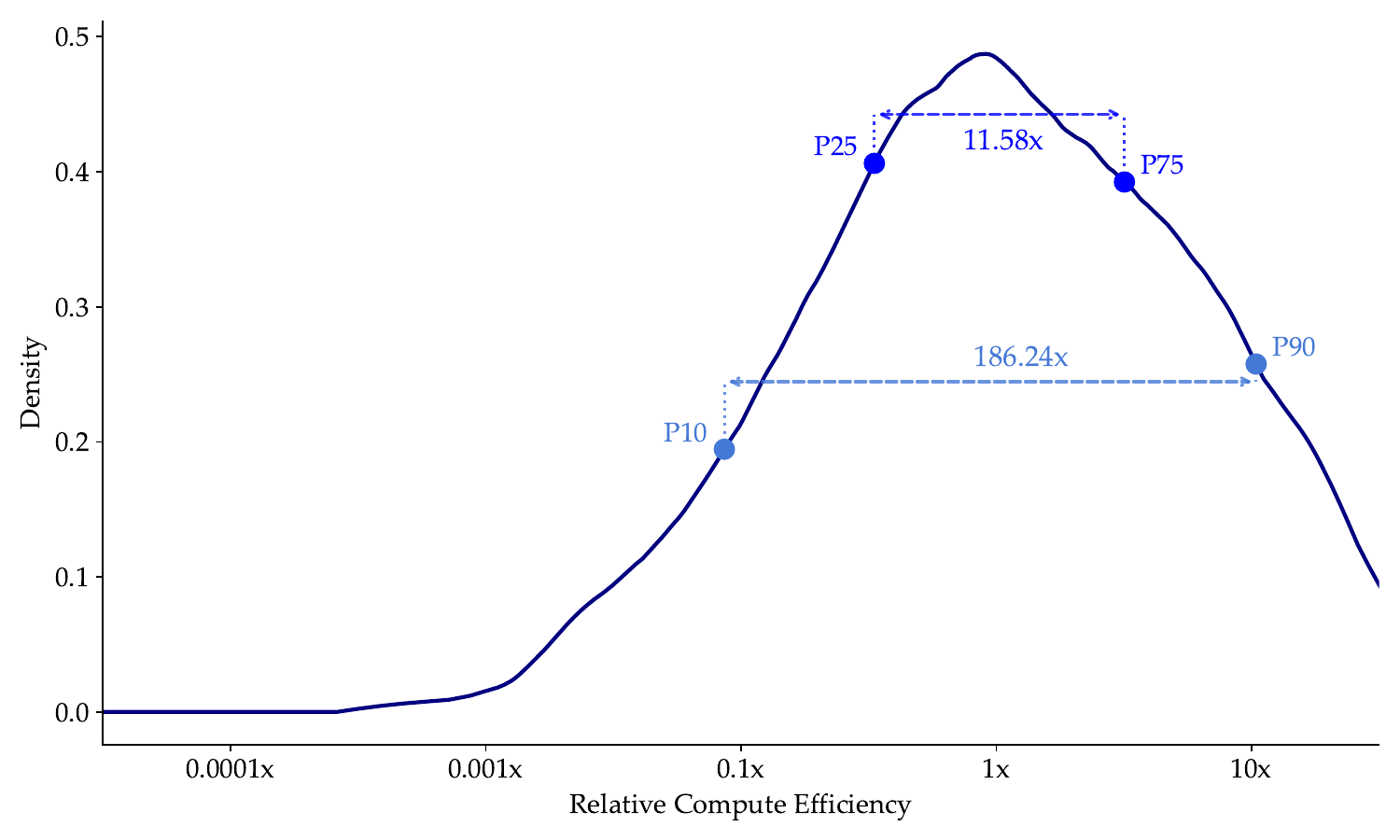}
        \vspace{-0.5em}

    \end{subfigure}
    
    \vspace{-0.3em} 

    \begin{minipage}{\textwidth}
        \scriptsize \singlespacing \textit{Notes:} The figure reports results for MATH Level 5 benchmark scores (785 LLMs). Panel (a) visualizes Eq. \eqref{main_reg} using a binned scatter plot that absorbs developer and period effects. Panel (b) shows the implied compute–performance curves for the first and last periods as predicted by the regression.  Panel (c) reports the estimated developer fixed effects (triangles). The shaded dots add model effects ($\varepsilon_i$) to the developer fixed effects with colors indicating the size of the models. Panel (d) uses a binned scatter plot to project company effects against the average of model size deviations from the average size of published models in a given period (log differences). Panel (e) displays the distribution of model effects/residuals ($\varepsilon_i$) for main developers. Significance levels in Panel (a) and (d): \sym{*} $p<0.10$, \sym{**} $p<0.05$, \sym{***} $p<0.01$
    \end{minipage}
\end{figure}

\begin{figure} [H]
    \caption{Contributions to Top Model Over Time}
  \label{top_models_main}
    \centering
    \begin{subfigure}[b]{0.89\textwidth}
    \caption{Top Scoring Models}
    \includegraphics[width=\textwidth]{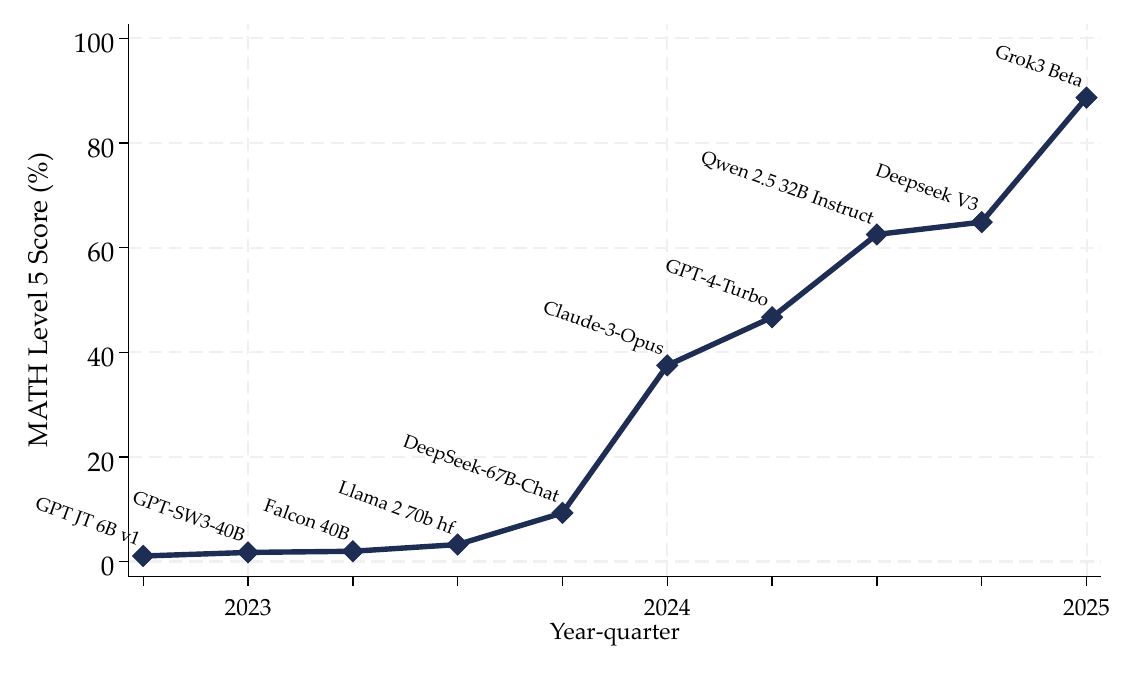}
    \end{subfigure}
    \begin{subfigure}[b]{0.89\textwidth}
    \caption{Required FLOPs for a Given MMLU-Pro Score}
    \includegraphics[width=\textwidth]{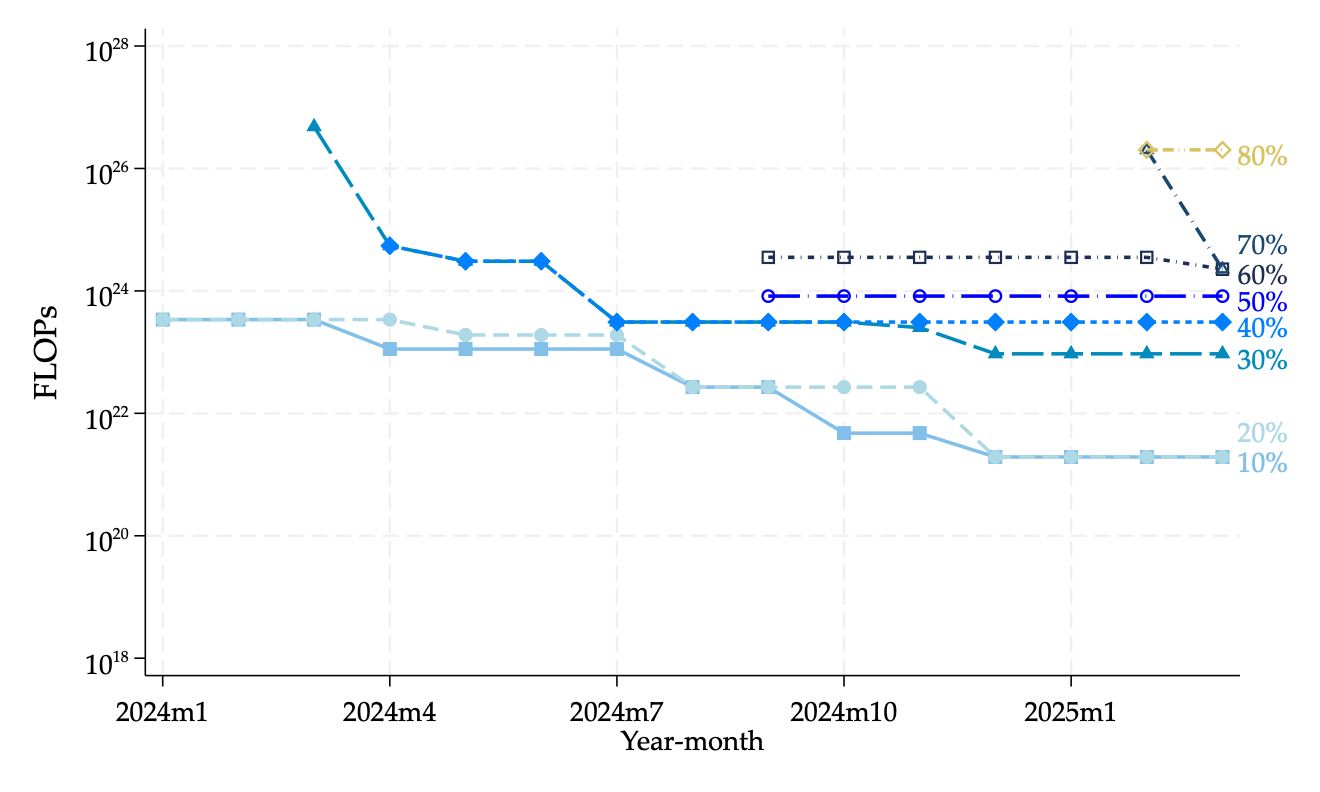}
    \end{subfigure}
                   \begin{minipage}{\textwidth}
        \scriptsize\singlespacing \textit{Notes:} Panel (a) shows the top scoring model on the MATH Level 5 benchmark each quarter. Panel (b) shows, over time, the smallest models (in FLOPs) in the data that reach a given MATH Level 5 performance level. 
     \end{minipage}

    \end{figure}

\begin{figure} [H]
    \caption{Sources of Performance Growth: Frontier Models and Smaller, Efficient Models}
  \label{Math_meek_models_main}
    \centering
    \begin{subfigure}[b]{0.95\textwidth}
    \caption{Source of Benchmark Score Growth: How Top Performing Models Become Better}
    \includegraphics[width=\textwidth]{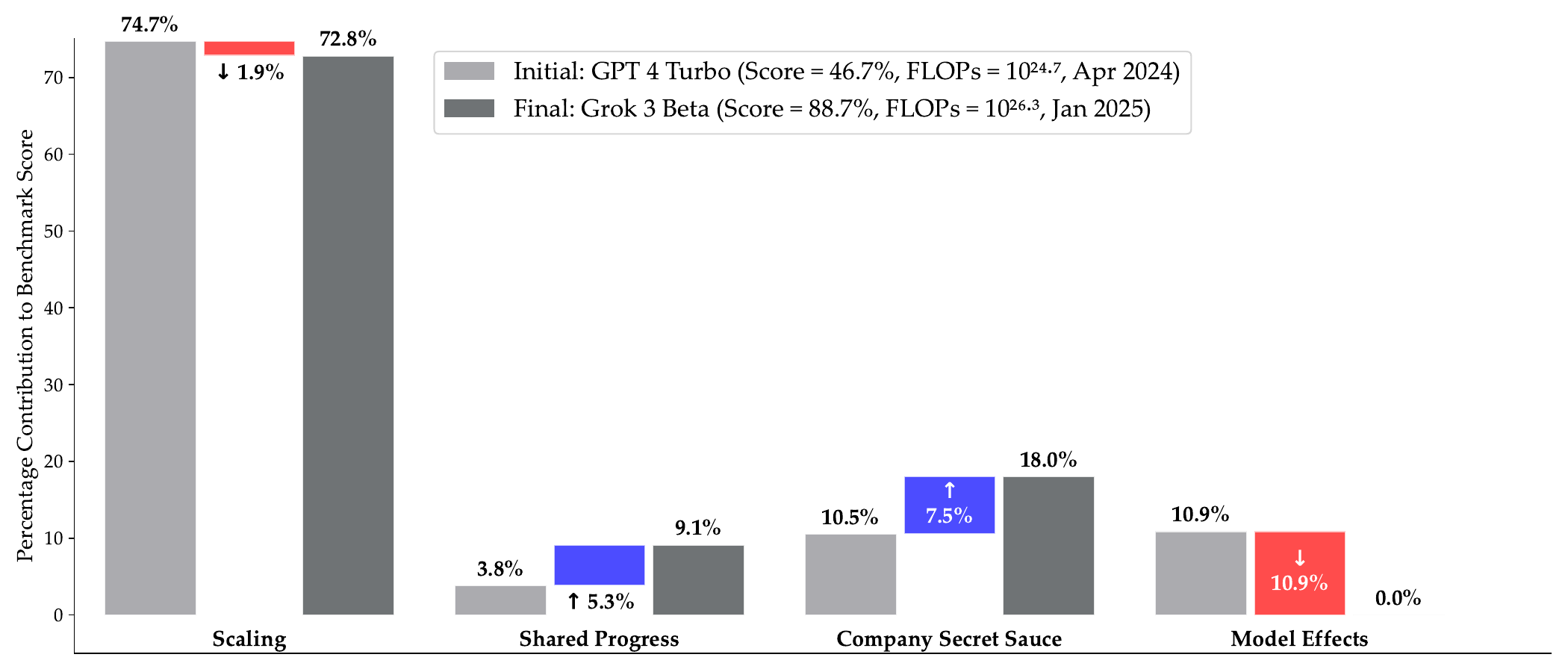}
    \end{subfigure}
    \begin{subfigure}[b]{0.95\textwidth}
    \caption{Source of Benchmark Score Growth: How Models Become More Efficient}
    \includegraphics[width=\textwidth]{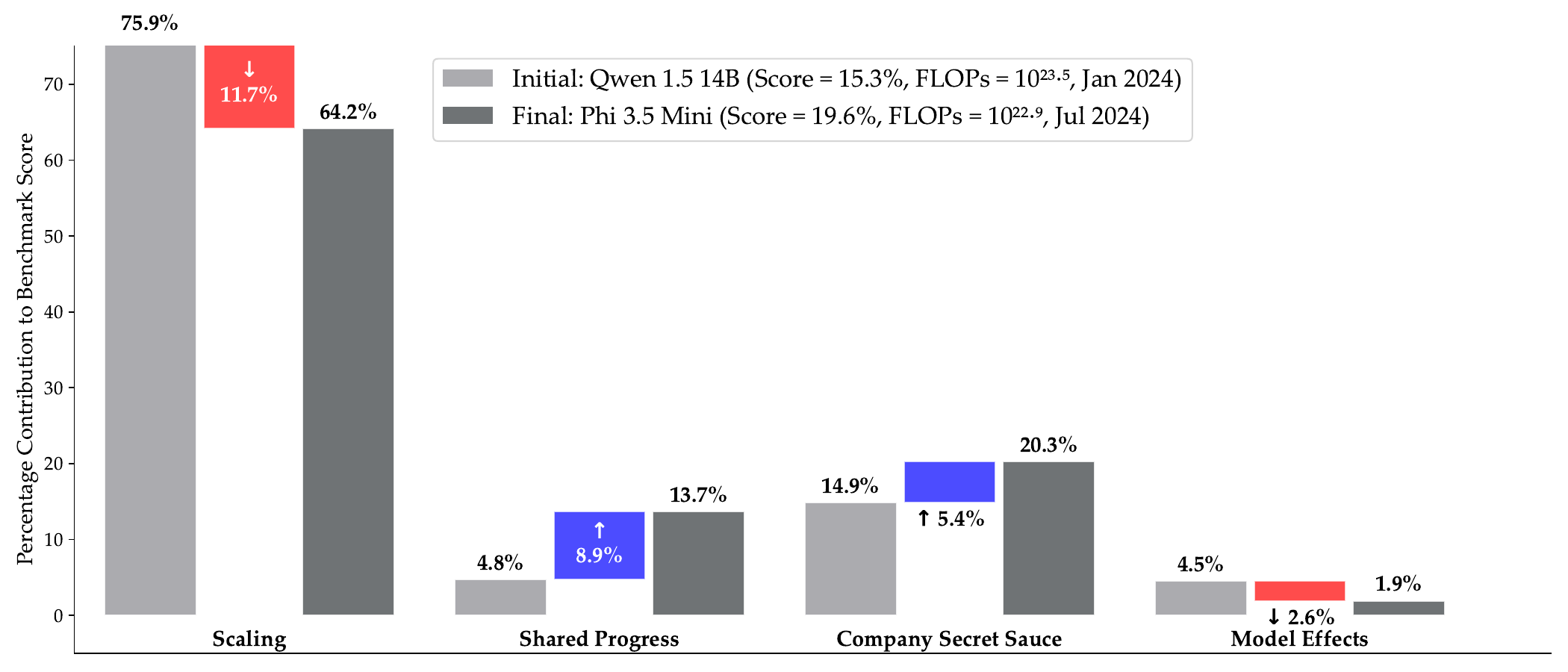}
    \end{subfigure}
          \begin{minipage}{\textwidth}
        \scriptsize\singlespacing \textit{Notes:} Panel (a) applies our regression-based decomposition to the highest-performing models for MATH Level 5 in the first and last time periods of our sample. Panel (b) applies our regression-based decomposition to the first model and the smallest models to achieve at least 15\% on MATH Level 5. We compute contribution shares of scaling, shared progress, company effects, and model effects by summing the estimated components of Eq. \eqref{main_reg} and calculating each component’s share of the total. These shares are then applied to the benchmark scores.
     \end{minipage}
    \end{figure}

\newpage

\section{Robustness Test for the Empirical Specification}\label{robustness}
 \counterwithin{equation}{section}
\renewcommand\theequation{\thesection\arabic{equation}}
\renewcommand\thefigure{\thesection\arabic{figure}}
\renewcommand\thetable{\thesection\arabic{table}}

\setcounter{figure}{0} \renewcommand{\thefigure}{E.\arabic{figure}}
\setcounter{table}{0} \renewcommand{\thetable}{E.\arabic{table}}
\setcounter{equation}{0} \renewcommand{\theequation}{E.\arabic{equation}}

\subsection{Additional Regression Specifications} \label{flex_fits}
Here, we tests the sensitivity of our regression analysis to alternative logistic transformations using a generalized logistic function $F(y_{i})= \left(\left(\frac{y^{v}_{i}}{1-y^{v}_{i}}\right)-b\right)/a$. Because the data are not rich enough to fit all parameters jointly, we first fit a two-parameter version with $v=1$ (as in Figure \ref{fig: raw_data_scatter}) and then conduct sensitivity tests for $v=0.7$ and $v=1.2$. Our results are very similar under these alternative specifications.

\subsubsection{General logistic function with scale parameter $v=0.7$ }

\begin{figure} [H]
  \label{fig:fit_asy_07}
    \centering
    \caption{Data Fit Asymmetric Logit of 0.7 and our Baseline Symmetric Logit}
    \includegraphics[width=0.8\textwidth]{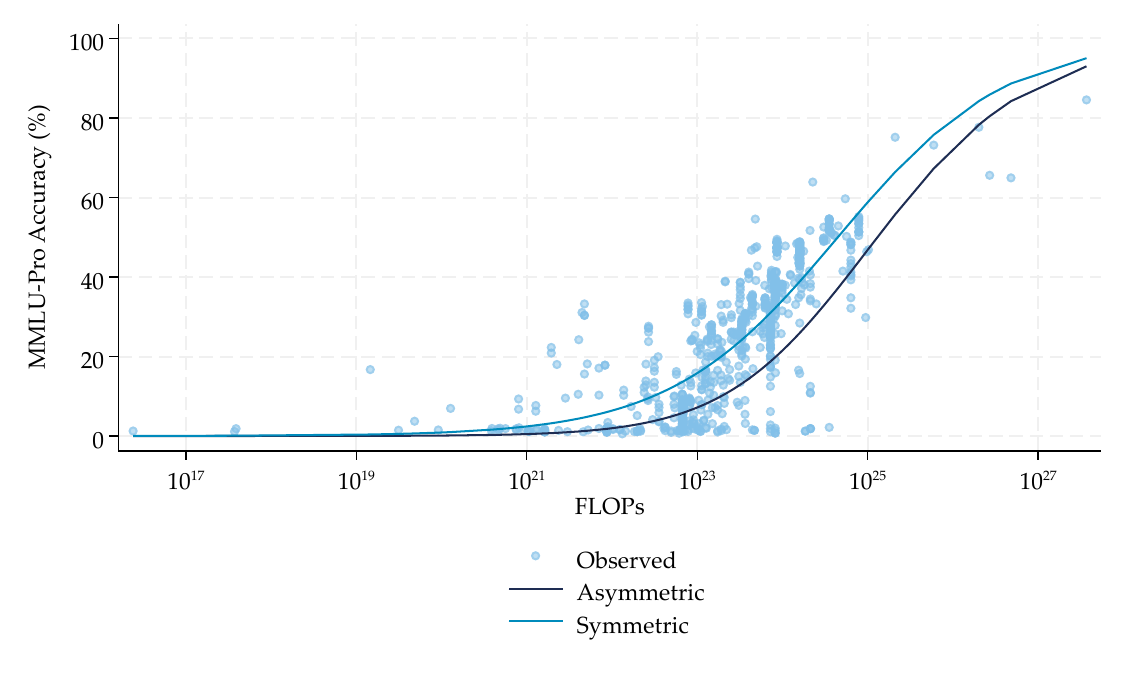}
       \begin{minipage}{\textwidth}
        \scriptsize \textit{Notes:} The figure show the raw data of MMLU-Pro benchmark scores (vertical axis) against log FLOPs (horizontal axis). We fit two functions through the data: a two-parameter logistic curve as in Figure \ref{fig: raw_data_scatter} and an augmented version of this curve with a shape parameter $v=0.7$.
     \end{minipage}
    \end{figure}

    \begin{figure} [H]
  \label{fig: shap07}
    \centering
    \caption{Shapley $R^2$ Decomposition: Asymmetric Logit 0.7}
    \includegraphics[width=0.8\textwidth]{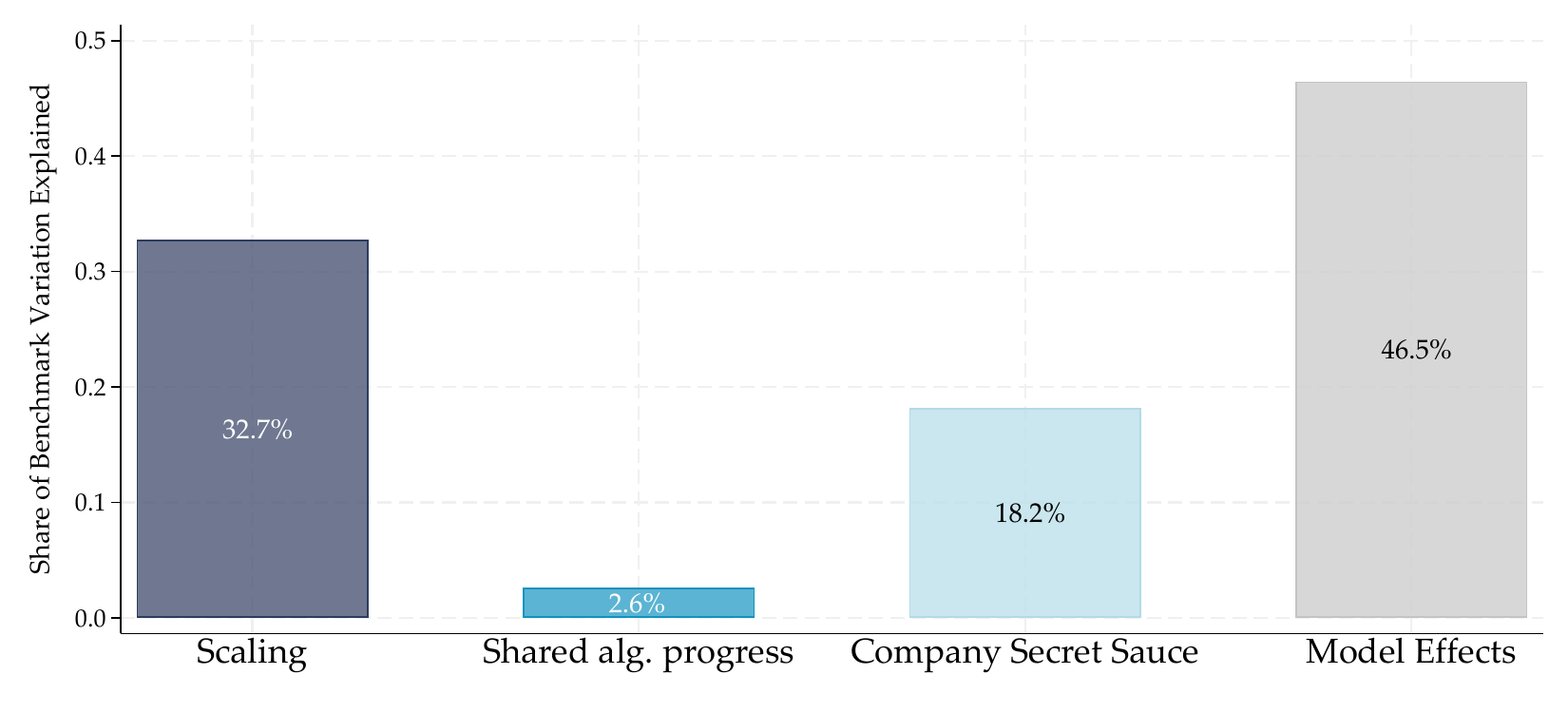}
                \begin{minipage}{0.8\textwidth}
        \scriptsize \textit{Notes:} The figure reports a Shapley decomposition of the regression $R^2$ into contributions from scale, shared algorithmic progress,
and company factors using our baseline sample and an asymmetric logit transformation with shape parameter $v=0.7$ using the command shapley2 in Stata. Model effects captures all variance in MMLU-pro benchmark score not explained by scaling, shared algorithmic progress, or company effects. 
    \end{minipage}
    \end{figure}

\begin{figure}[H]
    \caption{Main Regression results: Asymmetric Logit: 0.7}
    \label{fig: mainresults}
    \centering
    \begin{subfigure}[t]{0.49\textwidth}
        \centering
            \caption{Scaling and Performance}
        \includegraphics[width=\textwidth, height=5cm, keepaspectratio]{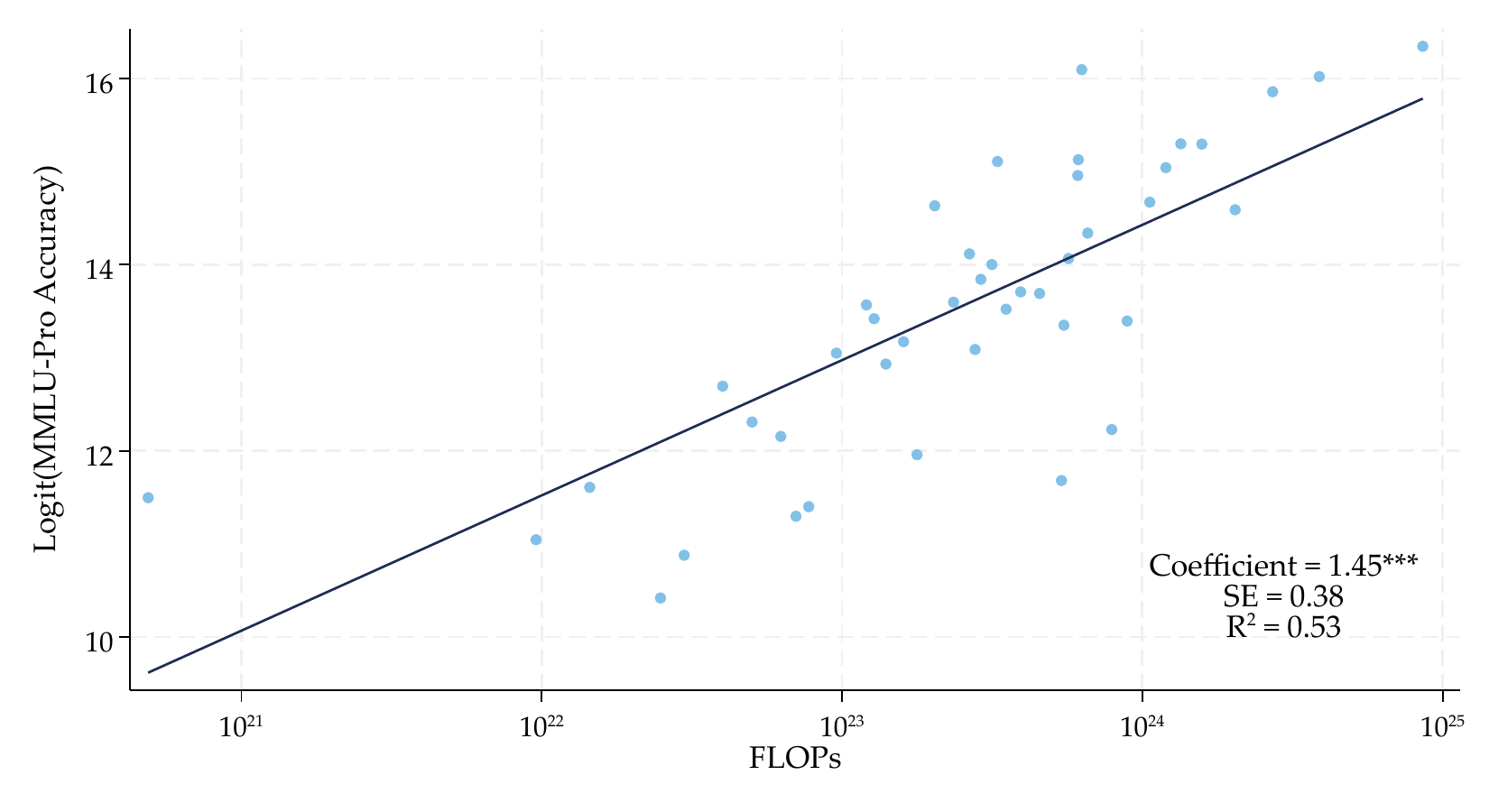}
        \vspace{-0.5em}
    
    \end{subfigure}
    \hfill
    \begin{subfigure}[t]{0.49\textwidth}
        \centering
          \caption{Shared Algorithmic Progress: Compute Factor Gain}
        \includegraphics[width=\textwidth, height=5cm, keepaspectratio]{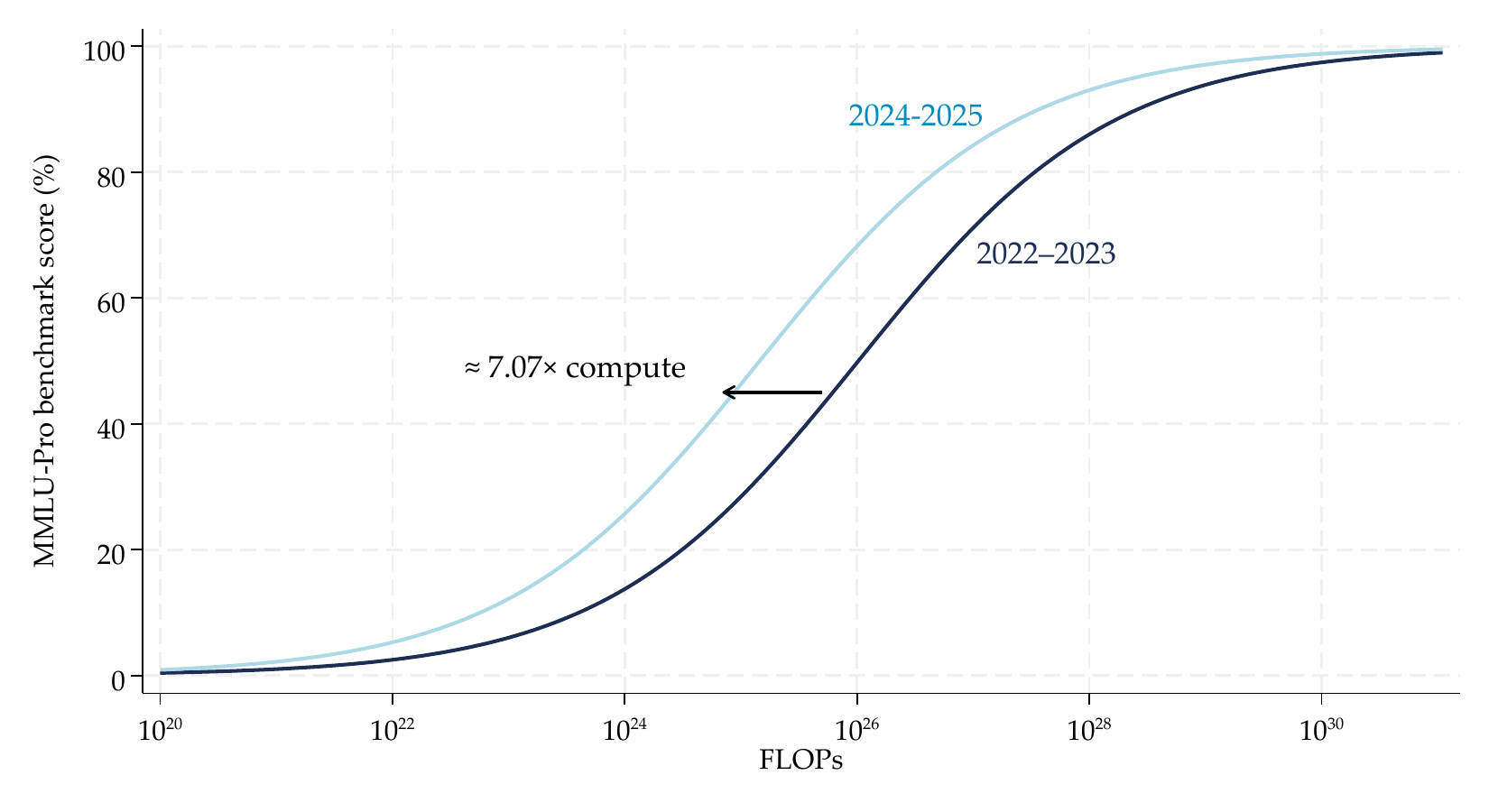}
        \vspace{-0.5em}
      
    \end{subfigure}
    
    \vspace{-0.3em} 

    \begin{subfigure}[t]{\textwidth}
        \centering
                \caption{Company and Model Effects in Compute Factors }
        \includegraphics[width=\textwidth, height=8cm, keepaspectratio]{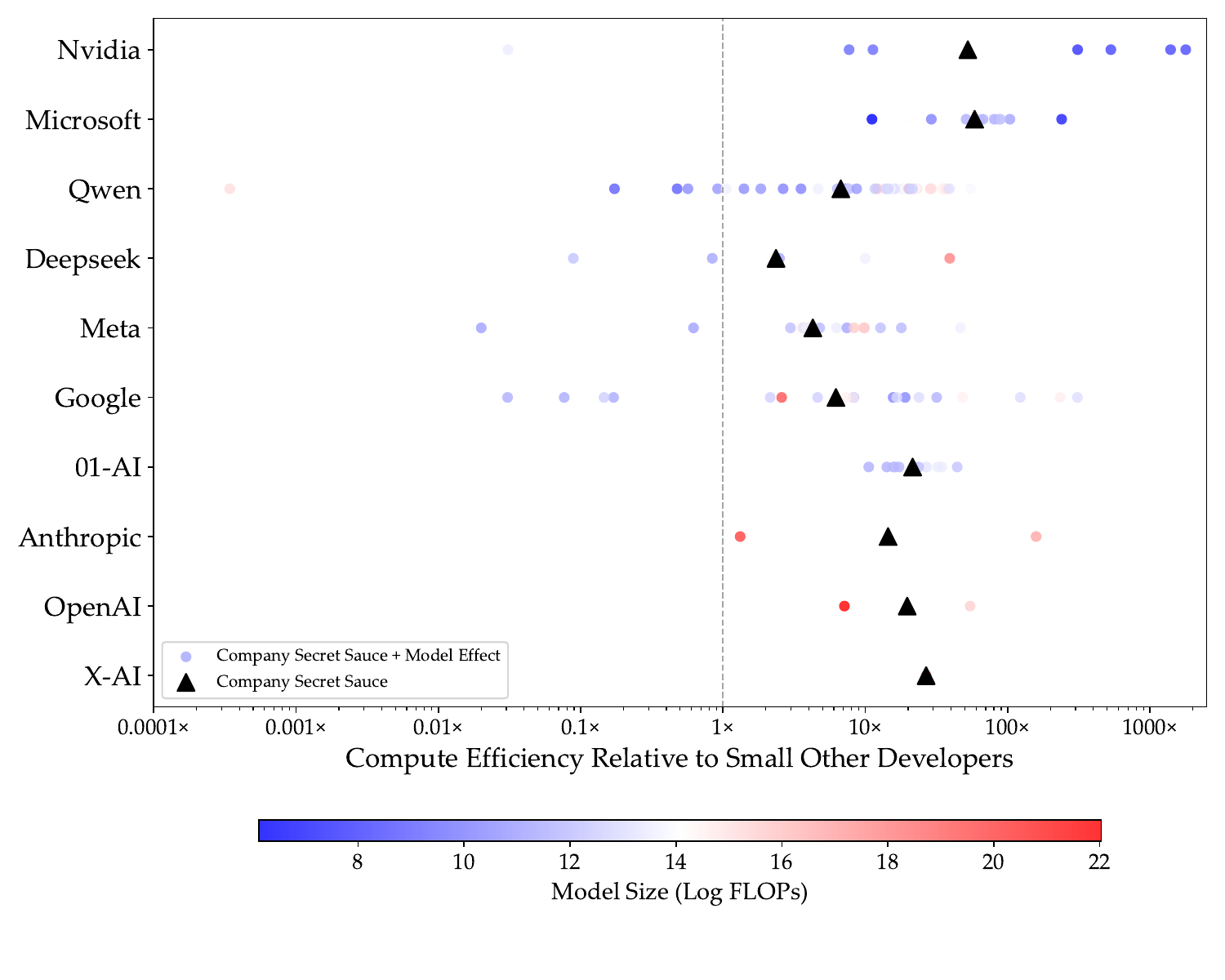}
        \vspace{-0.5em}

    \end{subfigure}
    
    \begin{subfigure}[t]{0.49\textwidth}
        \centering
              \caption{Company Effects and Relative Model Size}
        \includegraphics[width=\textwidth, height=5cm, keepaspectratio]{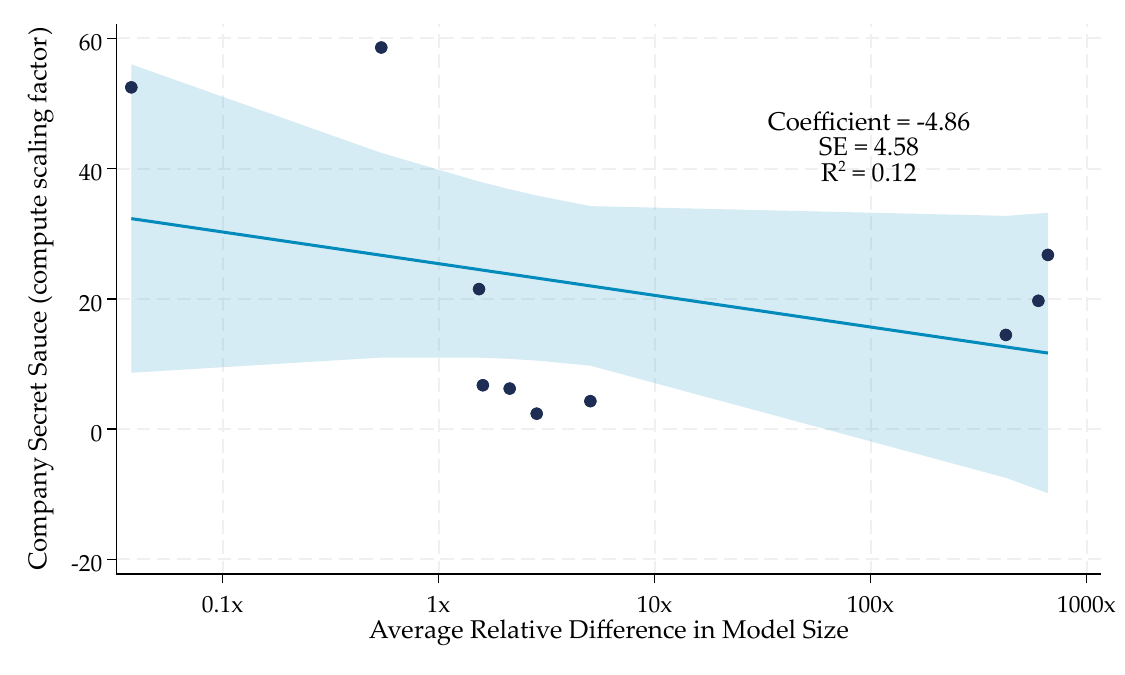}
        \vspace{-0.5em}
  
    \end{subfigure}
    \hfill
    \begin{subfigure}[t]{0.49\textwidth}
        \centering
                \caption{Model-Specific Effects in Compute Factors}
        \includegraphics[width=\textwidth, height=5cm, keepaspectratio]{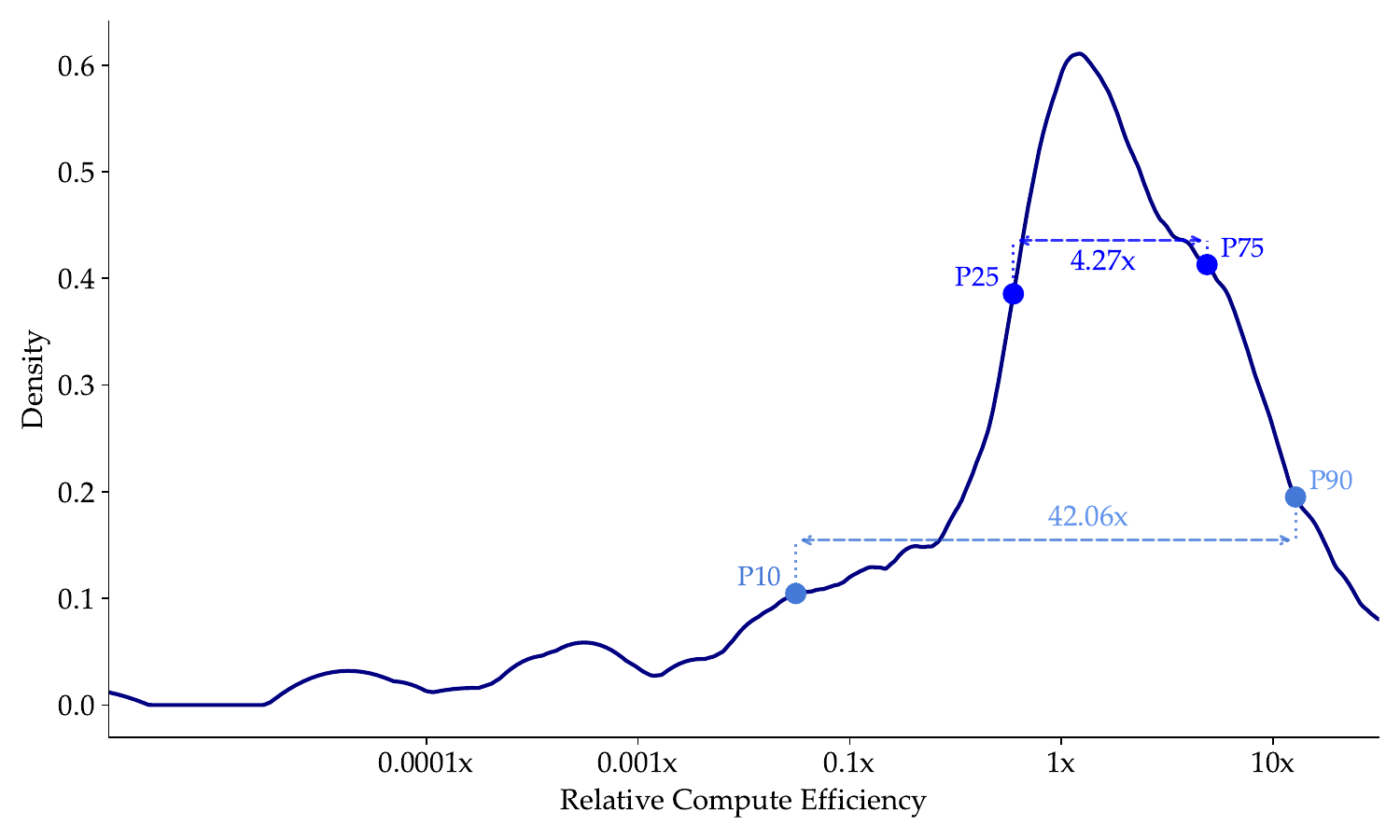}
        \vspace{-0.5em}

    \end{subfigure}
    
    \vspace{-0.3em} 

    \begin{minipage}{\textwidth}
        \scriptsize \singlespacing \textit{Notes:} The figure reports results for an asymmetrically transformed MMLU-Pro Benchmark scores  with shape parameter $v=0.7$. Panel (a) visualizes Eq. \eqref{main_reg} using a binned scatter plot that partials out developer and period effects. Panel (b) shows the implied compute–performance curves for the first and last periods as predicted by the regression.  Panel (c) reports the estimated developer fixed effects (triangles). The shaded dots add model effects ($\varepsilon_i$) to the developer fixed effects with colors indicating the size of the models Panel (d) uses a binned scatter plot to project company effects against the average of model size deviations from the average size of published models in a given period (log differences). Panel (e) displays the distribution of $\varepsilon_i$  for main developers and highlights the 90–10 percentile gap alongside the corresponding implied compute-factor difference.
    \end{minipage}
\end{figure}

\begin{figure} [H]
    \caption{Sources of Performance Growth: Frontier Models and Smaller, Efficient Models: Asymmetric Logit 0.7}
  \label{meek_models_main}
    \centering
    \begin{subfigure}[b]{0.95\textwidth}
    \caption{Source of Benchmark Score Growth: How Top Performing Models Become Better}
    \includegraphics[width=\textwidth]{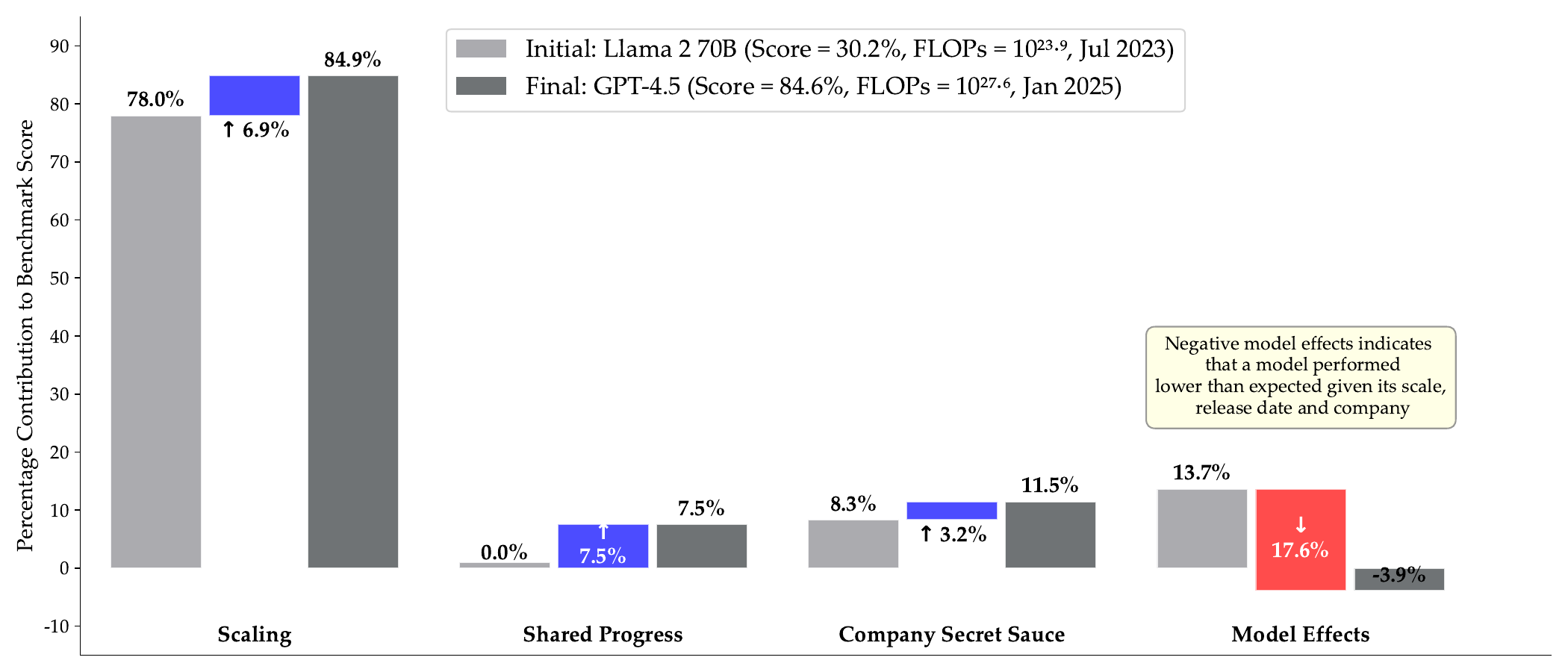}
    \end{subfigure}
    \begin{subfigure}[b]{0.95\textwidth}
    \caption{Source of Benchmark Score Growth: How Models Become More Efficient}
    \includegraphics[width=\textwidth]{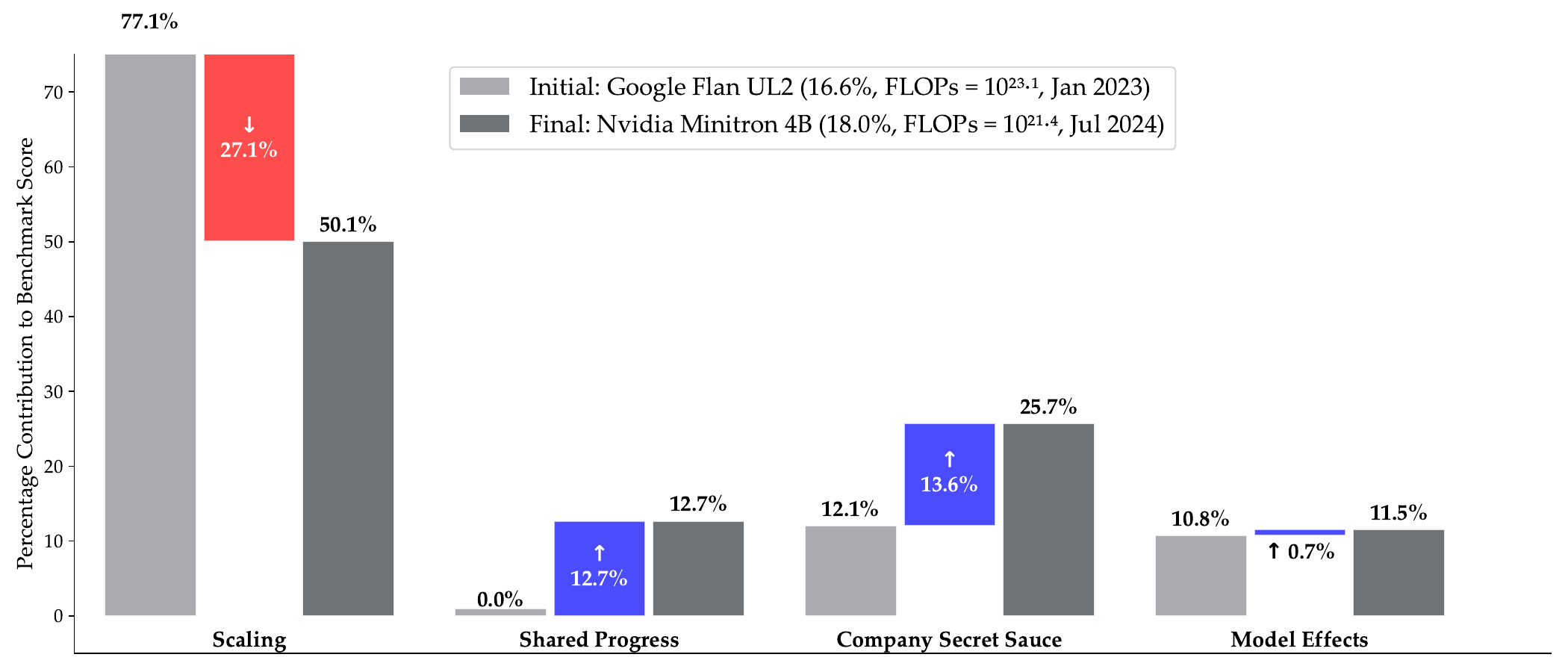}
    \end{subfigure}
          \begin{minipage}{\textwidth}
        \scriptsize\singlespacing \textit{Notes:} Panel (a) applies our regression-based decomposition with the asymmetric logit transformation that uses a shape parameter $v=0.7$ to the highest-performing models for MMLU-Pro in the first and last time periods of our sample. Panel (b) applies our regression-based decomposition to the first model and the smallest models to achieve at least 20\% on MMLU-Pro. We compute contribution shares of scaling, shared progress, company effects, and model effects by summing the estimated components of Eq. \eqref{main_reg} and calculating each component’s share of the total. These shares are then applied to the benchmark scores.
     \end{minipage}
    \end{figure}

\clearpage
\newpage
\subsubsection{General logistic function with scale parameter $v=1.2$}

\begin{figure} [H]
    \caption{Asymmetric Logit: Flatter Upper Tail}
  \label{fig:12 logit_fit_data}
    \centering
    \includegraphics[width=\textwidth]{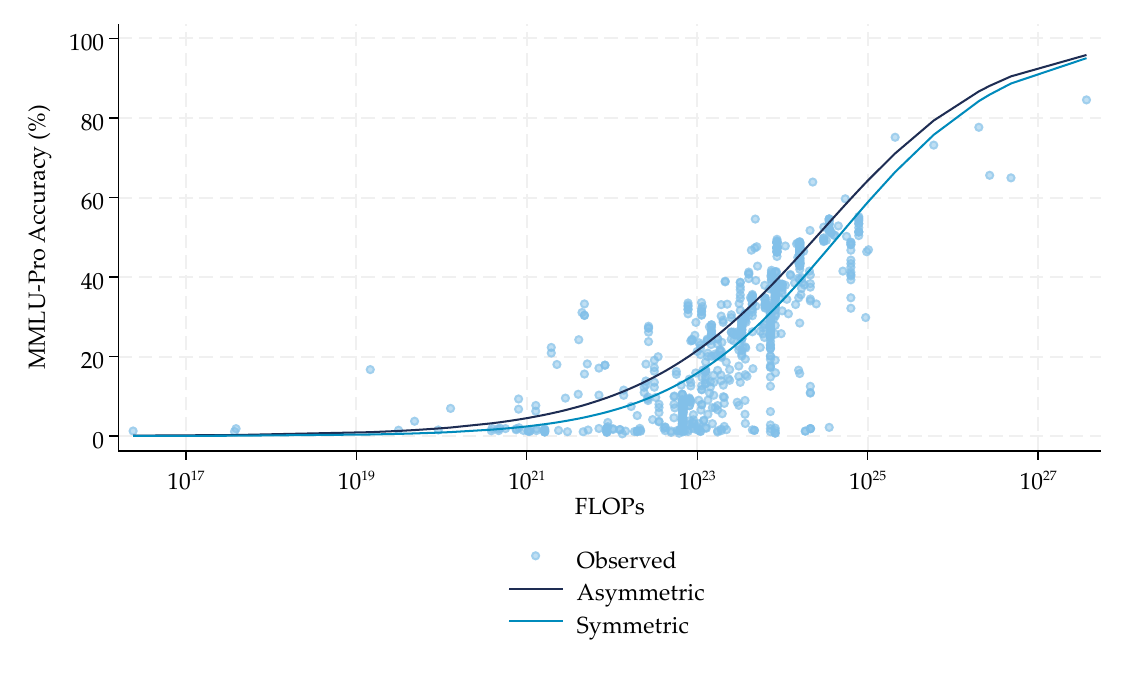}
          \begin{minipage}{\textwidth}
        \scriptsize \textit{Notes:} The figure show the raw data of MMLU-Pro benchmark scores (vertical axis) against log FLOPs (horizontal axis). We fit two functions through the data: a two-parameter logistic curve as in Figure \ref{fig: raw_data_scatter} and an augmented version of this curve with a shape parameter $v=1.2$.
     \end{minipage}
\end{figure}

\begin{figure} [H]
  \label{fig: shap07}
    \centering
    \caption{Shapley $R^2$ Decomposition: Asymmetric Logit 1.2}
    \includegraphics[width=0.8\textwidth]{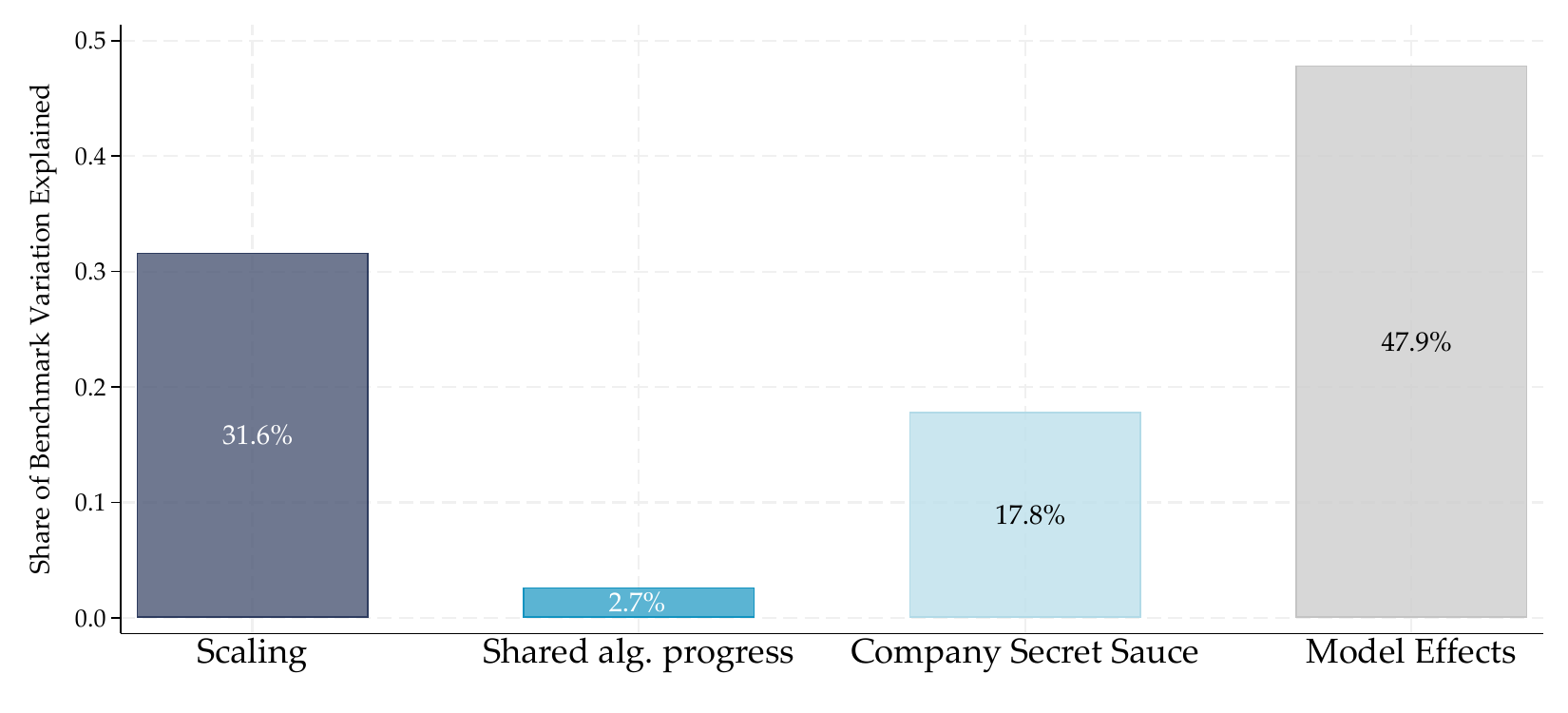}
            \begin{minipage}{0.8\textwidth}
        \scriptsize \textit{Notes:} The figure reports a Shapley decomposition of the regression $R^2$ into contributions from scale, shared algorithmic progress,
and company factors using our baseline sample and an asymmetric logit transformation with shape parameter $v=1.2$ using the command shapley2 in Stata. Model effects captures all variance in MMLU-pro benchmark score not explained by scaling, shared algorithmic progress, or company effects. 
    \end{minipage}
    \end{figure}

\begin{figure}[H]
    \caption{Main Regression results: Asymmetric Logit: 1.2}
    \label{fig: mainresults}
    \centering
    \begin{subfigure}[t]{0.49\textwidth}
        \centering
            \caption{Scaling and Performance}
        \includegraphics[width=\textwidth, height=5cm, keepaspectratio]{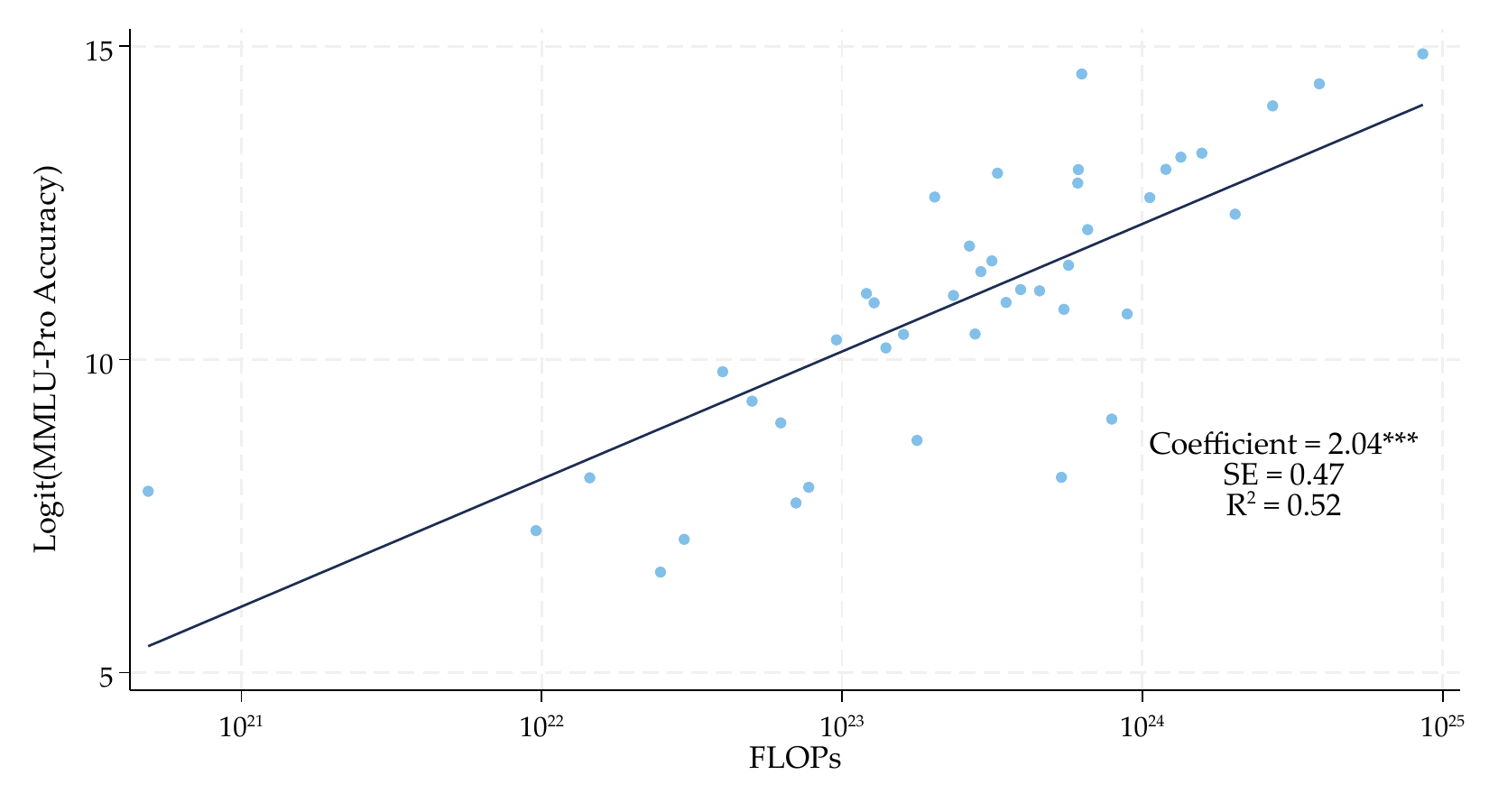}
        \vspace{-0.5em}
    
    \end{subfigure}
    \hfill
    \begin{subfigure}[t]{0.49\textwidth}
        \centering
          \caption{Shared Algorithmic Progress: Compute Factor Gain}
        \includegraphics[width=\textwidth, height=5cm, keepaspectratio]{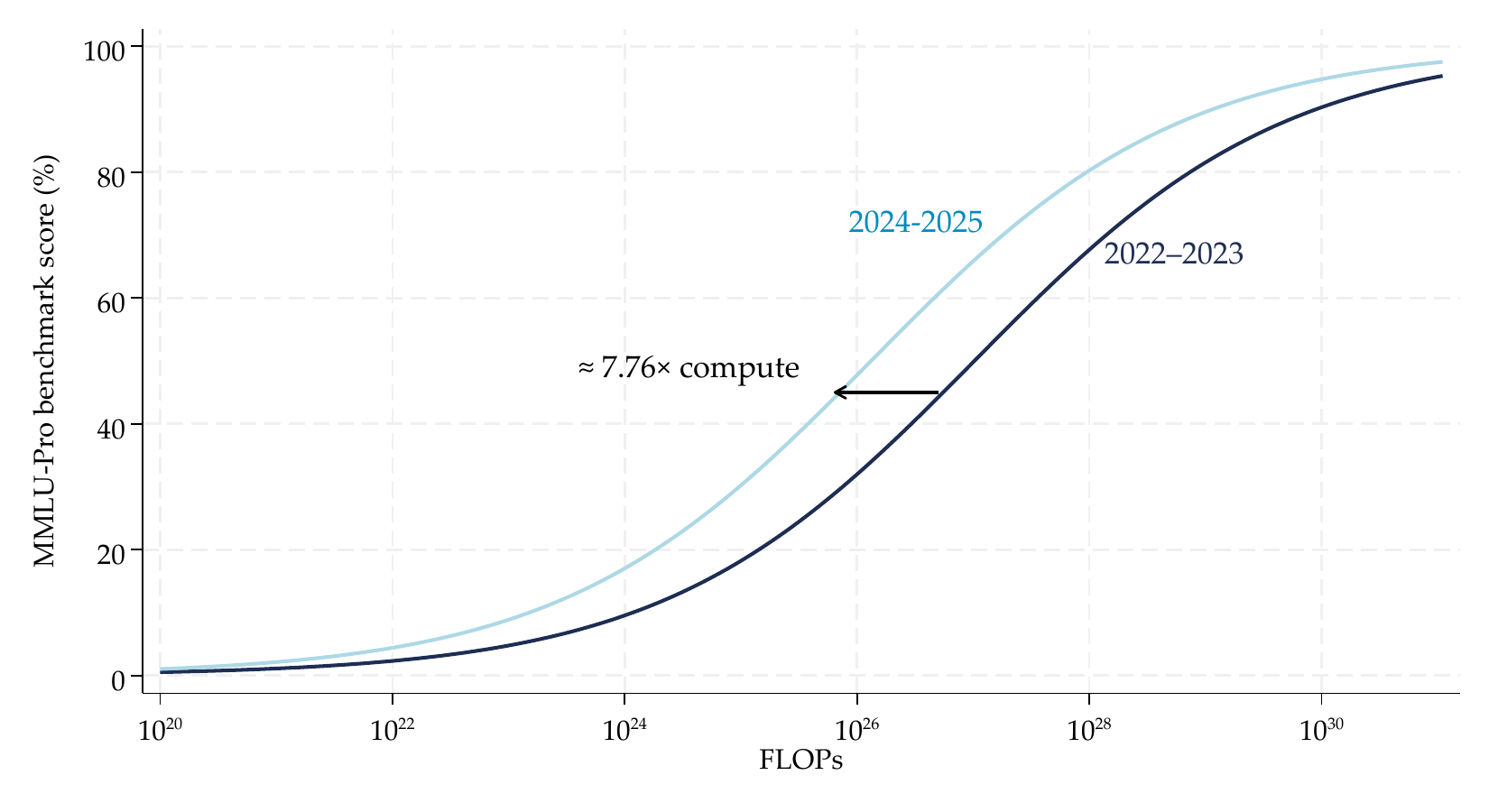}
        \vspace{-0.5em}
      
    \end{subfigure}
    
    \vspace{-0.3em} 

    \begin{subfigure}[t]{\textwidth}
        \centering
                \caption{Company and Model Effects in Compute Factors}
        \includegraphics[width=\textwidth, height=8cm, keepaspectratio]{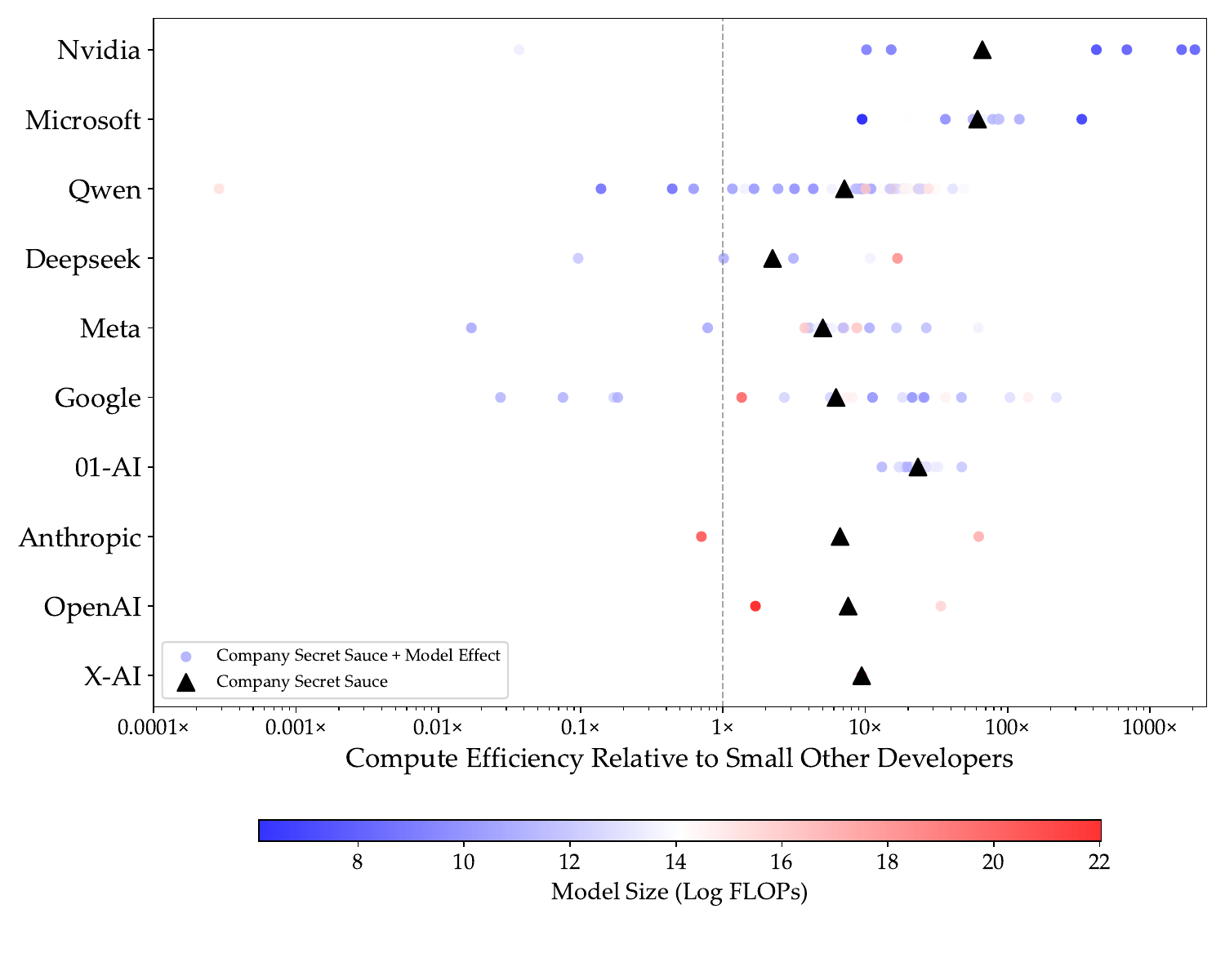}
        \vspace{-0.5em}

    \end{subfigure}
    
    \begin{subfigure}[t]{0.49\textwidth}
        \centering
              \caption{Company Effects and Relative Model Size}
        \includegraphics[width=\textwidth, height=5cm, keepaspectratio]{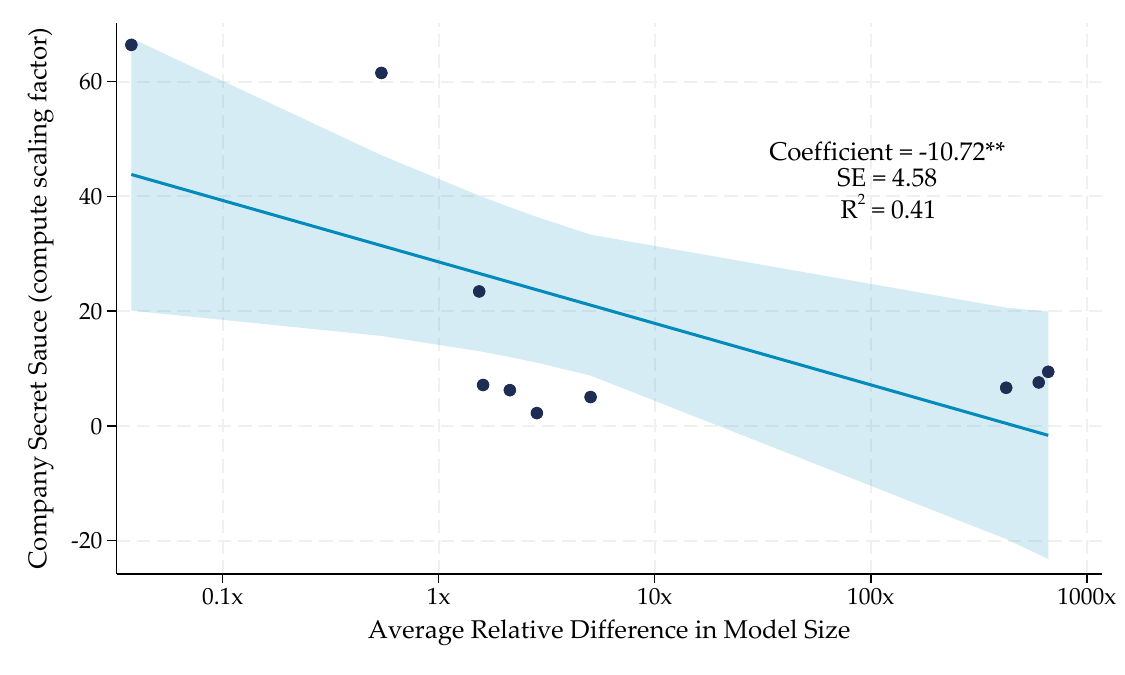}
        \vspace{-0.5em}
  
    \end{subfigure}
    \hfill
    \begin{subfigure}[t]{0.49\textwidth}
        \centering
                \caption{Model-Specific Effects in Compute Factors}
        \includegraphics[width=\textwidth, height=5cm, keepaspectratio]{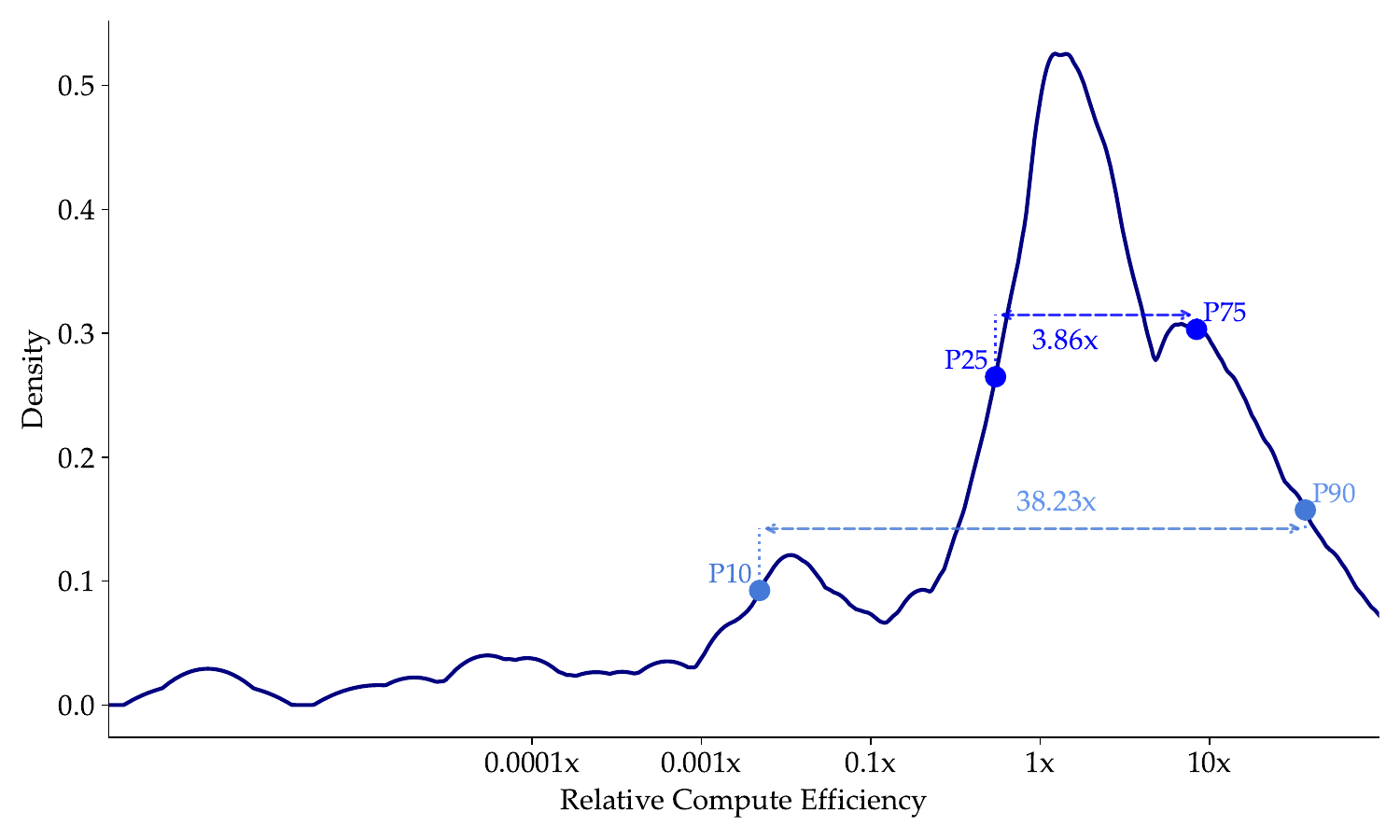}
        \vspace{-0.5em}

    \end{subfigure}
    
    \vspace{-0.3em} 

    \begin{minipage}{\textwidth}
        \scriptsize \singlespacing \textit{Notes:} The figure reports results for an asymmetrically transformed MMLU-Pro Benchmark scores with shape parameter $v=1.2$. Panel (a) visualizes Eq. \eqref{main_reg} using a binned scatter plot that partials out developer and period effects. Panel (b) shows the implied compute–performance curves for the first and last periods as predicted by the regression.  Panel (c) reports the estimated developer fixed effects (triangles). The shaded dots add model effects ($\varepsilon_i$) to the developer fixed effects with colors indicating the size of the models Panel (d) uses a binned scatter plot to project company effects against the average of model size deviations from the average size of published models in a given period (log differences). Panel (e) displays the distribution of $\varepsilon_i$  for main developers and highlights the 90–10 percentile gap alongside the corresponding implied compute-factor difference.
    \end{minipage}
\end{figure}

\begin{figure} [H]
    \caption{Sources of Performance Growth: Frontier Models and Smaller, Efficient Models: Asymmetric Logit 1.2}
  \label{meek_models_main}
    \centering
    \begin{subfigure}[b]{0.95\textwidth}
    \caption{Source of Benchmark Score Growth: How Top Performing Models Become Better}
    \includegraphics[width=\textwidth]{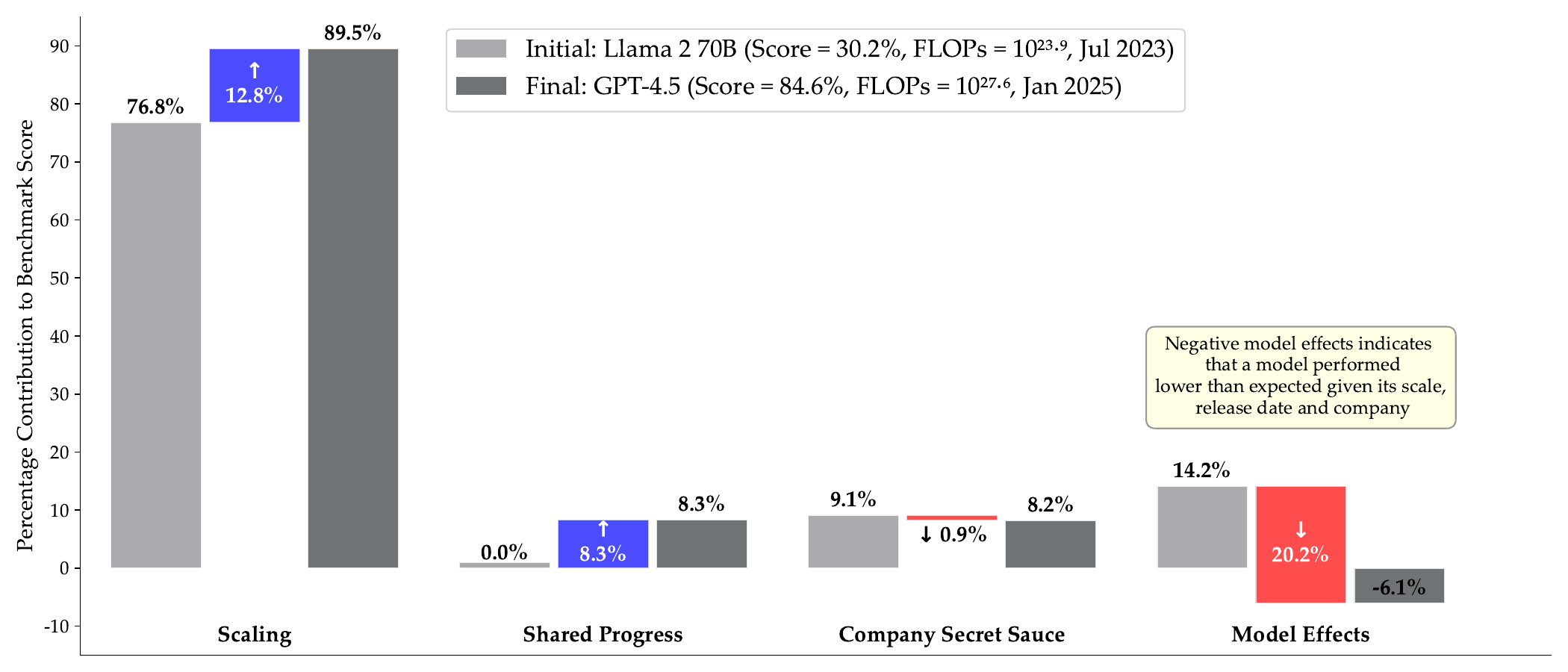}
    \end{subfigure}
    \begin{subfigure}[b]{0.95\textwidth}
    \caption{Source of Benchmark Score Growth: How Models Become More Efficient}
    \includegraphics[width=\textwidth]{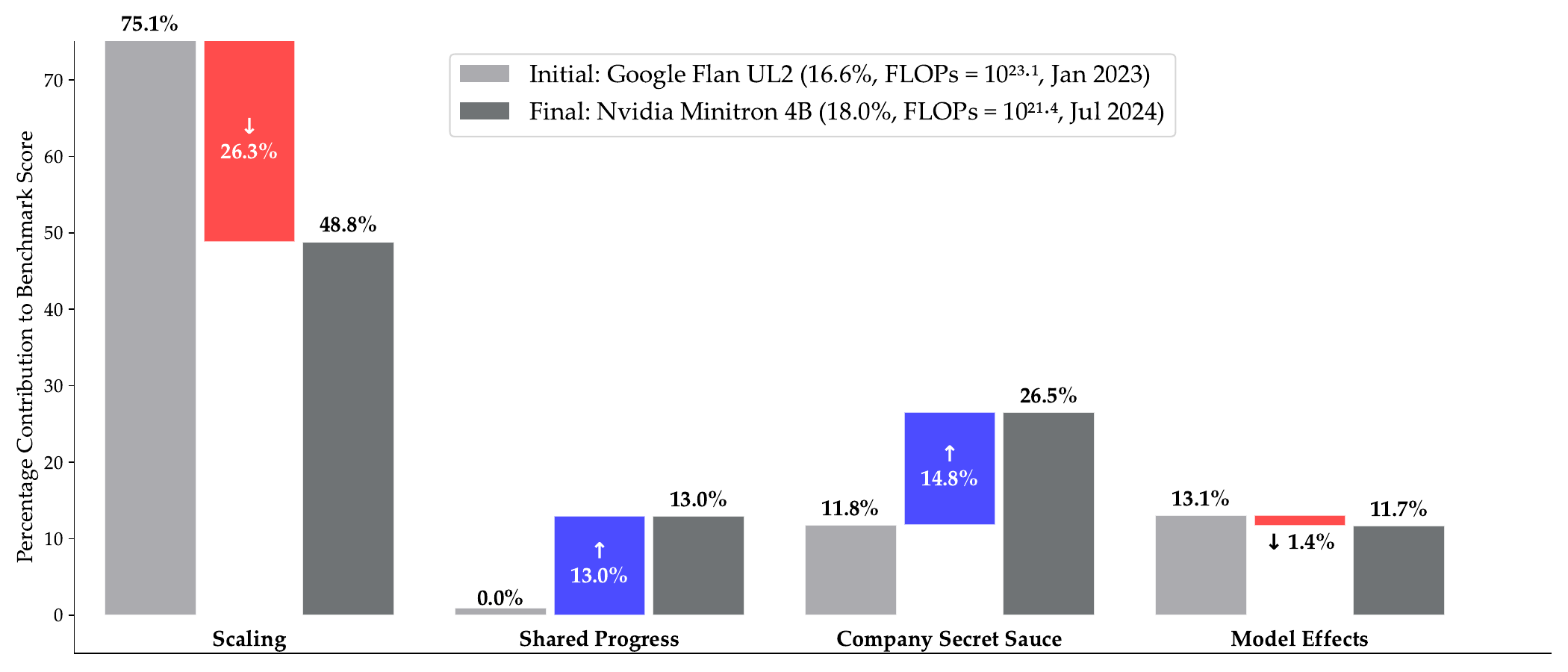}
    \end{subfigure}
          \begin{minipage}{\textwidth}
        \scriptsize\singlespacing \textit{Notes:} Panel (a) applies our regression-based decomposition with the asymmetric logit transformation that uses a shape parameter $v=1.2$ to the highest-performing models for MMLU-Pro in the first and last time periods of our sample. Panel (b) applies our regression-based decomposition to the first model and the smallest models to achieve at least 20\% on MMLU-Pro. We compute contribution shares of scaling, shared progress, company effects, and model effects by summing the estimated components of Eq. \eqref{main_reg} and calculating each component’s share of the total. These shares are then applied to the benchmark scores.
     \end{minipage}
    \end{figure}

\newpage

\subsection{Using Alternative Time Effects Structures}\label{Sec:alt_time}
\begin{figure} [H]
    \caption{Relative Compute Efficiency with Different Measures of Time}
  \label{fig: time_robust_check}
    \centering
    \includegraphics[width=0.8\textwidth]{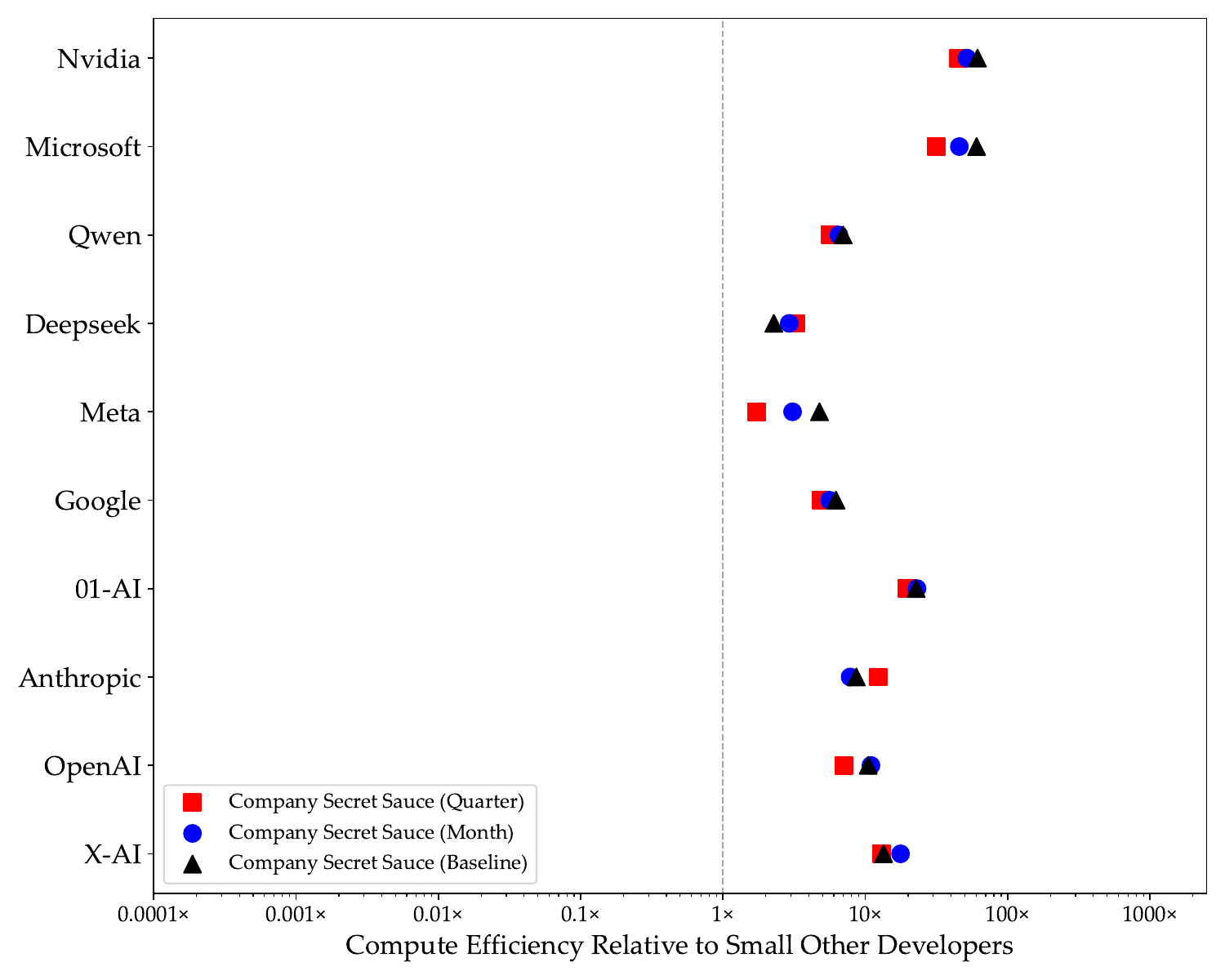}
   \begin{minipage}{\textwidth}
        \scriptsize \textit{Notes:} The figure show relative company compute factors calculated with three different sample splits of time. The black dot corresponds to our  main specification - splitting time into three periods. The red dot is a quarterly time metric with 2023q2 as the base quarter. The blue dot uses monthly time, measured continuously, with a quadratic term for extra flexibility. The three specifications show that our company secret sauce measure is robust to method of time measurement. 
     \end{minipage}
\end{figure}

\begin{figure} [H]
    \caption{Robustness Check: Shapley Decomposition with Different Time Measurements}
  \label{fig: time_robust_check_shapley}
    \centering
    \begin{subfigure}[b]{0.49\textwidth}
    \caption{Baseline Specification: Yearly}
    \includegraphics[width=\textwidth]{new_figures/shapley_stacked_bar_baseline.pdf}
    \end{subfigure}
    \begin{subfigure}[b]{0.49\textwidth}
    \caption{Robustness Check: Quarterly}
    \includegraphics[width=\textwidth]{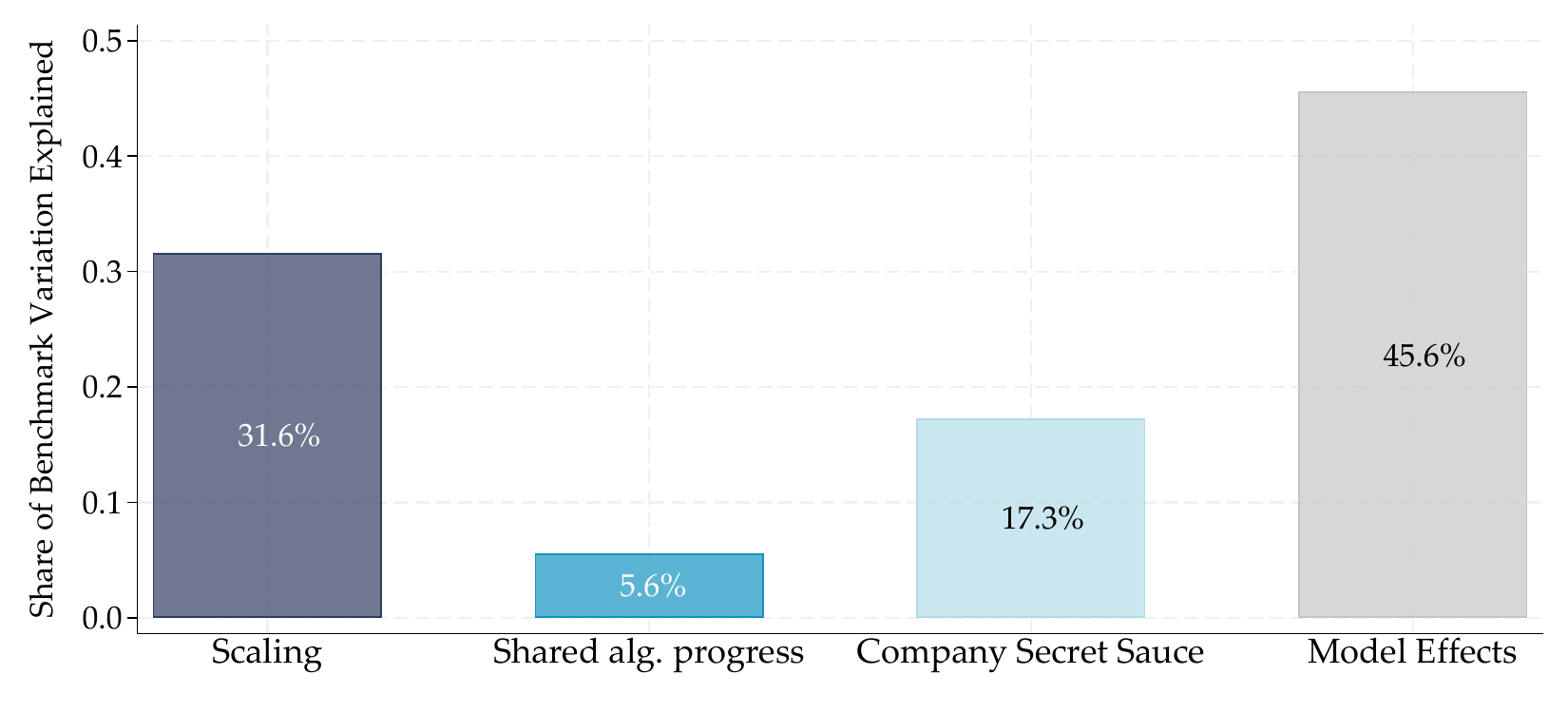}
    \end{subfigure}
    \begin{subfigure}[b]{0.49\textwidth}
    \caption{Robustness Check: Monthly Continuous Time}
    \includegraphics[width=\textwidth]{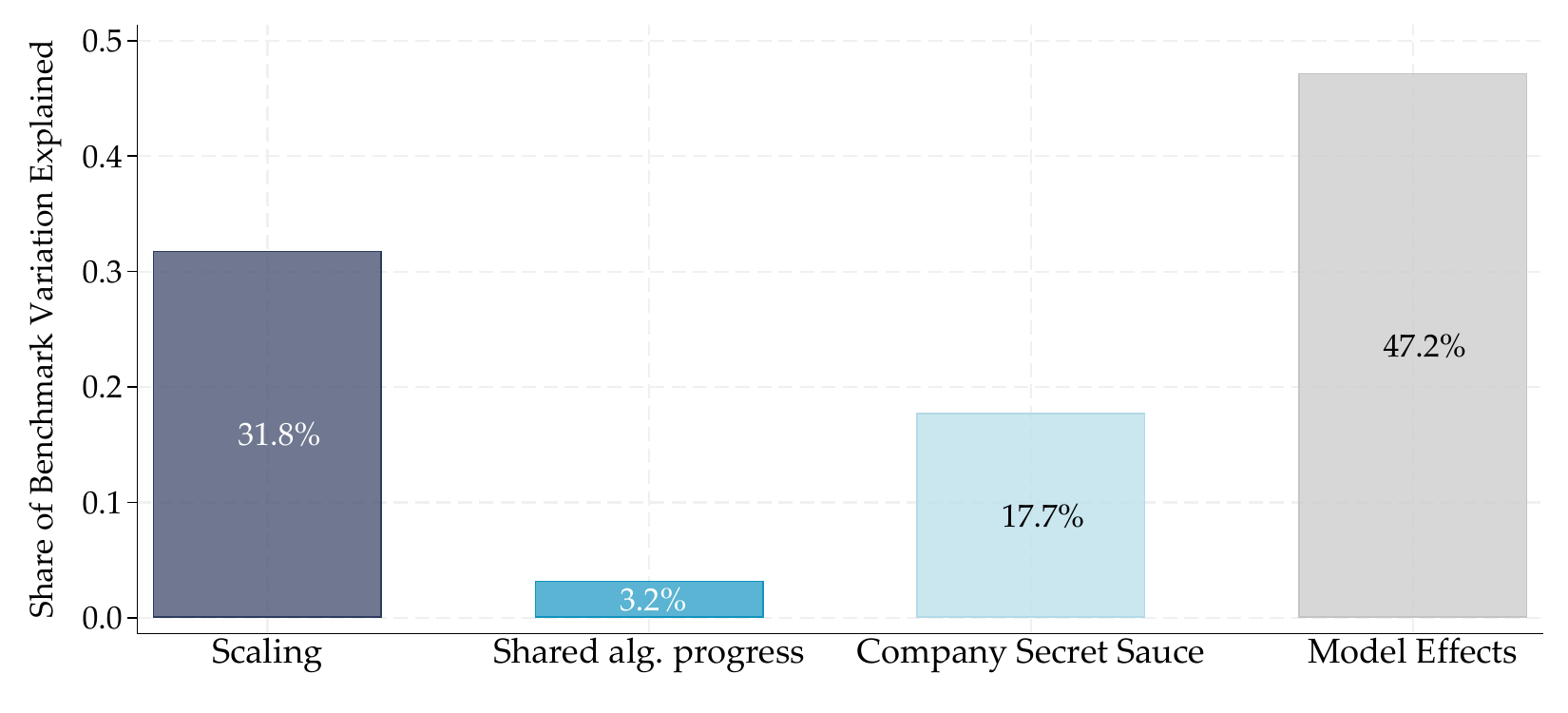}
    \end{subfigure}
                   \begin{minipage}{\textwidth}
        \scriptsize\singlespacing \textit{Notes:} All panels report a Shapley decomposition of the regression $R^2$ into contributions from scale, shared algorithmic progress and company secret sauce for the indicated sample using the command shapley2 in Stata. Model effects are the remaining residual variation. Panel (a) uses our baseline specification: time split into three chunks of when the leaderboard was active: 2022q4-2023q3, 2023q4-2024q2, and 2024q3-2025q1 (roughly yearly). Panel (b) shows time measured with quarterly indicator variables. Panel (c) shows time measured by a continuous monthly variable with a quadratic month squared term for extra flexibility. The robustness check shows that our choice of time measurement does not meaningfully change results and explained variance.
     \end{minipage}

    \end{figure}

One concern is that our time-periods are too broad, such that certain developer publication timelines could bias our estimates of secret sauce and shared algorithmic progress. Here, we use two different specifications: one with quarterly time dummies (despite some quarters have insufficient observations), and one with a linear and quadratic time trend in year-months instead of a set of dummies. Figure \ref{fig: time_robust_check} shows that, for each of these alternative specification, our secret sauce estimates are similar to our baseline. Additionally, Figure \ref{fig: time_robust_check_shapley} replicates Figure \ref{fig: shapley_main} showing that our variance decomposition results remain almost unchanged. 

\newpage

\subsection{Do Compute Coefficients Change Over Time?}\label{time_interac}

We estimate an alternative specification that interacts period fixed effects with log compute to examine whether the estimated compute elasticity varies over time. This specification tests whether shared algorithmic progress operates by shifting the compute coefficient itself (i.e., the exponent on log compute), rather than acting as a multiplicative factor that uniformly scales compute. Specifically, we estimate:

\begin{equation}
Y_i
= \beta_0
+ \beta_c \log(c_i)
+ \boldsymbol{\delta}_t
+ \boldsymbol{\nu}_j
+ \sum_{t} \beta_{c,t}\,\mathbf{1}\{T_i = t\}\,\log(c_i)
+ \varepsilon_i,
\end{equation}
where $\beta_{c,t}$  captures period-specific deviations in the compute coefficient relative to the baseline period, which remains 2022q4–2023q3. The baseline category is still 2022q4-2023q3. If shared algorithmic progress altered the exponent on compute, we would expect the interaction coefficients $\beta_{c,t}$ coefficients  to be positive and statistically significant. Table \ref{interact_period} reports the estimated interaction terms. While the coefficients are uniformly positive, none are statistically significant. We therefore find no evidence of a statistically meaningful shift in the compute elasticity over time.

\begin{table}[H]

\centering
\caption{Robustness Check: Interacting Time and Compute}
\def\sym#1{\ifmmode^{#1}\else\(^{#1}\)\fi}
\begin{tabular}{l*{1}{c}}
\hline\hline
\label{interact_period}
                    &\multicolumn{1}{c}{(1)}\\
                    &\multicolumn{1}{c}{Logit(MMLU-Pro)}\\
\hline
Log(FLOPs)          &       0.642\sym{**}  \\
                    &      (0.252)         \\
[1em]
2023q4-2024q2 $\times$ Log(FLOPs)&       0.128         \\
                    &      (0.169)         \\
[1em]
2024q3-2025q1 $\times$ Log(FLOPs)&       0.160         \\
                    &      (0.193)         \\
\hline
Company Effects          &       YES             \\
Period Effects          &       YES             \\
Observations        &         809         \\
R$^2$        &         0.53         \\

\hline\hline
\end{tabular}
                   \begin{minipage}{\textwidth}
        \scriptsize\singlespacing \textit{Notes:} The table reports the result of our main regression specification, as reported in \ref{main_reg} with an additional Log(FLOPs) $\times$ period interaction term. The baseline period is 2022q4-2023q3. Significance levels: \sym{*} $p<0.10$, \sym{**} $p<0.05$, \sym{***} $p<0.01$.
     \end{minipage}
\end{table}

\newpage
\printbibliography[segment=1, title={Appendix References}]

\end{document}